\documentclass{article}

\usepackage{natbib}
\setcitestyle{compress}
    \PassOptionsToPackage{numbers, compress}{natbib}
\usepackage[preprint]{neurips_data_2023}

\usepackage[utf8]{inputenc} 
\usepackage[T1]{fontenc}    
\usepackage{hyperref}       
\usepackage{url}            
\usepackage{booktabs}       
\usepackage{amsfonts}       
\usepackage{nicefrac}       
\usepackage{microtype}      
\usepackage[dvipsnames]{xcolor}
\usepackage{soul}
\usepackage{graphicx}
\usepackage{caption}
\usepackage{subcaption}
\usepackage{multirow}
\usepackage{listings}
\usepackage{cancel}
\usepackage{algorithm}
\usepackage{wrapfig,lipsum}
\usepackage[noend]{algpseudocode}
\usepackage[frozencache,cachedir=.]{minted}
\newminted{python}{frame=lines,framerule=2pt}
\usepackage{tablefootnote}
\usepackage{svg}
\usepackage{amsmath,amsthm}
\usepackage{enumitem}
\usepackage{pifont} 
\usepackage{tcolorbox}

\usepackage{wrapfig}

\newcommand{\pisft}{\pi^\text{SFT}} %

\colorlet{LightOrchid}{Orchid!40!White}
\colorlet{LightYellowGreen}{YellowGreen!40!White}
\colorlet{LightYellowOrange}{YellowOrange!40!White}
\colorlet{LightSalmon}{Salmon!60!White}
\colorlet{LightAquamarine}{Aquamarine!40!White}

\colorlet{LightBlue}{Blue!40!White} 
\colorlet{LightYellow}{Yellow!40!White} 
\colorlet{LightRed}{Red!40!White} 

\definecolor{detailcolor}{RGB}{70,130,180} 

\newlist{dlist}{enumerate}{1}
\setlist[dlist,1]{
  label={\bfseries\ding{227} Detail \arabic*:},
  leftmargin=*,
  align=left,
  labelsep=2mm,
}

\title{The N+ Implementation Details of RLHF with PPO: A Case Study on TL;DR Summarization}

\newcommand{\hflogo}{{\includegraphics[scale=0.10]{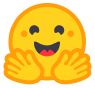}}}
\newcommand{\milalogo}{\includegraphics[scale=0.04]{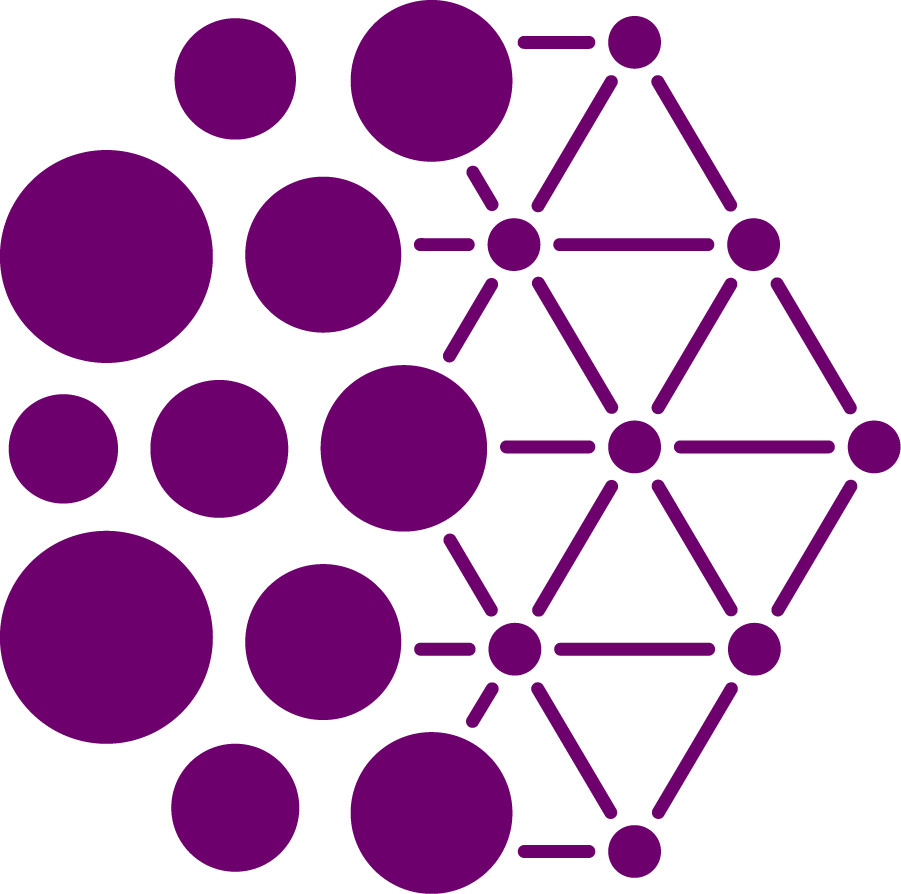}}
\newcommand{\fuxilogo}{\includegraphics[scale=0.10]{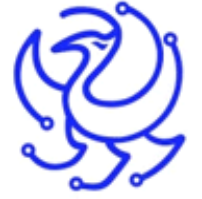}}

\author{Shengyi Huang\hflogo   \quad Michael Noukhovitch\milalogo \quad \textbf{Arian Hosseini}\milalogo\\
  \quad   \quad 
  \textbf{Kashif Rasul}\hflogo \quad  \textbf{Weixun Wang}\fuxilogo \quad \textbf{Lewis Tunstall}\hflogo
   \\
   \quad \hflogo Hugging Face \\
   \quad \milalogo Mila, Universit\'e de Montr\'eal \\
   \quad \fuxilogo Fuxi AI Lab, NetEase
   \\
  \texttt{costa@huggingface.co}
}

\begin{document}

\maketitle

\begin{abstract}

 This work is the first to openly reproduce the Reinforcement Learning from Human Feedback (RLHF) \emph{scaling behaviors} reported in OpenAI's seminal TL;DR summarization work~\citep{stiennon2020learning}. We create an RLHF pipeline from scratch, enumerate over 20 key implementation details, and share key insights during the reproduction. Our RLHF-trained Pythia models demonstrate significant gains in response quality that scale with model size with our 2.8B, 6.9B models outperforming OpenAI's released 1.3B checkpoint.
We publicly release the trained model checkpoints and code to facilitate further research and accelerate progress in the field (\url{https://github.com/vwxyzjn/summarize_from_feedback_details}). 

\end{abstract}

\begin{figure}[h]
    \centering
     \includegraphics[width=0.6\linewidth]{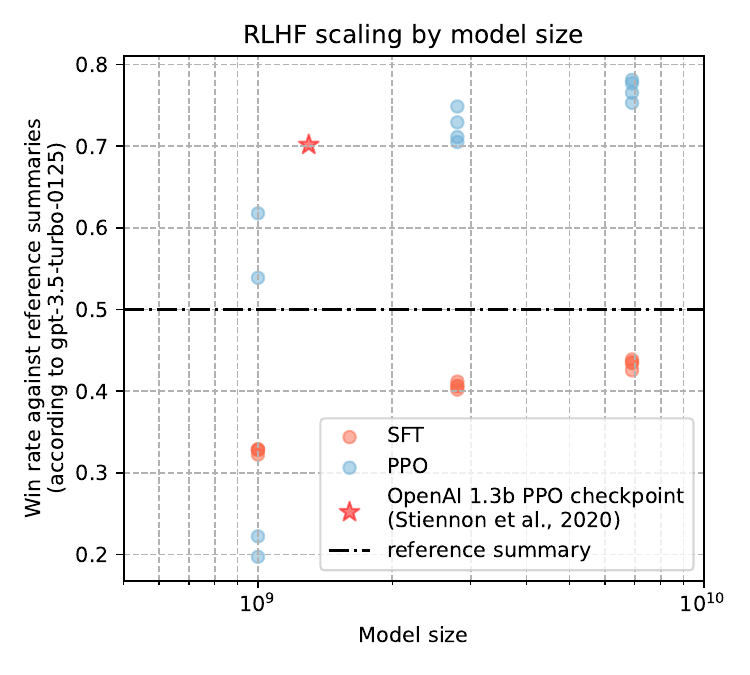}
    \caption{The win rate of our models’ summaries over the human-generated reference summaries on the \emph{validation split} of the TL;DR dataset, judged by GPT 3.5. Our SFT / RM / PPO models were trained with four random seeds across the 1B, 2.8B, and 6.9B Pythia~\cite{biderman2023emergent} model sizes using the same \texttt{3e-6} learning rate.
     }
    \label{fig:rlhf-scaling}
\end{figure}

\section{Introduction}
There has been tremendous development in pre-trained large language models (LLMs) over the years~\citep{radford2018improving,radford2019language,brown2020language,rae2021scaling}. Given the previous tokens, these LLMs are trained to predict the next token accurately, and they can be prompted to solve a wide range of natural language processing (NLP) tasks. However, the next-token-prediction objective differs from the fundamental objective of ``outputting contents that humans prefer''. To address this gap, Reinforcement Learning from Human Feedback (RLHF)~\citep{stiennon2020learning,ouyang2022training,bai2022training} is introduced as a pipeline to collect pair-wise human preferences, train a reward model (RM) to model these preferences and use Reinforcement Learning (RL) to create a model that outputs contents that humans prefer.





It has proven challenging to reproduce OpenAI's RLHF pipeline~\citep{ouyang2022training,openai2024gpt4} in the open-source community for several reasons: 1) RL and RLHF have many subtle implementation details that can significantly impact training stability~\citep{Engstrom2020Implementation,shengyi2022the37implementation,huang2024thenimplementation}, 2) the models are challenging to evaluate for the instruction following tasks (e.g., evaluating the quality of 800 lines of generated code snippet for a coding task), 3) they take a long time to train and iterate. 

This work addresses the aforementioned three challenges by taking a step back and reproducing OpenAI's earlier but seminal RLHF work in TL;DR summarization~\citep{stiennon2020learning}. TL;DR is one of the most popular benchmarks for RLHF methods alongside instruction following tasks such as Anthropic's HH-RLHF \citep{bai2022training} and AlpacaFarm \citep{dubois_alpacafarm_2023}. But summarization tasks are much easier to evaluate than general instruction following tasks because summaries are typically short and bad summaries usually have bad accuracy, coverage, or make-up facts. The reduced context and generation length also mean more efficient training, allowing us to iterate quickly and polish a working RLHF pipeline. Specifically, our contributions are as follows:



\textbf{We reproduced the RLHF scaling behaviors in \citet{stiennon2020learning}.} Our end-to-end pipeline demonstrates that larger models lead to improved ROUGE scores for SFT models, higher validation accuracy for RMs, and higher win rates of the generated summaries over reference summaries for the final RL policies.


\textbf{Our highly reproducible RLHF pipeline uses a single learning rate -- no LR sweeps.} To simplify the setup and improve reproducibility, we use the \emph{same learning rate} for SFT, RM, and PPO training. In contrast, the original setup ran hyperparameter sweeps separately for SFT, RM, and PPO model training. To ensure researchers can reliably reproduce our work, we ran our model training for four random seeds, including failure cases for analysis. 

\textbf{We enumerate over 20 relevant implementation details and offer detailed insights.} This paper delves into the details of the TL;DR datasets, including their specifications, tokenization processes, and token length distributions. We then detail the training setups, implementation details, and results for both the Supervised Fine-Tuning (SFT) and Reward Model (RM) components. Additionally, we explore the details of PPO implementation and how they impact performance. We provide visualizations to compare the behavior of aligned models versus base models.

\textbf{Our work is fully open source and transparent.} We make our complete source code available at \url{https://github.com/vwxyzjn/summarize_from_feedback_details}. We also release model checkpoints and training metrics in Appendix~\ref{sec:model-checkpoints}.

\section{Preliminaries}

In tasks for which it is difficult to design a reward function, RLHF is a technique that trains a reward model from human preferences and then performs RL training against the reward model~\citep{christiano2017deep}. At a larger scale, RLHF has been used to fine-tune large language models (LLMs) to output contents that align more with human preferences~\citep{ziegler2019fine,stiennon2020learning,ouyang2022training,bai2022training,openai2024gpt4,team2023gemini}. RLHF typically has three steps, as shown below.

\textbf{Step 1: Train an SFT policy:} The pre-trained LLMs are fine-tuned on the set of human demonstrations using the next-token prediction loss. In this reproduction work, these human demonstrations come from the human summaries of Reddit posts from a filtered TL;DR dataset~\citep {stiennon2020learning}. In later work, the human demonstrations could come from paid contracted labelers~\citep {ouyang2022training} on a larger variety of tasks.

\textbf{Step 2: Collect preference pairs and train an RM:} Various policies, such as the trained SFT policy, are then used to sample completions, and the human labelers would indicate which completions they prefer. Given the preference dataset, we initialize an RM from the SFT policy by adding a randomly initialized linear head that outputs a scalar score. The RM is trained to predict the log probability that a completion would be preferred by the labelers. Specifically, the RM loss is
\begin{equation}
\mathcal{L}_R(r_{\phi}) = -\mathbb{E}_{(x, y_c, y_r)\sim \mathcal{D}_\text{PREF}}\bigl[\log \sigma(r_{\phi}(x, y_c)- r_{\phi}(x, y_r))\bigr],
\end{equation}
where $\sigma(x) = \frac{1}{1 + e^{-x}}$ is the sigmoid function, $\mathcal{D}_\text{PREF}$ the human preference dataset, $x$ the prompt to the model (in this case, the Reddit post), $y_c$ the chosen/preferred completion by a labeler, $y_r$ the rejected completion by the labeler, $\phi$ are the parameters of the RM $r$. When plugging in the $\sigma$ function, we get the same form \citet{bai2022training} use:
\begin{equation}
\label{eq:rm-alternative-form}
    \mathcal{L}_R(r_{\phi}) = \mathbb{E}_{(x, y_c, y_r)\sim \mathcal{D}_\text{PREF}}\bigl[\log ({1 + e^{r_{\phi}(x, y_r) - r_{\phi}(x, y_c)}})\bigr]
\end{equation}

\textbf{Step 3: Train an RL policy against the RM:} Initializing from the SFT policy, the RL policy then samples completions given prompts and has the RM produce a score based on these completions. The reward of the RL policy then includes this score and a KL penalty to ensure the RL policy does not deviate too much from the SFT policy. Specifically, the reward of the RL problem is 
\begin{equation}\label{eq:RL}
R(x, y) = \left(r_{\phi}(x, y) - \beta\mathbb{D}_{\textrm{KL}}\bigl[\pi_{\theta}(y\mid x)\mid \mid \pisft(y\mid x)\bigr]\right)
\end{equation}
where $\beta$ is a parameter controlling the strength of the KL penalty, $\theta$ the parameters of RL policy $\pi_{\theta}$. Then, PPO is used to maximize the RLHF objective  $\max_{\pi_{\theta}}  \mathbb{E}_{x\sim \mathcal{D}_\text{SFT}, y\sim \pi_{\theta}(y \mid x)} R(x, y)$, where $\mathcal{D}_\text{SFT}$ is the prompts in the SFT dataset. 

\textbf{RL-free approaches:}  The RLHF + PPO pipeline can be quite computationally expensive because 1) the training program typically needs to load 3-4 models into the GPU memory and 2) RL policy training needs online generations and running the RM. To alleviate these two problems, researchers have proposed RL-free approaches~\citep{rafailov2023direct,azar2023general,hong2024reference}. One of the most widely-used RL-free approaches is Direct preference optimization (DPO), which has the following loss:
\begin{equation}\label{eq:optimum_model}
    \mathcal{L}_\text{DPO}(\pi_{\theta}) = -\mathbb{E}_{(x, y_c, y_r)\sim \mathcal{D}_\text{PREF}}\left[\log \sigma \left(\beta \log \frac{\pi_{\theta}(y_c\mid x)}{\pisft(y_c\mid x)} - \beta \log \frac{\pi_{\theta}(y_r\mid x)}{\pisft(y_r)}\right)\right].
\end{equation}
Note that DPO implicitly does the reward modeling: we can extract the reward score using the following formula:
\begin{equation}\label{eq:dpo-implicit-rm}
    r(x, y) = \beta \log\frac{\pi_\theta(y \mid x)}{\pisft(y \mid x)}.
\end{equation}
DPO is a more accessible alignment technique that has been implemented in popular RLHF libraries such as TRL~\cite {vonwerra2022trl}. DPO has also been used to align larger models effectively (e.g., Zephyr 7B~\citep{tunstall2023zephyr}, Tulu 70B~\citep{ivison2023camels}, and Mixtral 8x7B~\citep{jiang2024mixtral}).

\section{Dataset Details}
We start with a solid understanding of the dataset, the tokenization process, and the token lengths. This section provides an in-depth analysis and visualization of the TL;DR datasets from \citet{stiennon2020learning}, which includes an SFT dataset~\footnote{\url{https://huggingface.co/datasets/vwxyzjn/summarize_from_feedback_tldr_3_filtered}} and a preference dataset~\footnote{\url{https://huggingface.co/datasets/openai/summarize_from_feedback}}. 

\begin{dlist}[resume] 
  \item \label{detail:ds-specification}{\textbf{Dataset -> Specification}}
\end{dlist}

The SFT dataset is fairly intuitive -- it contains the subreddit, title, post, and reference summary columns. On the other hand, the preference dataset is a lot more nuanced. 

The \texttt{train} split of the preference dataset contains the subreddit, title, and post columns; it also contains two sampled summaries, their sampling policies, an internal batch number, the belonging split, which summary the human rater prefers, and optionally, a note or confidence level. 

The \texttt{validation} split of the preference dataset contains the same information as above, and \emph{definitely} includes a confidence level. Furthermore, the \texttt{validation} split contains small batches of data for CNN/DM news articles.

\begin{table}[t]
\centering
\caption{Query pre-processing example. The left example has 512, which is greater than the max query token length of 512, so the pre-processing truncates the last paragraph of the post. Colors show how the contents are tokenized.}
\label{tab:preprocessing}
\begin{tabular}{p{7cm}|p{7cm}}
\toprule
Before: 519 tokens & After: 449 tokens \\
\midrule
\sethlcolor{LightOrchid}\hl{S}\sethlcolor{LightYellowGreen}\hl{UB}\sethlcolor{LightYellowOrange}\hl{RED}\sethlcolor{LightSalmon}\hl{DIT}\sethlcolor{LightAquamarine}\hl{:}\sethlcolor{LightOrchid}\hl{ r}\sethlcolor{LightYellowGreen}\hl{/}\sethlcolor{LightYellowOrange}\hl{tif}\sethlcolor{LightSalmon}\hl{u}

\sethlcolor{LightYellowGreen}\hl{TIT}\sethlcolor{LightYellowOrange}\hl{LE}\sethlcolor{LightSalmon}\hl{:}\sethlcolor{LightAquamarine}\hl{ T}\sethlcolor{LightOrchid}\hl{IF}\sethlcolor{LightYellowGreen}\hl{U}\sethlcolor{LightYellowOrange}\hl{:}\sethlcolor{LightSalmon}\hl{ by}\sethlcolor{LightAquamarine}\hl{ ruining}\sethlcolor{LightOrchid}\hl{ my}\sethlcolor{LightYellowGreen}\hl{ chance}\sethlcolor{LightYellowOrange}\hl{ at}\sethlcolor{LightSalmon}\hl{ losing}\sethlcolor{LightAquamarine}\hl{ my}\sethlcolor{LightOrchid}\hl{ virginity}

\sethlcolor{LightSalmon}\hl{POST}\sethlcolor{LightAquamarine}\hl{:}\sethlcolor{LightOrchid}\hl{ I}\sethlcolor{LightYellowGreen}\hl{'ll}\sethlcolor{LightYellowOrange}\hl{ never}\sethlcolor{LightSalmon}\hl{ forget}\sethlcolor{LightAquamarine}\hl{ this}\sethlcolor{LightOrchid}\hl{ moment}\sethlcolor{LightYellowGreen}\hl{...}

 & \sethlcolor{LightOrchid}\hl{S}\sethlcolor{LightYellowGreen}\hl{UB}\sethlcolor{LightYellowOrange}\hl{RED}\sethlcolor{LightSalmon}\hl{DIT}\sethlcolor{LightAquamarine}\hl{:}\sethlcolor{LightOrchid}\hl{ r}\sethlcolor{LightYellowGreen}\hl{/}\sethlcolor{LightYellowOrange}\hl{tif}\sethlcolor{LightSalmon}\hl{u}

\sethlcolor{LightYellowGreen}\hl{TIT}\sethlcolor{LightYellowOrange}\hl{LE}\sethlcolor{LightSalmon}\hl{:}\sethlcolor{LightAquamarine}\hl{ T}\sethlcolor{LightOrchid}\hl{IF}\sethlcolor{LightYellowGreen}\hl{U}\sethlcolor{LightYellowOrange}\hl{:}\sethlcolor{LightSalmon}\hl{ by}\sethlcolor{LightAquamarine}\hl{ ruining}\sethlcolor{LightOrchid}\hl{ my}\sethlcolor{LightYellowGreen}\hl{ chance}\sethlcolor{LightYellowOrange}\hl{ at}\sethlcolor{LightSalmon}\hl{ losing}\sethlcolor{LightAquamarine}\hl{ my}\sethlcolor{LightOrchid}\hl{ virginity}

\sethlcolor{LightSalmon}\hl{POST}\sethlcolor{LightAquamarine}\hl{:}\sethlcolor{LightOrchid}\hl{ I}\sethlcolor{LightYellowGreen}\hl{'ll}\sethlcolor{LightYellowOrange}\hl{ never}\sethlcolor{LightSalmon}\hl{ forget}\sethlcolor{LightAquamarine}\hl{ this}\sethlcolor{LightOrchid}\hl{ moment}\sethlcolor{LightYellowGreen}\hl{...}

\\
... & ...\\
\sethlcolor{LightYellowOrange}\hl{ But}\sethlcolor{LightSalmon}\hl{ for}\sethlcolor{LightAquamarine}\hl{ some}\sethlcolor{LightOrchid}\hl{ reason}\sethlcolor{LightYellowGreen}\hl{,}\sethlcolor{LightYellowOrange}\hl{ the}\sethlcolor{LightSalmon}\hl{ combination}\sethlcolor{LightAquamarine}\hl{ of}\sethlcolor{LightOrchid}\hl{ my}\sethlcolor{LightYellowGreen}\hl{ emotions}\sethlcolor{LightYellowOrange}\hl{,}\sethlcolor{LightSalmon}\hl{ inexper}\sethlcolor{LightAquamarine}\hl{ience}\sethlcolor{LightOrchid}\hl{,}\sethlcolor{LightYellowGreen}\hl{ and}\sethlcolor{LightYellowOrange}\hl{ shock}\sethlcolor{LightSalmon}\hl{ produced}\sethlcolor{LightAquamarine}\hl{ the}\sethlcolor{LightOrchid}\hl{ worst}\sethlcolor{LightYellowGreen}\hl{ possible}\sethlcolor{LightYellowOrange}\hl{ words}\sethlcolor{LightSalmon}\hl{ to}\sethlcolor{LightAquamarine}\hl{ come}\sethlcolor{LightOrchid}\hl{ out}\sethlcolor{LightYellowGreen}\hl{,}\sethlcolor{LightYellowOrange}\hl{ "}\sethlcolor{LightSalmon}\hl{I}\sethlcolor{LightAquamarine}\hl{ love}\sethlcolor{LightOrchid}\hl{ you}\sethlcolor{LightYellowGreen}\hl{".}

\sethlcolor{LightAquamarine}\hl{She}\sethlcolor{LightOrchid}\hl{ got}\sethlcolor{LightYellowGreen}\hl{ up}\sethlcolor{LightYellowOrange}\hl{,}\sethlcolor{LightSalmon}\hl{ put}\sethlcolor{LightAquamarine}\hl{ her}\sethlcolor{LightOrchid}\hl{ clothes}\sethlcolor{LightYellowGreen}\hl{ on}\sethlcolor{LightYellowOrange}\hl{,}\sethlcolor{LightSalmon}\hl{ didn}\sethlcolor{LightAquamarine}\hl{'t}\sethlcolor{LightOrchid}\hl{ say}\sethlcolor{LightYellowGreen}\hl{ a}\sethlcolor{LightYellowOrange}\hl{ thing}\sethlcolor{LightSalmon}\hl{ and}\sethlcolor{LightAquamarine}\hl{ walked}\sethlcolor{LightOrchid}\hl{ out}\sethlcolor{LightYellowGreen}\hl{ leaving}\sethlcolor{LightYellowOrange}\hl{ me}\sethlcolor{LightSalmon}\hl{ on}\sethlcolor{LightAquamarine}\hl{ my}\sethlcolor{LightOrchid}\hl{ couch}\sethlcolor{LightYellowGreen}\hl{ with}\sethlcolor{LightYellowOrange}\hl{ a}\sethlcolor{LightSalmon}\hl{ bon}\sethlcolor{LightAquamarine}\hl{er}\sethlcolor{LightOrchid}\hl{.}\sethlcolor{LightYellowGreen}\hl{ The}\sethlcolor{LightYellowOrange}\hl{ best}\sethlcolor{LightSalmon}\hl{ moment}\sethlcolor{LightAquamarine}\hl{ of}\sethlcolor{LightOrchid}\hl{ my}\sethlcolor{LightYellowGreen}\hl{ life}\sethlcolor{LightYellowOrange}\hl{ had}\sethlcolor{LightSalmon}\hl{ just}\sethlcolor{LightAquamarine}\hl{ turned}\sethlcolor{LightOrchid}\hl{ into}\sethlcolor{LightYellowGreen}\hl{ my}\sethlcolor{LightYellowOrange}\hl{ worst}\sethlcolor{LightSalmon}\hl{.}\sethlcolor{LightAquamarine}\hl{ Shortly}\sethlcolor{LightOrchid}\hl{ after}\sethlcolor{LightYellowGreen}\hl{ my}\sethlcolor{LightYellowOrange}\hl{ drunk}\sethlcolor{LightSalmon}\hl{ roomm}\sethlcolor{LightAquamarine}\hl{ates}\sethlcolor{LightOrchid}\hl{ stumbled}\sethlcolor{LightYellowGreen}\hl{ in}\sethlcolor{LightYellowOrange}\hl{ and}\sethlcolor{LightSalmon}\hl{ I}\sethlcolor{LightAquamarine}\hl{ had}\sethlcolor{LightOrchid}\hl{ to}\sethlcolor{LightYellowGreen}\hl{ report}\sethlcolor{LightYellowOrange}\hl{ of}\sethlcolor{LightSalmon}\hl{ my}\sethlcolor{LightAquamarine}\hl{ failures}\sethlcolor{LightOrchid}\hl{.}\sethlcolor{LightYellowGreen}\hl{ My}\sethlcolor{LightYellowOrange}\hl{ only}\sethlcolor{LightSalmon}\hl{ consolation}\sethlcolor{LightAquamarine}\hl{ was}\sethlcolor{LightOrchid}\hl{ the}\sethlcolor{LightYellowGreen}\hl{ smell}\sethlcolor{LightYellowOrange}\hl{ on}\sethlcolor{LightSalmon}\hl{ my}\sethlcolor{LightAquamarine}\hl{ hands}\sethlcolor{LightOrchid}\hl{.}

\sethlcolor{LightSalmon}\hl{TL}\sethlcolor{LightAquamarine}\hl{;}\sethlcolor{LightOrchid}\hl{DR}\sethlcolor{LightYellowGreen}\hl{:} 

& 

\sethlcolor{LightYellowOrange}\hl{ But}\sethlcolor{LightSalmon}\hl{ for}\sethlcolor{LightAquamarine}\hl{ some}\sethlcolor{LightOrchid}\hl{ reason}\sethlcolor{LightYellowGreen}\hl{,}\sethlcolor{LightYellowOrange}\hl{ the}\sethlcolor{LightSalmon}\hl{ combination}\sethlcolor{LightAquamarine}\hl{ of}\sethlcolor{LightOrchid}\hl{ my}\sethlcolor{LightYellowGreen}\hl{ emotions}\sethlcolor{LightYellowOrange}\hl{,}\sethlcolor{LightSalmon}\hl{ inexper}\sethlcolor{LightAquamarine}\hl{ience}\sethlcolor{LightOrchid}\hl{,}\sethlcolor{LightYellowGreen}\hl{ and}\sethlcolor{LightYellowOrange}\hl{ shock}\sethlcolor{LightSalmon}\hl{ produced}\sethlcolor{LightAquamarine}\hl{ the}\sethlcolor{LightOrchid}\hl{ worst}\sethlcolor{LightYellowGreen}\hl{ possible}\sethlcolor{LightYellowOrange}\hl{ words}\sethlcolor{LightSalmon}\hl{ to}\sethlcolor{LightAquamarine}\hl{ come}\sethlcolor{LightOrchid}\hl{ out}\sethlcolor{LightYellowGreen}\hl{,}\sethlcolor{LightYellowOrange}\hl{ "}\sethlcolor{LightSalmon}\hl{I}\sethlcolor{LightAquamarine}\hl{ love}\sethlcolor{LightOrchid}\hl{ you}\sethlcolor{LightYellowGreen}\hl{".}

\sethlcolor{LightAquamarine}\hl{TL}\sethlcolor{LightOrchid}\hl{;}\sethlcolor{LightYellowGreen}\hl{DR}\sethlcolor{LightYellowOrange}\hl{:} \\ 
\bottomrule
\end{tabular}
\end{table}

\begin{dlist}[resume] 
  \item \label{detail:ds-specification}{\textbf{Dataset -> Do not truncate the sentence, truncate the paragraph}}
\end{dlist}
The next step is to tokenize the query. The query token goes through the following two transformations (\href{https://github.com/openai/summarize-from-feedback/blob/700967448d10004279f138666442bf1497d0e705/summarize_from_feedback/utils/experiment_helpers.py\#L196-L199}{\texttt{utils/experiment\_helpers.py\#L196-L199}}, \href{https://github.com/openai/summarize-from-feedback/blob/700967448d10004279f138666442bf1497d0e705/summarize_from_feedback/tasks.py#L98-L165}{\texttt{tasks.py\#L98-L165}})

\begin{enumerate}
    \item \textbf{Format the query} input string using the following template.
    \begin{itemize}
        \item \texttt{SUBREDDIT: r/\{subreddit\}\textbackslash{}n\textbackslash{}nTITLE: \{title\}\textbackslash{}n\textbackslash{}nPOST: \{post\}\textbackslash{}n\textbackslash{}nTL;DR:}
    \end{itemize}
    \item \textbf{Clever truncation} to ensure the query token length is not greater than 512.
    \begin{itemize}
        \item The formatted query is tokenized using the tokenizer. If the query token length is not greater than 512, it is padded from the left with either padding tokens or repeated white spaces.
        \item If the query token length exceeds 512, the pre-processing process will attempt to remove the last paragraph. Specifically, it finds the last index of \texttt{\textbackslash n} in the post and removes the content after. Table~\ref{tab:preprocessing} shows an example. This is a much more sophisticated form of truncation compared to a hard truncation on a maximum token length.
    \end{itemize}
    \item \textbf{No trailing space after} ``\sethlcolor{LightSalmon}\hl{TL}\sethlcolor{LightAquamarine}\hl{;}\sethlcolor{LightOrchid}\hl{DR}\sethlcolor{LightYellowGreen}\hl{:}'' to make sure there is no weird generation issues due to the nature of tokenization.
\end{enumerate}

\begin{dlist}[resume] 
  \item \label{detail:use-an-eos}{\textbf{Dataset -> Prepend a leading space to completion; append an EOS token to the completions; use a special padding token \texttt{[PAD]}; do not use EOS token synonymously as \texttt{[PAD]}}}
\end{dlist}
When tokenizing the concatenation of queries and responses for the SFT and preference dataset, we always do the following:
\begin{enumerate}
    \item Prepends a leading space to the completion, so there is always a space between \sethlcolor{LightSalmon}\hl{TL}\sethlcolor{LightAquamarine}\hl{;}\sethlcolor{LightOrchid}\hl{DR}\sethlcolor{LightYellowGreen}\hl{:}  and the completion such as below.
    \item Append an EOS \texttt{<|endoftext|>} token to the completion.
    \item When needed to pad the sequence to a maximum length, we always use a special padding token \texttt{[PAD]}.
\end{enumerate}

For example,  we would add the EOS token and \texttt{[PAD]} token to the reference summary as follows:

\sethlcolor{LightOrchid}\hl{\texttt{ long}}\sethlcolor{LightYellowGreen}\hl{\texttt{ relationship}}\sethlcolor{LightYellowOrange}\hl{\texttt{;}}\sethlcolor{LightSalmon}\hl{\texttt{ fell}}\sethlcolor{LightAquamarine}\hl{\texttt{ in}}\sethlcolor{LightOrchid}\hl{\texttt{ love}}\sethlcolor{LightYellowGreen}\hl{\texttt{ with}}\sethlcolor{LightYellowOrange}\hl{\texttt{ another}}\sethlcolor{LightSalmon}\hl{\texttt{ person}}\sethlcolor{LightAquamarine}\hl{\texttt{;}}\sethlcolor{LightOrchid}\hl{\texttt{ admitted}}\sethlcolor{LightYellowGreen}\hl{\texttt{ it}}\sethlcolor{LightYellowOrange}\hl{\texttt{;}}\sethlcolor{LightSalmon}\hl{\texttt{ would}}\sethlcolor{LightAquamarine}\hl{\texttt{ like}}\sethlcolor{LightOrchid}\hl{\texttt{ it}}\sethlcolor{LightYellowGreen}\hl{\texttt{ to}}\sethlcolor{LightYellowOrange}\hl{\texttt{ disappear}}\sethlcolor{LightSalmon}\hl{\texttt{,}}\sethlcolor{LightAquamarine}\hl{\texttt{ though}}\sethlcolor{LightOrchid}\hl{\texttt{ it}}\sethlcolor{LightYellowGreen}\hl{\texttt{ doesn}}\sethlcolor{LightYellowOrange}\hl{\texttt{'t}}\sethlcolor{LightSalmon}\hl{\texttt{.}}\sethlcolor{LightAquamarine}\hl{\texttt{<|endoftext|>}}\sethlcolor{LightOrchid}\hl{\texttt{[PAD]}}\sethlcolor{LightYellowGreen}\hl{\texttt{[PAD]}}\sethlcolor{LightYellowOrange}\hl{\texttt{[PAD]}}\sethlcolor{LightSalmon}...

We do \emph{not} recommend using the common practice which uses the EOS token synonymously with the \texttt{[PAD]} token (e.g., \texttt{tokenizer.pad\_token\_id = tokenizer.eos\_token\_id}). This is because the EOS token would then be masked out as a padding token during SFT training, and the model would not learn to end a summary -- a trained model would often continue to sample summary texts without stopping. This could exacerbate existing issues with RLHF models generating longer outputs \citep{stiennon2020learning, dubois_alpacafarm_2023}. With a clear EOS token and padding token distinction, our final trained endpoint always learns to end summaries with the EOS token, as shown in Figure~\ref{tab:ai2-visual}.

While \citet{stiennon2020learning} choose \texttt{<|endoftext|>} as the EOS token, it may be possible to use another token like \texttt{<|im\_end|>}\footnote{\url{https://github.com/openai/openai-python/blob/release-v0.28.0/chatml.md}} instead as the EOS token. We suspect the key practice is to end the completion with some special token, so the model can learn when to stop.

\begin{figure}[t]
    \centering
    \includegraphics[width=0.90\linewidth]{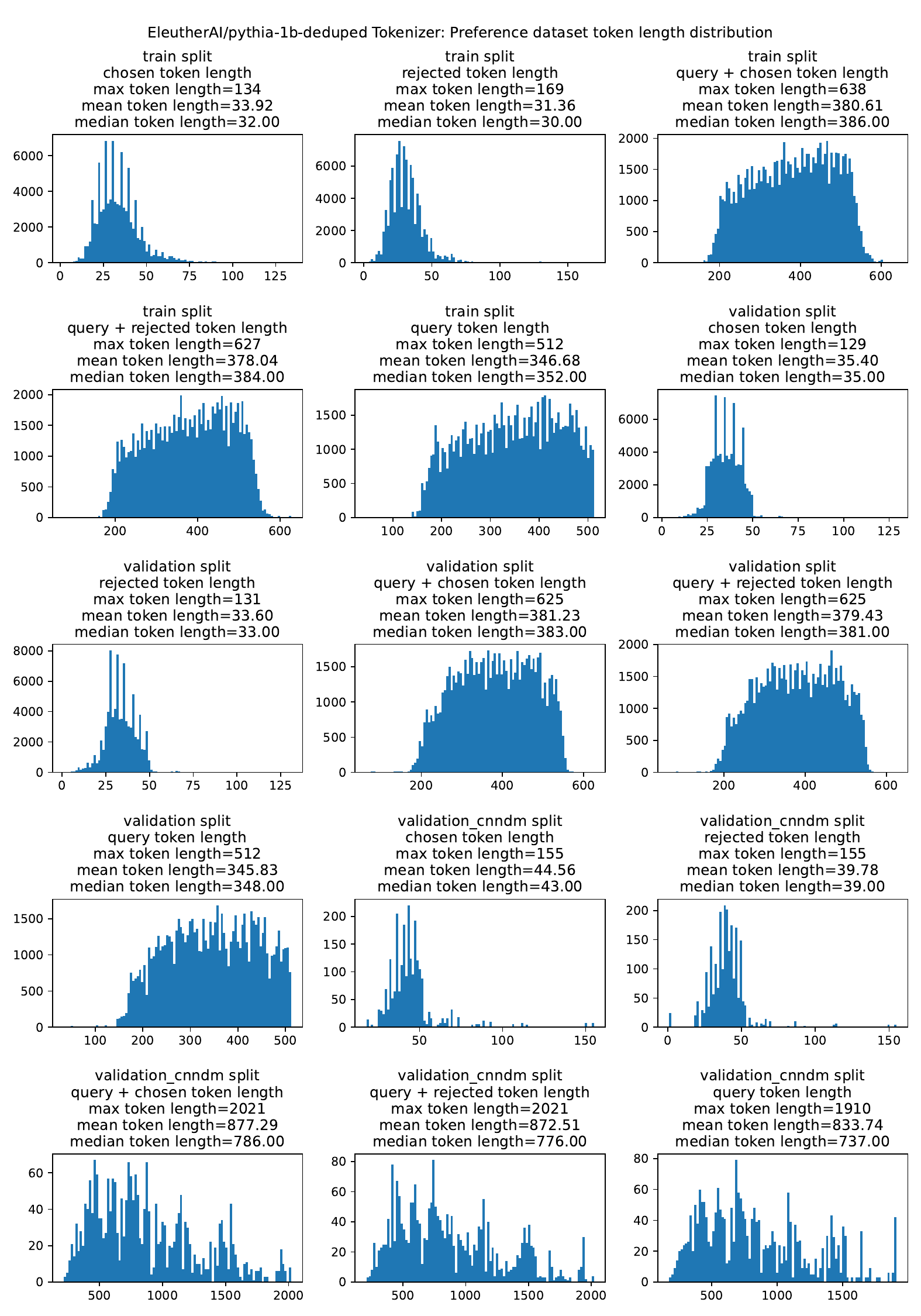}
    \caption{The token length visualization of the preference dataset.
     }
    \label{fig:token-length-vis-pref}
\end{figure}

\begin{figure}[t]
    \centering
    \includegraphics[width=0.90\linewidth]{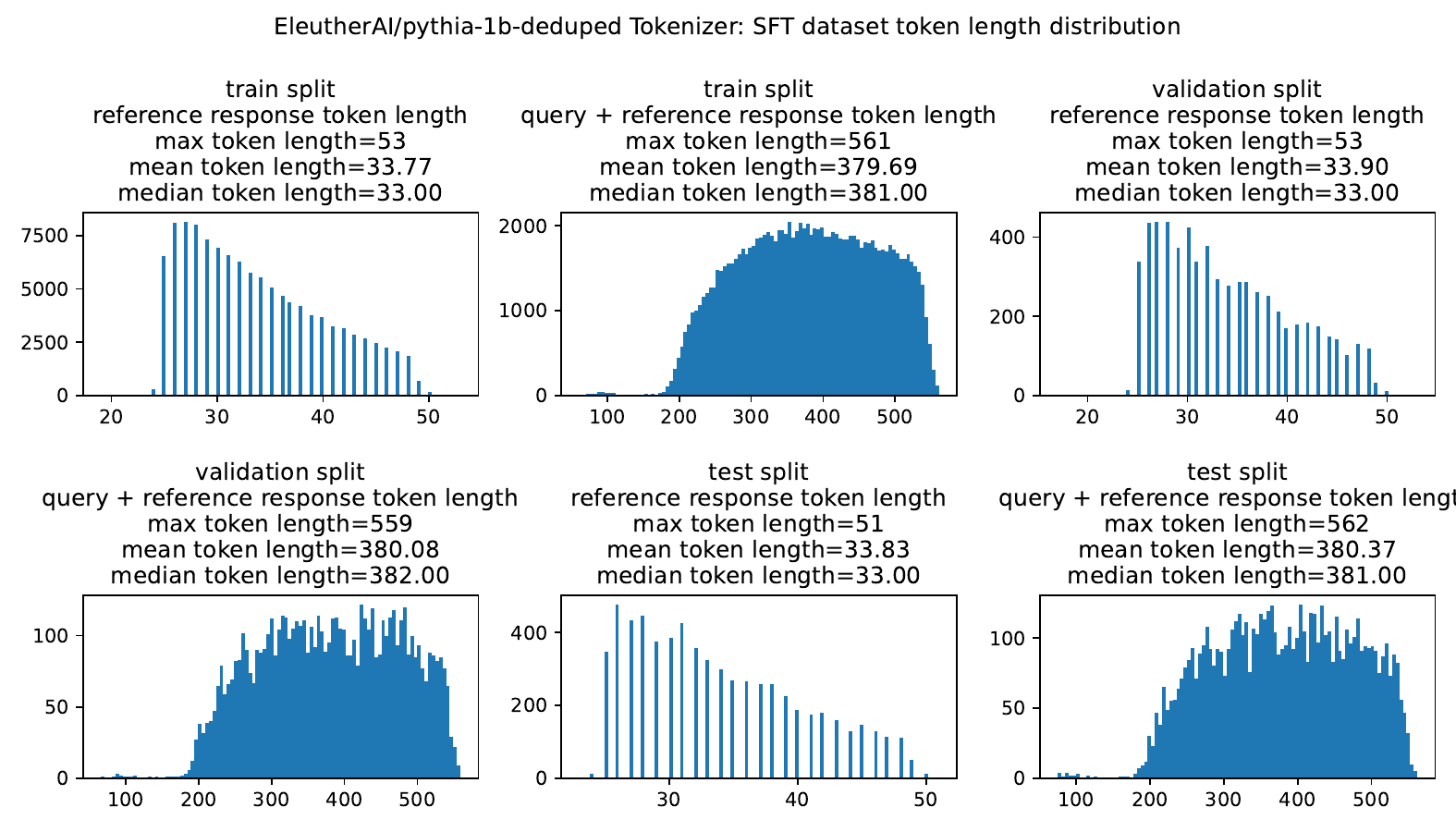}
    \caption{The token length visualization of the preference dataset.
     }
    \label{fig:token-length-vis-sft}
\end{figure}

\begin{dlist}[resume] 
  \item \label{detail:ds-token-len}{\textbf{Dataset -> SFT and preference datasets have different tokenization length}}
\end{dlist}
The SFT dataset had already been filtered such that all the reference summary lengths were controlled -- they have a maximum of 48 tokens using the GPT2 tokenizer. In our case, we used Pythia's tokenizer~\citep{biderman2023emergent}, with which the reference summaries have a maximum of 53 tokens. However, an interesting fact is that the summary lengths in the preference dataset are \emph{not} controlled to be the same. Figures~\ref{fig:token-length-vis-pref} and~\ref{fig:token-length-vis-sft} show the length distribution. Several observations:

\begin{enumerate}
    \item The chosen/rejected response token length in the preference dataset can be as long as 169, significantly exceeding the 53 tokens found in the SFT dataset.
    \item The median chosen response token length is 32, which is slightly longer than that of the rejected response token of 30. 
\end{enumerate}

\begin{dlist}[resume] 
  \item \label{detail:ds-specification}{\textbf{Dataset -> Pre-tokenize the dataset: right pad the concatenation of queries and responses; left pad the queries}}
\end{dlist}
To pre-tokenize the dataset for training, we right pad the concatenation of queries and responses and left pad the queries, as shown below.
\begin{enumerate}
    \item \textbf{SFT dataset for SFT training}: we concatenate the query and the reference response together and pad from the right, so during training each sampled batch will have the shape \texttt{(B, 562)}.
    \item \textbf{Preference dataset for RM training}: we concatenate the query-chosen and query-rejected responses together and pad from the right, so during the RM training, each sampled batch will have the shape \texttt{(B, 638)}.
    \item \textbf{Preference dataset for RM evaluation}: During RM evaluation, the sampled batch in the TL;DR splits will have shape \texttt{(B, 638)}. Note that in the preference dataset, there is also a split that measures the RM's generalization ability to the CNN/DM dataset, and this split has a much longer token length; in particular, a sampled batch from this dataset will have shape \texttt{(B, 2021)}.
    \item \textbf{SFT dataset for PPO training}: we pad the query from the left to make generations compatible with \texttt{transformers} (since decoder models require left padding for generations), so each sampled batch will have shape \texttt{(B, 512)}.
\end{enumerate}



\begin{dlist}[resume] 
  \item \label{detail:ds-divsers-validation}{\textbf{Dataset -> The validation split of the preference dataset has a lot of OOD data.}}
\end{dlist}

As illustrated in Table~\ref{table:split-pairs} (see Appendix~\ref{sec:details-pref-paris} for details on the exact policy comparisons and their counts), the sampling policies employed in the preference dataset exhibit significant diversity, which is out of the distribution of the sampling policies used in the train split. As a result, the validation set serves as a great measure of the generalization ability of the (RM).

\begin{table}[t]
\centering
\caption{The number of unique pairs of policies compared differs in each preference dataset split. In particular, notice the validation set contains highly diverse pairs (see Appendix~\ref{sec:details-pref-paris} for details on the exact policy comparisons and their counts).}
\label{table:split-pairs}
\begin{tabular}{lc}
\toprule
\textbf{Split name}         & \textbf{The number of unique pairs of policies compared} \\ \midrule
train              & 47    \\
validation         & 241   \\
    validation\_cnndm & 7     \\ \bottomrule
\end{tabular}
\end{table}


\section{General Details}



\begin{dlist}[resume] 
  \item \label{detail:ds-specification}{\textbf{Model -> Disable dropout to ensure PPO's ratio calculation still works}}
\end{dlist}

We disable the dropout layers during training, similar to the settings in \citet{ziegler2019fine,huang2024thenimplementation}. This is important for PPO training, especially because with dropout activated, the log probabilities of tokens will not be reproducible, making calculating the KL penalty unreliable while also causing the ratios of the PPO to be not 1s during the first epoch, causing PPO optimization problems. For consistency, we also disable dropout for SFT and RM training.

\begin{dlist}[resume] 
  \item \label{detail:ds-specification}{\textbf{Setup -> Tech stack}}
\end{dlist}
We used the \texttt{transformers}~\citep{Wolf_Transformers_State-of-the-Art_Natural_2020} library's implementation of the Pythia models in conjunction with \texttt{deepspeed}'s ZeRO Stage 2~\citep{rasley2020deepspeed,rajbhandari2020zero} to help fit the models into the GPU memory; for 6.9B PPO training we also offload the reference policy and reward model to CPU. We launch experiments using \texttt{accelerate}~\citep{accelerate} with \texttt{bf16} mixed-precision training and track them with Weights and Biases~\citep{wandb}. We use 8xH100 machines and always upload the trained models to Hugging Face's model hub\footnote{\url{https://huggingface.co/models}}.

\begin{figure}[t]
    \centering
    \includegraphics[width=0.49\linewidth]{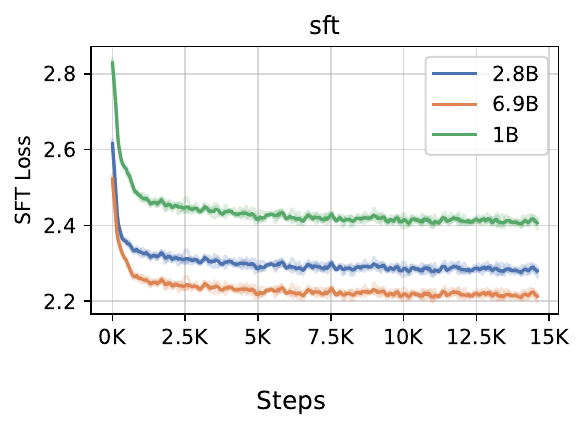}
    \includegraphics[width=0.49\linewidth]{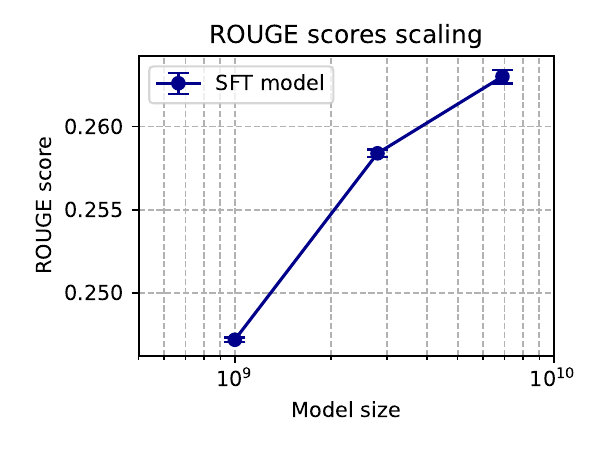}
    \caption{Left is the SFT training losses across the model sizes for one epoch of the train split of the SFT dataset (116,722 episodes). Right is the scaling behaviors of the ROUGE score between the trained SFT model summaries and the reference summaries}
    \label{fig:sft}
\end{figure}

\section{SFT Details}

\begin{dlist}[resume] 
  \item \label{detail:ds-specification}{\textbf{SFT Training -> Setups}}
\end{dlist}

Our SFT setup closely follows \citet{stiennon2020learning}, except for a modified learning rate (Table~\ref{tab:sft-hyper}).

\begin{table}[h]
\centering
\caption{SFT hyperparameters}
\begin{tabular}{lc}
\toprule
\textbf{Hyperparameter} & \textbf{Default Value} \\
\midrule
Number of Train Epochs & 1 (or 116,722 episodes) \\
Optimizer & AdamW ($\epsilon=1e-5$, $\texttt{lr}=3e-6$)  \\
Scheduler & Cosine  \\
Batch Size & 128 \\
\bottomrule
\end{tabular}
\label{tab:sft-hyper}
\end{table}

\subsection{SFT training results}

The SFT loss curves can be found in Figure~\ref{fig:sft}. Unsurprisingly, larger models have smaller next-token-prediction losses. After finishing the training, we also evaluated the ROUGE scores against the reference summaries in the validation set. We find a favorable scaling behavior, similar to Figure 14 (a) in \citet{stiennon2020learning}.

\begin{figure}[t]
    \centering
    \includegraphics[width=0.98\linewidth]{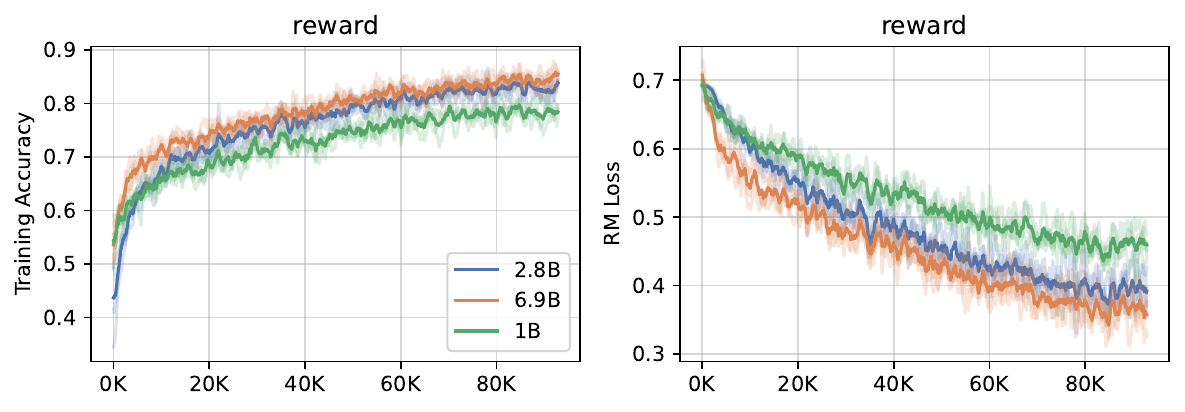}
    \includegraphics[width=0.49\linewidth]{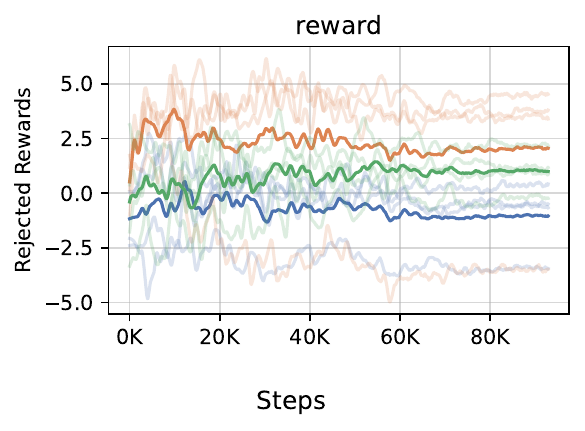}
    \includegraphics[width=0.49\linewidth]{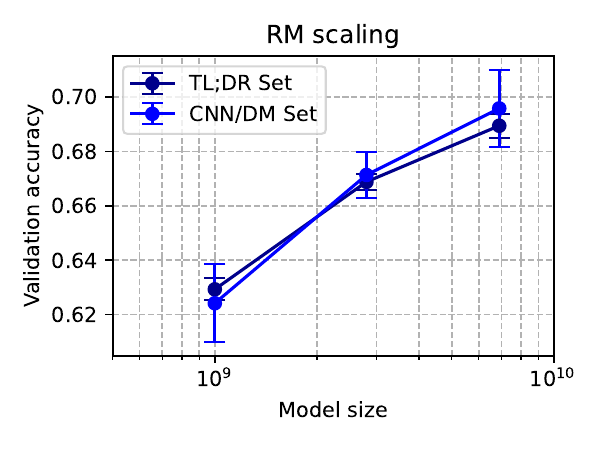}
    \caption{Top left is the RM training accuracy across the model sizes for one epoch of the train split of the preference dataset (92,858 episodes); top right is the RM loss; bottom left is the actual chosen reward scalars; bottom right is the scaling behavior of the reward modeling validation accuracy on the TL;DR set and CNN/DM set. Note the validation accuracy is lower because the validation set has out-of-distribution (OOD) data as explained in \ref{detail:ds-divsers-validation}.}
    \label{fig:rm}
\end{figure}
\section{Reward Model Details}

\begin{dlist}[resume] 
  \item \label{detail:ds-specification}{\textbf{RM Training -> Setups}}
\end{dlist}

We follow \citet{stiennon2020learning}'s original setting to train the RM, except that we used a different learning rate (Table~\ref{tab:rm-hyper}).

\begin{table}[h]
\centering
\caption{Reward modeling hyperparameters}
\label{tab:rm-hyper}
\begin{tabular}{lc}
\toprule
\textbf{Hyperparameter} & \textbf{Default Value} \\
\midrule
Number of Train Epochs & 1 (or 92,858 episodes) \\
Optimizer & AdamW ($\epsilon=1e-5$, $\texttt{lr}=3e-6$) \\
Scheduler & Cosine \\
Batch Size & 64 \\
\bottomrule
\end{tabular}
\end{table}

\begin{dlist}[resume] 
  \item \label{detail:ds-specification}{\textbf{RM Training -> Reward head initialization}}
\end{dlist}

We follow \citet{stiennon2020learning}'s original setting to initialize the RM from the trained SFT model and create a linear heard to output reward scalar with with
weights initialized according to $\mathcal{N}(0, 1/\sqrt{(d_{\text{model}} + 1)})$
(\href{https://github.com/openai/summarize-from-feedback/blob/700967448d10004279f138666442bf1497d0e705/summarize_from_feedback/query_response_model.py\#L106-L108}{\texttt{query\_response\_model.py\#L106-L108}}) \footnote{Note \cite{stiennon2020learning} have a minor typo of saying the initialization was according to $\mathcal{N}(0, 1/(d_{\text{model}} + 1))$, but the reference code clearly indicates otherwise.}

\begin{dlist}[resume] 
  \item \label{detail:ds-specification}{\textbf{RM Training -> Extract reward from the EOS token}}
\end{dlist}
When obtaining the scalar reward, the RM does a forward pass on the sequence and extracts the reward only on the EOS token. (\href{https://github.com/openai/summarize-from-feedback/blob/700967448d10004279f138666442bf1497d0e705/summarize_from_feedback/reward_model.py}{\texttt{reward\_model.py}}) This is implemented by finding the first index of the padding token and then minus 1. If the padding token does not exist, the extracted reward will then be logits corresponding to the last token of the sequence -- if that token is not the EOS token, its reward won't be used for PPO training, as explained later in PPO's EOS trick -- \ref{detail:eos-trick}).

Note that \citet{stiennon2020learning} choose the \texttt{<|endoftext|>} from the base model as the EOS token to extract the reward, but it is possible to use another special token. For example, Andrej Karpathy mentioned that the reward is extracted at \texttt{<|reward|>} in OpenAI's newer GPT systems\footnote{\url{https://youtu.be/bZQun8Y4L2A?t=956}}.

\begin{figure}[t]
    \centering
    \includegraphics[width=0.99\linewidth]{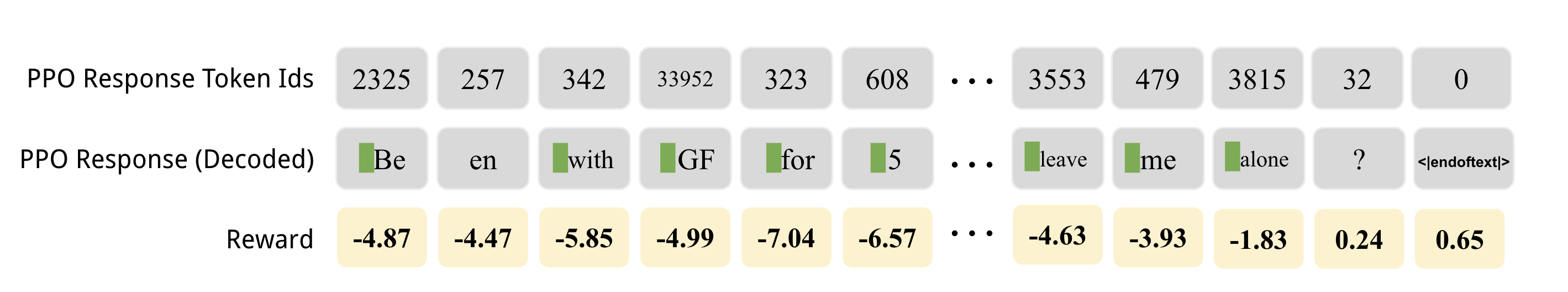}
    \caption{A 1B PPO model's response and its corresponding reward logits from a 1B RM. Here, we use Pythia's tokenizer, so \texttt{0} denotes the EOS token and \texttt{0.65} is the extracted EOS reward. Notice how the logits of non-EOS tokens are almost always negative -- we see this behavior in all the response-reward-logits pairs from all policies and RMs.}
    \label{fig:actual-rewards}
\end{figure}
\begin{dlist}[resume] 
  \item \label{detail:reward-logits}{\textbf{RM Training -> Most values in the reward logits are non-valid and negative; only the reward logit at the EOS token are valid}}
\end{dlist}
What do the reward logits actually look like in these trained RMs? We include an example in Figure~\ref{fig:actual-rewards}. We noticed the logits of non-EOS tokens are almost always negative in all the response-reward-logits pairs from all policies and RMs.

\begin{dlist}[resume] 
  \item \label{detail:ds-specification}{\textbf{RM Training -> Minor numerical differences between extracting reward with left and right padded queries}}
\end{dlist}

During RM training, the sequences are padded from the right with the shape \texttt{(B, 638)}. However, left-padding the query is required for generation in PPO training. The query has shape (B, 512), and after generation (with sequence length = 53), the query and response batch shape becomes (B, 565). As a result, we need to adjust the attention masks during RM forward calls.

Numerical note: Left-padding vs. right-padding can introduce minor numerical differences. For instance, in the 6.9B RM, the average reward scalar difference on the SFT dataset between the two padding methods is \texttt{-0.000544150301720947}. This difference is generally negligible.

\begin{dlist}[resume] 
  \item \label{detail:ds-specification}{\textbf{RM Training -> Reward normalization based on SFT demonstrations}}
\end{dlist}
\citet{stiennon2020learning} suggested that ``at the end of training, we normalize the reward model outputs such that the reference summaries from our dataset achieve a mean score of 0.'' We applied the same procedure by iterating through the SFT dataset and calculating the rewards of the query and reference responses, then calculating the mean reward and setting it as a bias in the reward head.

\subsection{RM training results}

The RM training loss, accuracy, and chosen reward value can be found in Figure~\ref{fig:rm}. The training accuracy and losses appear stable. Overall, larger RMs have higher validation accuracy on both TL;DR and CNN/DM sets. Note the validation accuracy on the CNN/DM is very encouraging -- the RM has never trained on CNN/DM data! We also performed a comprehensive evaluation of the trained RM on the validation set and calculated the aggregated mean and standard deviation for each batch, split, and confidence in Table~\ref{tab:various-accuraries}.

\begin{table}[]

    \caption{The mean and standard deviation of various metrics of the reward models across four random seeds. The table shows the metric names across different batches, confidences, and splits. There is limited documentation from \citet{stiennon2020learning} about these batches and splits, but nevertheless interesting to see this table.}
    \label{tab:various-accuraries}
\begin{tabular}{|ccc|c|c|c|}
\toprule[1pt]
\multicolumn{3}{|c|}{\textbf{Metric Names}}                                                                                                                                                                                    & \textbf{1B} & \textbf{2.8B} & \textbf{6.9B} \\ \midrule
\multicolumn{1}{|c|}{\multirow{4}{*}{Reward}}                                                         & \multicolumn{2}{c|}{Max}                                                                                      & 8.273 ± 0.993 & 5.961 ± 2.45 & 11.75 ± 2.203     \\ \cline{2-6} 
\multicolumn{1}{|c|}{}                                                                                & \multicolumn{2}{c|}{Mean}                                                                                     & 2.114 ± 0.939 & 0.925 ± 2.386 & 4.783 ± 1.545      \\ \cline{2-6} 
\multicolumn{1}{|c|}{}                                                                                & \multicolumn{2}{c|}{Min}                                                                                      & -5.461 ± 1.754 & -5.039 ± 2.547 & -3.016 ± 1.421      \\ \cline{2-6} 
\multicolumn{1}{|c|}{}                                                                                & \multicolumn{2}{c|}{Std}                                                                                      & 1.657 ± 0.086 & 1.361 ± 0.206 & 1.912 ± 0.078      \\ \midrule
\multicolumn{1}{|c|}{\multirow{29}{*}{\begin{tabular}[c]{@{}c@{}}Validation\\ Accuracy\end{tabular}}} & \multicolumn{1}{c|}{\multirow{18}{*}{\begin{tabular}[c]{@{}c@{}}Batch\\ Number\end{tabular}}} & Overall Accuracy             & 0.628 ± 0.002 & 0.669 ± 0.003 & 0.689 ± 0.004      \\ \cline{3-6} 
\multicolumn{1}{|c|}{}                                                                                & \multicolumn{1}{c|}{}                                                                         & 6             & 0.661 ± 0.016 & 0.682 ± 0.024 & 0.709 ± 0.009      \\ \cline{3-6} 
\multicolumn{1}{|c|}{}                                                                                & \multicolumn{1}{c|}{}                                                                         & 7              & 0.694 ± 0.023 & 0.718 ± 0.011 & 0.732 ± 0.014      \\ \cline{3-6} 
\multicolumn{1}{|c|}{}                                                                                & \multicolumn{1}{c|}{}                                                                         & 8             & 0.598 ± 0.014 & 0.63 ± 0.008 & 0.636 ± 0.009      \\ \cline{3-6} 
\multicolumn{1}{|c|}{}                                                                                & \multicolumn{1}{c|}{}                                                                         & 9             & 0.578 ± 0.005 & 0.687 ± 0.017 & 0.691 ± 0.015      \\ \cline{3-6} 
\multicolumn{1}{|c|}{}                                                                                & \multicolumn{1}{c|}{}                                                                         & 10            & 0.626 ± 0.007 & 0.655 ± 0.015 & 0.69 ± 0.007      \\ \cline{3-6} 
\multicolumn{1}{|c|}{}                                                                                & \multicolumn{1}{c|}{}                                                                         & 11            & 0.508 ± 0.01 & 0.603 ± 0.004 & 0.653 ± 0.021      \\ \cline{3-6} 
\multicolumn{1}{|c|}{}                                                                                & \multicolumn{1}{c|}{}                                                                         & 12            & 0.686 ± 0.007 & 0.697 ± 0.009 & 0.704 ± 0.007     \\ \cline{3-6} 
\multicolumn{1}{|c|}{}                                                                                & \multicolumn{1}{c|}{}                                                                         & 13            & 0.771 ± 0.016 & 0.708 ± 0.013 & 0.745 ± 0.008      \\ \cline{3-6} 
\multicolumn{1}{|c|}{}                                                                                & \multicolumn{1}{c|}{}                                                                         & 14            & 0.577 ± 0.031 & 0.588 ± 0.01 & 0.634 ± 0.011      \\ \cline{3-6} 
\multicolumn{1}{|c|}{}                                                                                & \multicolumn{1}{c|}{}                                                                         & 15            & 0.628 ± 0.021 & 0.699 ± 0.011 & 0.671 ± 0.01     \\ \cline{3-6} 
\multicolumn{1}{|c|}{}                                                                                & \multicolumn{1}{c|}{}                                                                         & 16            & 0.707 ± 0.017 & 0.737 ± 0.002 & 0.761 ± 0.006      \\ \cline{3-6} 
\multicolumn{1}{|c|}{}                                                                                & \multicolumn{1}{c|}{}                                                                         & 17            & 0.752 ± 0.014 & 0.757 ± 0.003 & 0.734 ± 0.018      \\ \cline{3-6} 
\multicolumn{1}{|c|}{}                                                                                & \multicolumn{1}{c|}{}                                                                         & 18            & 0.733 ± 0.015 & 0.741 ± 0.025 & 0.771 ± 0.011     \\ \cline{3-6} 
\multicolumn{1}{|c|}{}                                                                                & \multicolumn{1}{c|}{}                                                                         & 19            & 0.636 ± 0.02 & 0.688 ± 0.012 & 0.714 ± 0.01      \\ \cline{3-6} 
\multicolumn{1}{|c|}{}                                                                                & \multicolumn{1}{c|}{}                                                                         & 20            & 0.671 ± 0.005 & 0.705 ± 0.008 & 0.711 ± 0.007      \\ \cline{3-6} 
\multicolumn{1}{|c|}{}                                                                                & \multicolumn{1}{c|}{}                                                                         & 22            & 0.587 ± 0.006 & 0.632 ± 0.009 & 0.651 ± 0.005      \\ \cline{2-6} 
\multicolumn{1}{|c|}{}                                                                                & \multicolumn{1}{c|}{\multirow{9}{*}{Confidence}}                                              & 1             & 0.693 ± 0.012 & 0.758 ± 0.005 & 0.795 ± 0.004      \\ \cline{3-6} 
\multicolumn{1}{|c|}{}                                                                                & \multicolumn{1}{c|}{}                                                                         & 2             & 0.669 ± 0.011 & 0.706 ± 0.012 & 0.718 ± 0.007      \\ \cline{3-6} 
\multicolumn{1}{|c|}{}                                                                                & \multicolumn{1}{c|}{}                                                                         & 3             & 0.635 ± 0.005 & 0.656 ± 0.011 & 0.674 ± 0.003      \\ \cline{3-6} 
\multicolumn{1}{|c|}{}                                                                                & \multicolumn{1}{c|}{}                                                                         & 4             & 0.58 ± 0.005 & 0.562 ± 0.006 & 0.589 ± 0.009      \\ \cline{3-6} 
\multicolumn{1}{|c|}{}                                                                                & \multicolumn{1}{c|}{}                                                                         & 6             & 0.563 ± 0.006 & 0.574 ± 0.012 & 0.581 ± 0.009      \\ \cline{3-6} 
\multicolumn{1}{|c|}{}                                                                                & \multicolumn{1}{c|}{}                                                                         & 7             & 0.568 ± 0.006 & 0.635 ± 0.007 & 0.655 ± 0.008      \\ \cline{3-6} 
\multicolumn{1}{|c|}{}                                                                                & \multicolumn{1}{c|}{}                                                                         & 8             & 0.609 ± 0.011 & 0.691 ± 0.008 & 0.704 ± 0.007      \\ \cline{3-6} 
\multicolumn{1}{|c|}{}                                                                                & \multicolumn{1}{c|}{}                                                                         & 9             & 0.694 ± 0.007 & 0.744 ± 0.005 & 0.765 ± 0.009      \\ \cline{2-6} 
\multicolumn{1}{|c|}{}                                                                                & \multicolumn{1}{c|}{\multirow{2}{*}{Split Valid}}   & 1             & 0.639 ± 0.003 & 0.667 ± 0.007 & 0.69 ± 0.007     \\ \cline{3-6} 
\multicolumn{1}{|c|}{}                                                                                & \multicolumn{1}{c|}{}                                                                         & 2             & 0.621 ± 0.003 & 0.669 ± 0.003 & 0.688 ± 0.002      \\ \midrule
\multicolumn{1}{|c|}{\multirow{14}{*}{\begin{tabular}[c]{@{}c@{}}Cnndm\\ Accuracy\end{tabular}}}      & \multicolumn{2}{c|}{Overall Accuracy}                                                                                    & 0.627 ± 0.013 & 0.665 ± 0.01 & 0.686 ± 0.003     \\ \cline{2-6} 
\multicolumn{1}{|c|}{}                                                                                & \multicolumn{1}{c|}{\multirow{3}{*}{Batch}}                                                   & Batch0\_cnndm & 0.679 ± 0.06 & 0.714 ± 0.027 & 0.723 ± 0.009      \\ \cline{3-6} 
\multicolumn{1}{|c|}{}                                                                                & \multicolumn{1}{c|}{}                                                                         & Cnndm0        & 0.772 ± 0.009 & 0.677 ± 0.017 & 0.714 ± 0.031       \\ \cline{3-6} 
\multicolumn{1}{|c|}{}                                                                                & \multicolumn{1}{c|}{}                                                                         & Cnndm2        & 0.564 ± 0.012 & 0.646 ± 0.013 & 0.666 ± 0.005     \\ \cline{2-6} 
\multicolumn{1}{|c|}{}                                                                                & \multicolumn{1}{c|}{\multirow{9}{*}{Confidence}}                                              & 1             & 0.589 ± 0.094 & 0.804 ± 0.043 & 0.815 ± 0.022      \\ \cline{3-6} 
\multicolumn{1}{|c|}{}                                                                                & \multicolumn{1}{c|}{}                                                                         & 2             & 0.641 ± 0.139 & 0.661 ± 0.107 & 0.732 ± 0.036     \\ \cline{3-6} 
\multicolumn{1}{|c|}{}                                                                                & \multicolumn{1}{c|}{}                                                                         & 3             & 0.5 ± 0.037 & 0.771 ± 0.023 & 0.736 ± 0.014     \\ \cline{3-6} 
\multicolumn{1}{|c|}{}                                                                                & \multicolumn{1}{c|}{}                                                                         & 4             & 0.597 ± 0.053 & 0.6 ± 0.028 & 0.615 ± 0.025      \\ \cline{3-6} 
\multicolumn{1}{|c|}{}                                                                                & \multicolumn{1}{c|}{}                                                                         & 6             & 0.671 ± 0.05 & 0.587 ± 0.031 & 0.568 ± 0.02      \\ \cline{3-6} 
\multicolumn{1}{|c|}{}                                                                                & \multicolumn{1}{c|}{}                                                                         & 7             & 0.743 ± 0.095 & 0.646 ± 0.036 & 0.741 ± 0.032      \\ \cline{3-6} 
\multicolumn{1}{|c|}{}                                                                                & \multicolumn{1}{c|}{}                                                                         & 8             & 0.594 ± 0.092 & 0.632 ± 0.056 & 0.662 ± 0.056     \\ \cline{3-6} 
\multicolumn{1}{|c|}{}                                                                                & \multicolumn{1}{c|}{}                                                                         & 9             & 0.65 ± 0.094 & 0.777 ± 0.054 & 0.812 ± 0.061     \\ \cline{2-6} 
\multicolumn{1}{|c|}{}                                                                                & \multicolumn{1}{c|}{Split Valid}                                                                    & 2        & 0.627 ± 0.013 & 0.665 ± 0.01 & 0.686 ± 0.003      \\
\bottomrule[1pt]
\end{tabular}
\end{table}

\begin{dlist}[resume] 
  \item \label{detail:different-accuracies}{\textbf{RM Training -> Different batches / confidences have different accuracies }}
\end{dlist}

As shown in Table~\ref{tab:various-accuraries}, different annotated batches could have different validation accuracies. Several observations:

\begin{enumerate}
    \item The 1B model's validation accuracy at batch 11 is 0.508, which is no different from a coin toss
    \item The 1B model's validation accuracy at batch 13 is 0.771, a much higher accuracy. 
    
    \item The trained RMs generally have high accuracy for high-confidence preference pairs, which makes sense (e.g., the 6.9B model's validation accuracy with accuracy 9 is 0.765). 
    \item Interestingly, the trained RMs also have high accuracy for very low-confidence preference pairs for some reason (e.g., 6.9B model's validation accuracy with accuracy 1 is 0.795). 
\end{enumerate}

\begin{dlist}[resume] 
  \item \label{detail:gpt3.5-rm-consistency}{\textbf{RM Training -> Preference consistency rate with GPT3.5}}
\end{dlist}

\begin{figure}[t]
    \centering
    \includegraphics[width=0.44\linewidth]{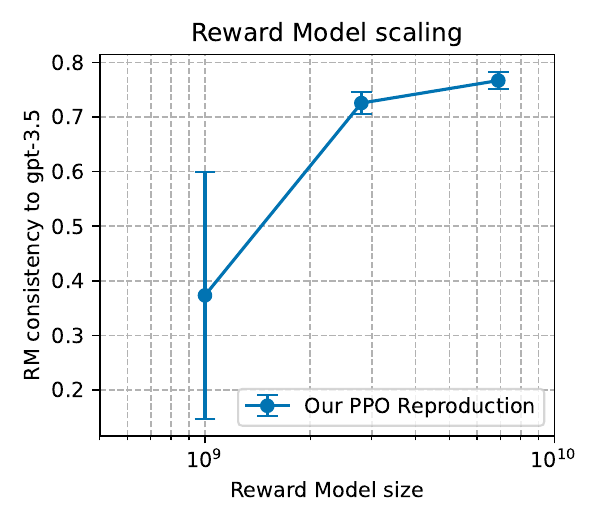}
    \includegraphics[width=0.55\linewidth]{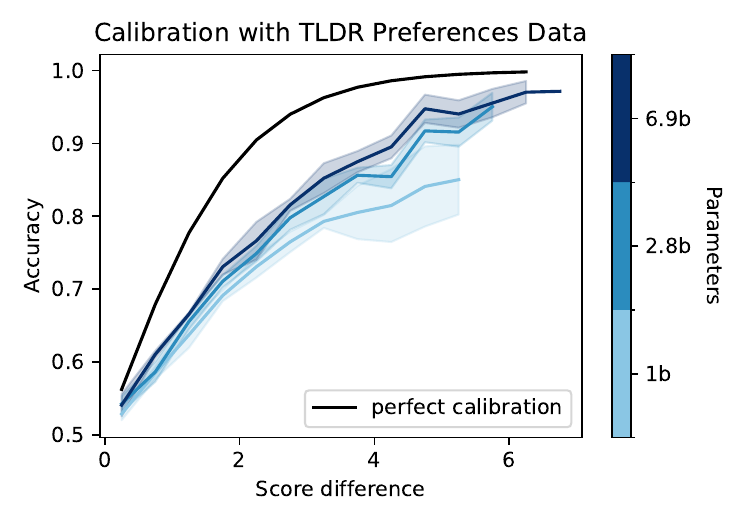}
    \caption{\textbf{(Left)} RM agreement rate with GPT3.5 across different model sizes. \textbf{(Right)} RM calibration -- the black line is the perfect calibration $\frac{1}{1+e^{-\Delta}}$, where $\Delta$ is the the score difference (Equation~\ref{eq:rm-alternative-form})~\citep{bai2022training}.}
    \label{fig:gpt3.5-rm-agreement}
\end{figure}

As per Goodhart’s law when a metric becomes the optimization goal, it ceases to be a good metric \citep{gao2023scaling}. To verify whether RM is overfitting the current dataset's accuracy after training, we introduced GPT3.5 as an external LLM-judge~\citep{zheng2024judging}. By comparing the preferences of GPT3.5 and RM for the same set of preference data, we assess the actual training effects of RM across different model sizes. As depicted in Figure \ref{fig:gpt3.5-rm-agreement}, we have observed the following:

\begin{enumerate}
\item For the 1B-sized model, the average preference consistency in multiple random experiments is close to 0.4, indicating that the 1B model has captured a different set of preference, contrary to GPT3.5.
\item The average preference consistency rates for the 2.8B and 6.9B models are 0.726 and 0.767, respectively, both exceeding 0.5. Compared to the 1B model, as the model size increases, RM can exhibit preferences similar to GPT3.5.
\item The difference in average preference consistency rates between the 2.8B and 6.9B models is 0.041, whereas the difference between the 2.8B and 1B models is 0.353. The gains from increasing model size are gradually diminishing (maybe also because the accuracy is already high).
\end{enumerate}

\begin{dlist}[resume] 
  \item \label{detail:dpo-rm-acc}{\textbf{RM Training -> RM calibration}}
\end{dlist}
RMs should predict the log probabilities that humans will prefer one completion versus others; to this end, \citet{bai2022training} propose a visualization technique to see if these probabilities are accurate and well-calibrated. The idea is to plot the score difference between the chosen and rejected pairs in the x-axis and the accuracy of the RM in the y-axis. Intuitively, the larger the score difference, the more confident the model is that one completion is better than the other. We plot the RM calibration in Figure~\ref{fig:gpt3.5-rm-agreement}.

Overall, we do find a positive correlation between accuracy and score difference -- this is a good sign because models become more accurate as they become more confident (i.e., higher score difference). However, the RMs are still under-calibrated, probably due to the diverse validation set (\ref{detail:ds-divsers-validation}) and different accuracies in these validation set (\ref{detail:different-accuracies}).



\begin{dlist}[resume] 
  \item \label{detail:dpo-rm-acc}{\textbf{RM Training -> Comparison with DPO's implicit reward modeling}}
\end{dlist}
We also trained equivalent DPO models to compare the validation accuracy. We use the same hyperparameters used for RM training, except DPO also has a $\beta$ hyperparameter, as shown in Table~\ref{tab:hyper-dpo}.

\begin{table}[h]
\centering
\caption{DPO hyperparameters}
\label{tab:hyper-dpo}
\begin{tabular}{lc}
\toprule
\textbf{Hyperparameter} & \textbf{Default Value} \\
\midrule
Number of Train Epochs & 1 (or 92,858 episodes) \\
Optimizer & AdamW ($\epsilon=1e-5$, $\texttt{lr}=3e-6$) \\
Scheduler & Cosine \\
Batch Size & 64 \\
$\beta$ (KL Penalty Coefficient for RLHF) & 0.05 \\
\bottomrule
\end{tabular}
\end{table}

\begin{figure}[t]
    \centering
    \includegraphics[width=0.98\linewidth]{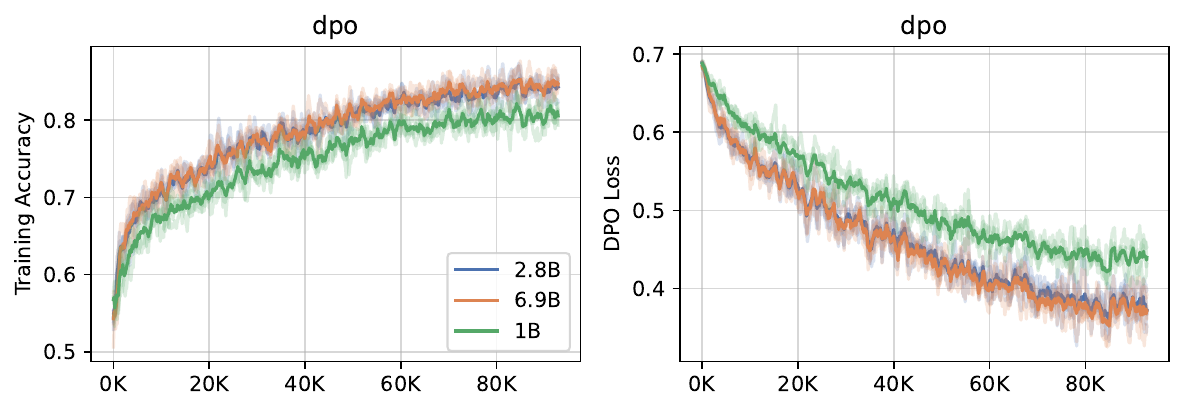}
    \includegraphics[width=0.49\linewidth]{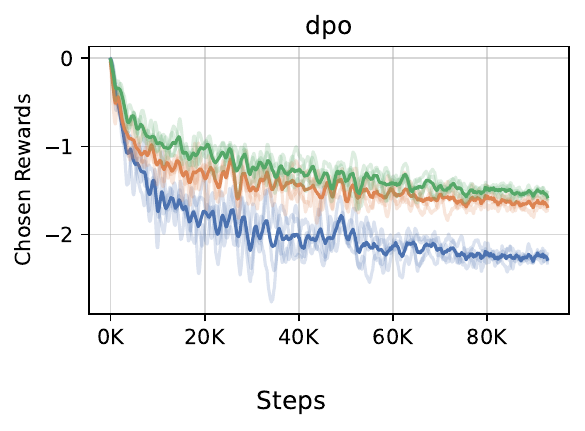}
    \includegraphics[width=0.49\linewidth]{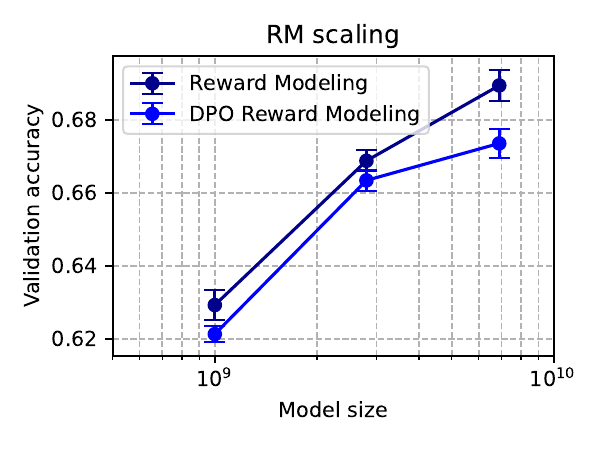}
    \caption{Top left is DPO's implicit RM training accuracy (Equation~\ref{eq:dpo-implicit-rm}), top right DPO's loss, bottom left DPO's actual chosen reward values, and bottom right the scaling behavior of validation accuracy on the TL;DR set between regular reward modeling and DPO's implicit modeling. We observed a regression of validation accuracy in DPO (\ref{detail:dpo-rm-acc}).}
    \label{fig:dpo}
\end{figure}

During training, we controlled the preference dataset iteration order as well, so this should be a fair comparison of explicit versus DPO's implicit reward modeling losses. The training curves can be found in Figure~\ref{fig:dpo}. There are a couple of interesting observations:
\begin{enumerate}
    \item \textbf{Validation accuracy regression in DPO}: We found a regression in the validation accuracy in DPO's final evaluation, and this finding holds true across 3 model sizes and 4 random seeds; this suggests DPO's implicit reward modeling may not be equivalent to the regular explicit reward modeling. There are several factors that we suspect may be responsible for this difference. First, regular reward modeling's loss only applies to the EOS token, whereas in DPO, the loss applies to all the tokens. Second, DPO also has the RLHF $\beta$ parameter in the loss, which is not present in regular reward modeling's loss (we chose $\beta=0.05$ to match PPO's setting). Third, by modeling the reward as the difference in logprobs between model and reference model, DPO's objective may be harder to optimize in practice than the RM objective. Whereas an RM can easily learn large changes in reward using the linear head, DPO must drastically change many tokens' logprobs to do the same. 
    \item \textbf{Decreasing chosen rewards}: DPO's chosen and rejected rewards both generally decrease, whereas regular reward modeling's chosen rewards fluctuate, see Figure~\ref{fig:rm}. 
\end{enumerate}
We advocate for more research on how DPO's loss systematically affects RM accuracies.

\section{PPO Details}

\begin{dlist}[resume] 
  \item \label{detail:ds-specification}{\textbf{PPO Training -> Setups}}
\end{dlist}

Our PPO setup closely follows \citet{stiennon2020learning}, except for a modified learning rate (Table~\ref{tab:ppo-hyper}).

\begin{table}[h]
\centering
\caption{PPO hyperparameters.}
\begin{tabular}{ll} 
\toprule
\textbf{Hyperparameter} & \textbf{Default Value} \\
\midrule
Episodes & 1,000,000 (or $\sim$8.56 epochs) \\ 
Optimizer & AdamW ($\epsilon=1e-5$, $\texttt{lr}=3e-6$) \\
Scheduler & Linear \\
Batch Size & 512 \\
$\beta$ (KL Penalty Coefficient for RLHF) & 0.05 \\
$\gamma$ (Discount Factor) & 1.0 \\ 
$\lambda$ (for GAE) & 0.95 \\ 
$N_\text{mb}$ Number of Mini-batches & 1 \\
$K$ (Number of PPO Update Iteration Per Epoch)& 4 \\
$\varepsilon$ (PPO's Policy Clipping Coefficient) & 0.2 \\
$\hat{\varepsilon}$ (Value Clipping Coefficient) & 0.2 \\
$c_1$ (Value Function Coefficient)& 0.1\\
Value Function Loss Clipping
& True\\
Sampling Temperature & 0.7 \\
\bottomrule
\end{tabular}
\label{tab:ppo-hyper}
\end{table}

\begin{dlist}[resume] 
  \item \label{detail:ds-specification}{\textbf{PPO Training -> Re-use the SFT dataset and shuffle when reaches the end}}
\end{dlist}
\citet{stiennon2020learning} trains the PPO models for 1M episodes, but the \texttt{train} split of the SFT dataset is only of size 116,722, so an educated guess is that the SFT dataset is re-used repeatedly during PPO training. Specifically, we should shuffle the SFT dataset and sample from it without replacement; once the dataset is depleted, we should reshuffle it again and sample without replacement; we continue this process until we reach 1M episodes. 
(\href{https://github.com/openai/summarize-from-feedback/blob/700967448d10004279f138666442bf1497d0e705/summarize_from_feedback/datasets/__init__.py#L27-L39}{\texttt{datasets/\_\_init\_\_.py\#L27-L39}})

\begin{dlist}[resume] 
  \item \label{detail:value-network-improve}{\textbf{PPO Training -> Value model initializes from the reward model; trained value model looks like a per-token RM.}}
\end{dlist}

Similar to the settings in \citet{stiennon2020learning}, we initialize the value network based on the reward model. This warm-starting of the value network can greatly improve initial gradients to the policy and reduce drift / alignment tax over training \citep{noukhovitch_language_2023}. Because of this, the values generated by the value network will look identical to the example in Figure~\ref{fig:actual-rewards} (\ref{detail:reward-logits}), where the values of most tokens are negative numbers except for the EOS token. 

\begin{figure}[h]
    \centering
     \includegraphics[width=0.6\linewidth]{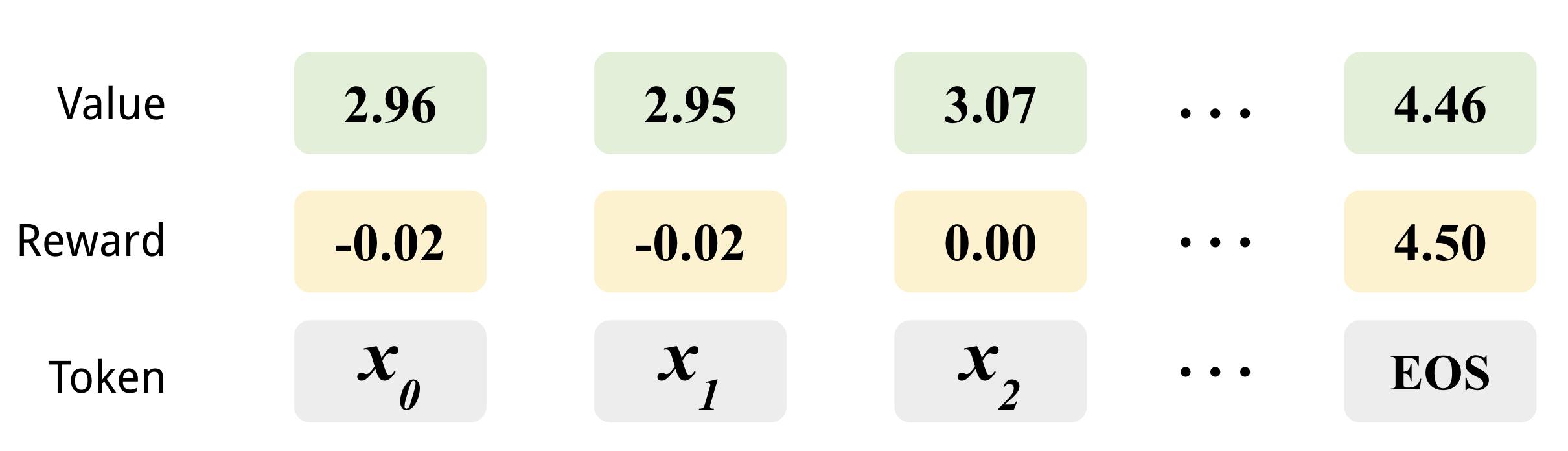}
    \caption{Reward and values of a completion. The score from the reward model at the EOS token is \texttt{4.50} while the rest of \texttt{reward} numbers are per-token KL penalty scores.}
    \label{fig:reward_vs_value}
\end{figure}

However, in RL training, the value function would aim to predict the end-of-episode return at each timestep / token, effectively acting as a per-token RM. In Figure~\ref{fig:reward_vs_value}, we show the rewards and values of a completion, where the \texttt{4.5000} is the score from the RM corresponding to the EOS token. The other values in the \texttt{rewards} are per-token KL penalty. See \url{https://wandb.ai/costa-huang/tldr_summarize/runs/9f6t868e/logs} for the full log.

\begin{dlist}[resume] 
  \item \label{detail:eos-trick}{\textbf{PPO Training -> ``EOS trick'' to ensure scores from the RM is valid}}
\end{dlist}
One interesting phenomenon we observed with the original checkpoint of \citet{stiennon2020learning} is that the generated summaries always have less than 48 tokens and also end with an EOS token -- this makes the comparison with the reference summaries more fair because the reference summaries are also less than 48 tokens (\ref{detail:ds-token-len}). We suspect the following processes likely achieve it:

\begin{enumerate}
    \item Always samples a fixed amount of 48 tokens (corresponding to 53 tokens in our reproduction) from the vocabulary (\href{https://github.com/openai/summarize-from-feedback/blob/700967448d10004279f138666442bf1497d0e705/summarize_from_feedback/policy.py#L48}{\texttt{policy.py\#L48}}). In particular, the model will continue to sample tokens even if it encounters an EOS token (this means after the EOS token the generations are unconditional).
    \item Given the 48 tokens, the script then ``truncates'' at the EOS token, filling the tokens after the EOS token as padding tokens (\href{https://github.com/openai/summarize-from-feedback/blob/700967448d10004279f138666442bf1497d0e705/summarize_from_feedback/sample.py#L146}{\texttt{sample.py\#L146}}, \href{https://github.com/openai/summarize-from-feedback/blob/700967448d10004279f138666442bf1497d0e705/summarize_from_feedback/tasks.py#L57-L62}{\texttt{tasks.py\#L57-L62}}).
    \item This ``truncated'' response is then passed to the reward model to get a score; if the response does not contain any EOS token, we suspect \cite{stiennon2020learning} replaced the score with -1, similar to the procedure described by \citet{ziegler2019fine,huang2024thenimplementation}.
\end{enumerate}



The EOS trick serves a couple of purposes for RL:
\begin{enumerate}
    \item \textbf{Defined reward scores:} It guarantees that the PPO model receives a defined reward score. This is important because the RM only backpropagates loss on the EOS token during training. \emph{Without an EOS token, the completion's reward is undefined.} The EOS trick assigns a constant -1 reward in these cases.
    \item \textbf{Constraining completion length:} The trick encourages the model to generate concise completions -- longer completions that lack an EOS token are penalized with a -1 reward.
\end{enumerate}
Essentially, the EOS trick helps ensure the completion ends with an EOS token, so rewards are well-defined.


\begin{figure}[t]
    \centering
    \includegraphics[width=0.98\linewidth]{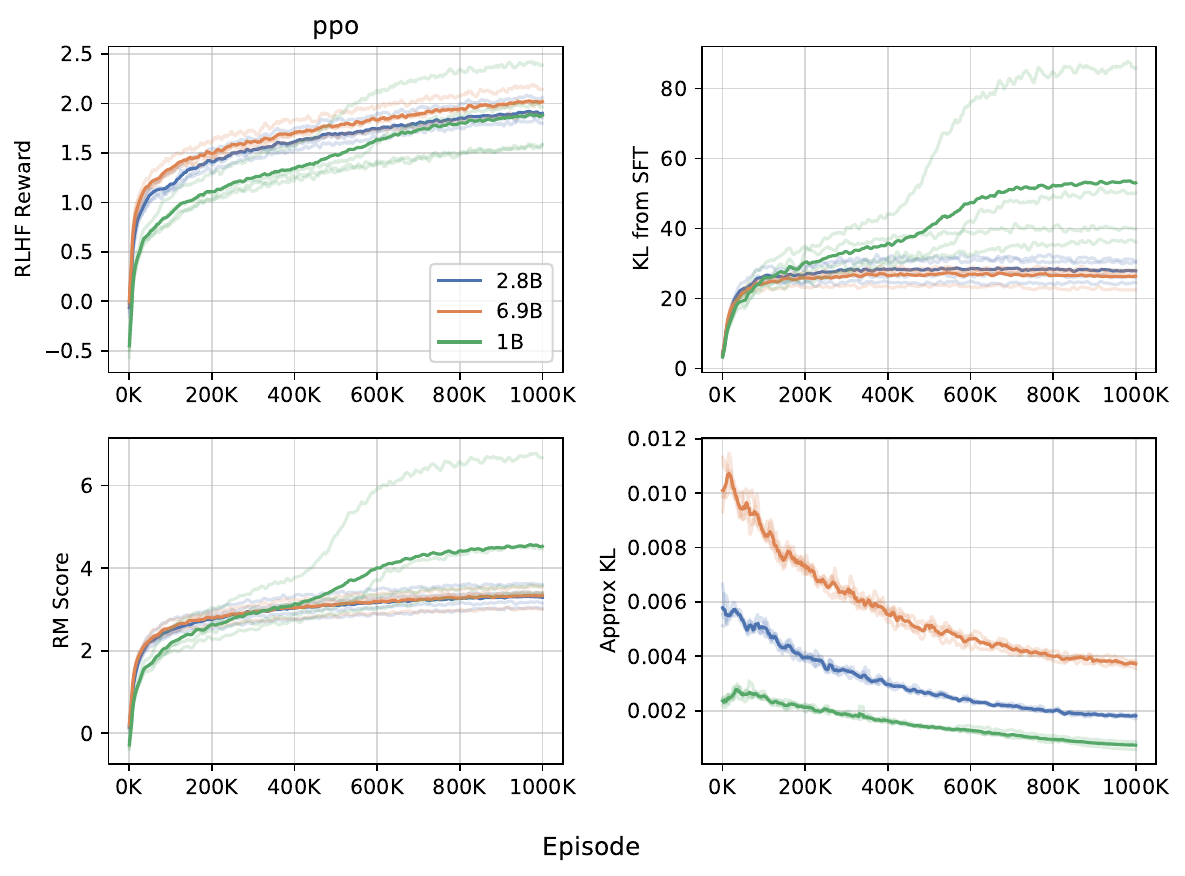}
    \caption{Top left shows PPO's RLHF's reward $R(x, y)$ (Equation~\ref{eq:RL}). The top right figure shows the mean of the sum of per-token KL divergence between the RL and SFT policies. The bottom left shows the scores obtained from the reward model.}
    \label{fig:ppo-learning}
\end{figure}

\begin{dlist}[resume] 
  \item \label{detail:reward-whitening}{\textbf{PPO Training -> (Optional) Reward whitening}}
\end{dlist}
As indicated in \citet{huang2024thenimplementation}, \citet{ziegler2019fine} implement a \texttt{whiten} function that looks like below, basically normalizing the values by subtracting its mean followed by dividing by its standard deviation. Optionally, \texttt{whiten} can shift back the mean of the whitened values with \texttt{shift\_mean=True}. 
In each minibatch, PPO could whiten the reward \texttt{whiten(rewards, shift\_mean=False)} without shifting the mean (\href{https://github.com/openai/lm-human-preferences/blob/cbfd210bb8b08f6bc5c26878c10984b90f516c66/lm_human_preferences/train_policy.py\#L325}{\texttt{lm\_human\_preferences/train\_policy.py\#L325})}.
\pagebreak
\begin{minted}{python}
def whiten(values, shift_mean=True):
    mean, var = torch.mean(values), torch.var(values, unbiased=False)
    whitened = (values - mean) * torch.rsqrt(var + 1e-8)
    if not shift_mean:
        whitened += mean
    return whitened
\end{minted}

\begin{dlist}[resume] 
  \item \label{detail:ds-specification}{\textbf{PPO Training -> Advantage whitening}}
\end{dlist}
Similar to practices identified in \cite{Engstrom2020Implementation,andrychowicz2021what,shengyi2022the37implementation}, PPO whitens the advantages \texttt{whiten(advantages)} with the shifted mean  \href{https://github.com/openai/lm-human-preferences/blob/cbfd210bb8b08f6bc5c26878c10984b90f516c66/lm_human_preferences/train_policy.py\#L338}{(\texttt{lm\_human\_preferences/train\_policy.py\#L338})}.

\begin{figure}[t]
    \centering
    \includegraphics[width=0.48\linewidth]{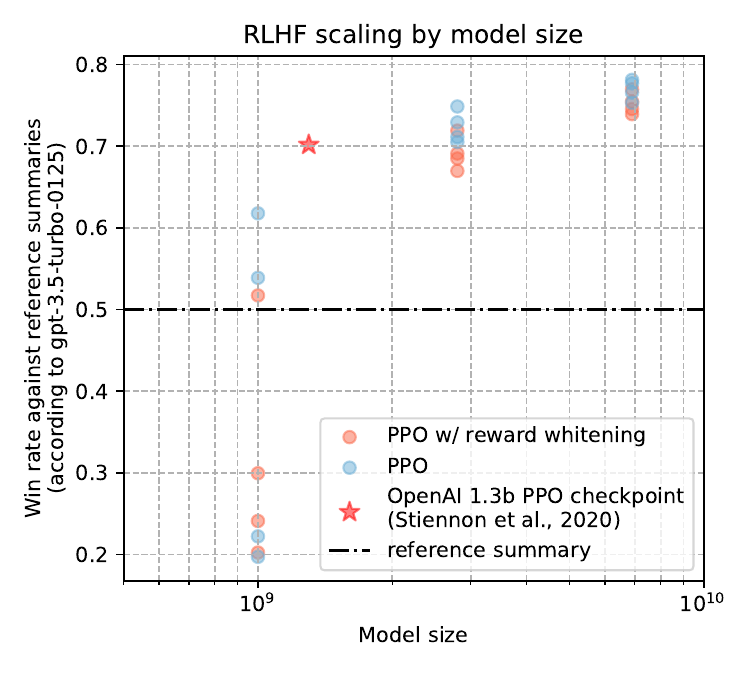}\includegraphics[width=0.48\linewidth]{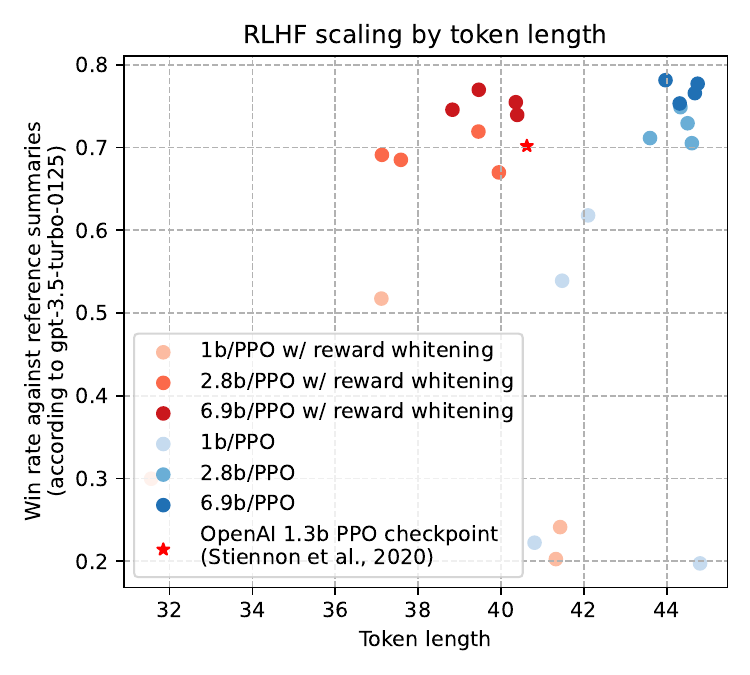}
    \caption{Left figure shows the RLHF preference scaling behavior across different model sizes with and without \ref{detail:reward-whitening}. The right figure then plots those data points with the x-axis being the average summary token length.}
    \label{fig:reward-whitening}
\end{figure}

\begin{figure}[t]
    \centering
    \includegraphics[width=0.48\linewidth]{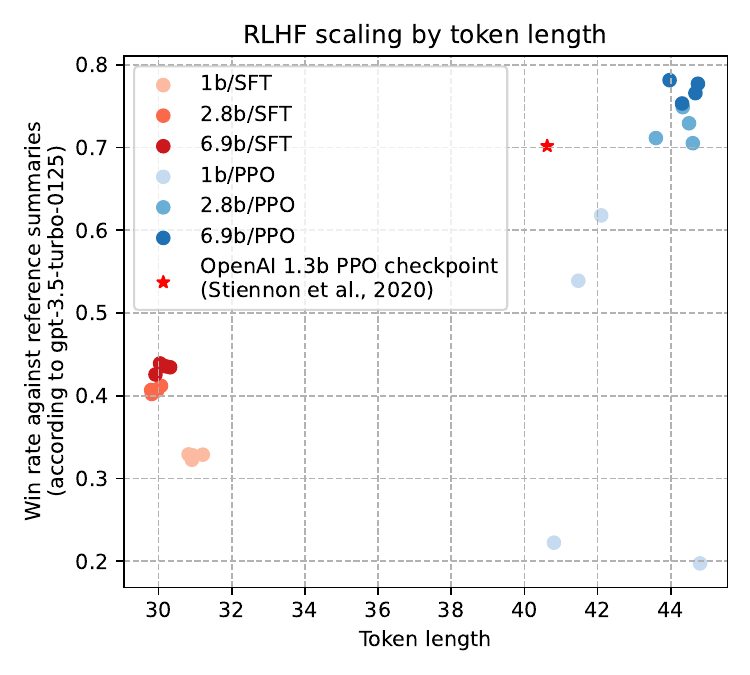}\includegraphics[width=0.48\linewidth]{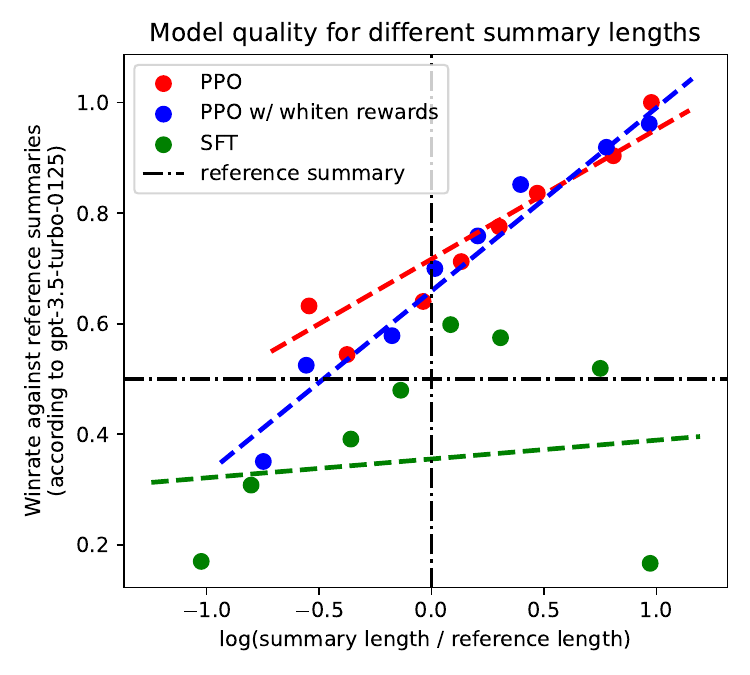}
    \caption{The left figure plots the win rate against reference summaries, with the x-axis being the average summary token length. The right figure shows the win rate of 6.9B SFT and PPO models for different summary lengths (one random seed; see Appendix~\ref{sec:full-log-length} for other seeds and model sizes).}
    \label{fig:log-length}
\end{figure}

\begin{table}[t]
\caption{Sample query, responses from the 1B SFT, PPO, and DPO models; scores are from a 6.9B model. We mark the response tokens ranked top 1 by the pre-trained model \sethlcolor{LightBlue}\hl{blue, meaning unshifted tokens}, tokens ranked within the top 3  \sethlcolor{LightYellow}\hl{yellow, meaning marginal tokens}, and tokens ranked beyond the top 3  \sethlcolor{LightRed}\hl{red, meaning shifted tokens}~\citep{Lin2024ReAlign}. Essentially,  \sethlcolor{LightRed}\hl{red} and \sethlcolor{LightYellow}\hl{yellow} tokens highlight what the SFT, PPO, and DPO models would do differently compared to the pre-trained model. We released the source code to load the model and generate this visualization in \url{https://github.com/vwxyzjn/summarize_from_feedback_details/blob/main/visualize_tokens.py}.}
\label{tab:ai2-visual}
\centering
\begin{small}
    \begin{tabular}{lp{7cm}l}
\toprule
Type & Content & Score (RM) \\
\midrule
Query & SUBREDDIT: r/AskReddit

TITLE: How do you get someone out of your head?

POST: Hi,
I'm 22, and I have been with my girlfriend for 5 years now. We recently moved together. We've always loved each other intensely.

Problem, I recently started to have feelings for an other person (a friend). This person has had a boyfriend for now 3 years, and has absolutely no ideas. Those feelings were so strong, it was hard to hide them. After 2 months of me being distant and really sad, my girlfriend forced me to say what was bothering me. I'm not a good liar, and now she knows.

We decided to give us a week alone, I went to my parents. 

Now, I'm completely lost. I keep on thinking about this person, and I hate that. I would like for those feelings to go away, to leave me alone. But I can't.  

What do I do? It's been 3 months now, and I'm just desperate.

TL;DR: & N/A \\\midrule
SFT Model Response & \sethlcolor{LightYellow}\hl{ I}\sethlcolor{LightRed}\hl{ have}\sethlcolor{LightRed}\hl{ feelings}\sethlcolor{LightBlue}\hl{ for}\sethlcolor{LightRed}\hl{ a}\sethlcolor{LightBlue}\hl{ friend}\sethlcolor{LightBlue}\hl{,}\sethlcolor{LightBlue}\hl{ and}\sethlcolor{LightBlue}\hl{ I}\sethlcolor{LightBlue}\hl{'m}\sethlcolor{LightBlue}\hl{ not}\sethlcolor{LightBlue}\hl{ sure}\sethlcolor{LightYellow}\hl{ how}\sethlcolor{LightBlue}\hl{ to}\sethlcolor{LightBlue}\hl{ get}\sethlcolor{LightBlue}\hl{ them}\sethlcolor{LightBlue}\hl{ out}\sethlcolor{LightBlue}\hl{ of}\sethlcolor{LightBlue}\hl{ my}\sethlcolor{LightBlue}\hl{ head}\sethlcolor{LightBlue}\hl{.}\sethlcolor{LightRed}\hl{<|endoftext|>} & -3.4151 \\\midrule
PPO Model Response & \sethlcolor{LightRed}\hl{ Be}\sethlcolor{LightRed}\hl{en}\sethlcolor{LightBlue}\hl{ with}\sethlcolor{LightRed}\hl{ GF}\sethlcolor{LightBlue}\hl{ for}\sethlcolor{LightBlue}\hl{ 5}\sethlcolor{LightBlue}\hl{ years}\sethlcolor{LightBlue}\hl{,}\sethlcolor{LightRed}\hl{ recently}\sethlcolor{LightBlue}\hl{ started}\sethlcolor{LightBlue}\hl{ to}\sethlcolor{LightBlue}\hl{ have}\sethlcolor{LightBlue}\hl{ feelings}\sethlcolor{LightBlue}\hl{ for}\sethlcolor{LightBlue}\hl{ another}\sethlcolor{LightBlue}\hl{ person}\sethlcolor{LightBlue}\hl{.}\sethlcolor{LightYellow}\hl{ I}\sethlcolor{LightRed}\hl{ love}\sethlcolor{LightBlue}\hl{ her}\sethlcolor{LightRed}\hl{ deeply}\sethlcolor{LightRed}\hl{ however}\sethlcolor{LightRed}\hl{ the}\sethlcolor{LightBlue}\hl{ feelings}\sethlcolor{LightBlue}\hl{ are}\sethlcolor{LightRed}\hl{ driving}\sethlcolor{LightBlue}\hl{ me}\sethlcolor{LightBlue}\hl{ crazy}\sethlcolor{LightBlue}\hl{.}\sethlcolor{LightYellow}\hl{ What}\sethlcolor{LightBlue}\hl{ do}\sethlcolor{LightBlue}\hl{ I}\sethlcolor{LightBlue}\hl{ do}\sethlcolor{LightBlue}\hl{?}\sethlcolor{LightRed}\hl{ Com}\sethlcolor{LightBlue}\hl{plet}\sethlcolor{LightBlue}\hl{ely}\sethlcolor{LightBlue}\hl{ lost}\sethlcolor{LightRed}\hl{<|endoftext|>} & 2.8743 \\\midrule
DPO Model Response & \sethlcolor{LightYellow}\hl{ I}\sethlcolor{LightRed}\hl{ recently}\sethlcolor{LightBlue}\hl{ started}\sethlcolor{LightBlue}\hl{ to}\sethlcolor{LightBlue}\hl{ have}\sethlcolor{LightBlue}\hl{ feelings}\sethlcolor{LightBlue}\hl{ for}\sethlcolor{LightYellow}\hl{ another}\sethlcolor{LightBlue}\hl{ person}\sethlcolor{LightYellow}\hl{.}\sethlcolor{LightRed}\hl{ My}\sethlcolor{LightBlue}\hl{ girlfriend}\sethlcolor{LightRed}\hl{ forced}\sethlcolor{LightBlue}\hl{ me}\sethlcolor{LightBlue}\hl{ to}\sethlcolor{LightBlue}\hl{ say}\sethlcolor{LightBlue}\hl{ what}\sethlcolor{LightBlue}\hl{ was}\sethlcolor{LightBlue}\hl{ bother}\sethlcolor{LightBlue}\hl{ing}\sethlcolor{LightBlue}\hl{ me}\sethlcolor{LightBlue}\hl{.}\sethlcolor{LightYellow}\hl{ She}\sethlcolor{LightRed}\hl{ now}\sethlcolor{LightBlue}\hl{ knows}\sethlcolor{LightBlue}\hl{.}\sethlcolor{LightYellow}\hl{ I}\sethlcolor{LightRed}\hl{ want}\sethlcolor{LightYellow}\hl{ those}\sethlcolor{LightBlue}\hl{ feelings}\sethlcolor{LightBlue}\hl{ to}\sethlcolor{LightBlue}\hl{ go}\sethlcolor{LightBlue}\hl{ away}\sethlcolor{LightBlue}\hl{,}\sethlcolor{LightBlue}\hl{ but}\sethlcolor{LightBlue}\hl{ I}\sethlcolor{LightBlue}\hl{ can}\sethlcolor{LightBlue}\hl{'t}\sethlcolor{LightBlue}\hl{.}\sethlcolor{LightYellow}\hl{ What}\sethlcolor{LightBlue}\hl{ do}\sethlcolor{LightBlue}\hl{ I}\sethlcolor{LightBlue}\hl{ do}\sethlcolor{LightBlue}\hl{?}\sethlcolor{LightRed}\hl{<|endoftext|>} & 1.354 \\\midrule
Reference response &  long relationship; fell in love with an other person; admitted it; would like it to disappear, though it doesn't.<|endoftext|> & -1.6587 \\\midrule
Base Model Response &  How do you get someone out of your head?

A:

I think you're in a situation where you need to get out of your head.  You're not in a relationship, and you're not in a relationship with someone who is a good fit for you.  You're in a relationship with someone who is not a good fit for you.  You're in a & -6.7223 \\
\bottomrule
\end{tabular}
\end{small}
\end{table}
\subsection{PPO training results}
We include several PPO's learning curves in Figure~\ref{fig:ppo-learning}. We also evaluate the final model checkpoint on the validation set of the SFT dataset using GPT as a judge against the reference summaries in Figure~\ref{fig:rlhf-scaling} (the GPT prompt can be found in Appendix~\ref{sec:gpt-judge-prompt}). We also conducted an ablation study in which we used reward whitening (\ref{detail:reward-whitening}), and the results are in Figure~\ref{fig:reward-whitening}. Finally, to help understand the correlation between summary length and win rate, we plot the win rate against the $\log(\text{summary length} / \text{reference summary length})$ at Figure~\ref{fig:log-length}.

Several observations:
\begin{enumerate}
    \item \textbf{RLHF objective goes up.} Our PPO implementation at least optimizes the RLHF objective, increasing the score total. 
    \item \textbf{Good scaling behaviors.} The preference rate of the PPO models scales nicely with the model checkpoint sizes. In particular, GPT prefers our best 6.9B model nearly 80\% of the time.
    \item \textbf{Over-optimization in 1B models.} For 1B models, the KL divergence seems high (around 50 and 85 for two runs). From an optimization point of view, there is nothing wrong with them because these two runs got higher RLHF Reward $R(x, y)$ (Equation~\ref{eq:RL}), but GPT then judges these two checkpoints to have poor human preference: less than 20\% of time GPT prefers them over reference summaries)
    \begin{itemize}
        \item Upon inspection of these overoptimized samples, we find the PPO policy would concatenate the strings like ``Mybestfriendrecentlyblockedmeinsocialmedia(atleastonce),anditreallyhurtsme(especiallyafterIwasignoredforaweek).
        Opinionsandadvicewouldbegreatlyappreciated'' (see \url{https://wandb.ai/costa-huang/tldr_summarize/runs/6qn2rlaq} as an example).
    \end{itemize}
    \item \textbf{Reward whitening makes the model generate shorter outputs.} We conducted an ablation study with and without reward whitening in Figure~\ref{fig:reward-whitening}. Our experiments show that reward whitening makes the model's completions get a lower preference rate, and the completions are shorter than those without reward whitening. However, when inspecting the length-controlled comparisons in Figure~\ref{fig:log-length} (right), the models perform similarly with or without reward whitening in different summary lengths.
    \item \textbf{PPO models significantly outperform SFT when controlling for length.} As shown in Figure~\ref{fig:log-length} (left), while PPO gets a higher win rate than SFT, the models' responses are generally longer compared to SFT responses, so the summary length is a confounding factor. To address this issue, we control for ratio of summary length to reference length in Figure~\ref{fig:log-length} (right) and show that PPO models outperform SFT models across all summary lengths. We also find that PPO win-rate increases with summary length. This implies that either GPT3.5 prefers longer summaries or longer summaries better optimize true human preference (perhaps implicitly) \citep{dubois_alpacafarm_2023}.

\end{enumerate}

\subsection{Visualizing the aligned models vs pre-trained models}
\label{sec:ai2-visual}
\citet{Lin2024ReAlign} proposed an interesting visualization regarding how aligned models would behave differently from pre-trained models. The idea is to sample a response from the aligned LLM and check if the pre-trained LLM would greedy sample the same tokens; if so, then color the text blue (unshifted tokens); if the token is within the top 3 probability, color the text yellow; else color the text red (shifted tokens). In simpler terms, the red tokens correspond to what aligned models do differently. We include such visualization of 1B models in Table \ref{tab:ai2-visual}. There are more visualizations of models in the Appendix~\ref{sec:ai2-visual-more}. Several observations:

\begin{enumerate}
    \item \textbf{Pre-trained model would continue sampling.} As a result, the generated summary would go significantly beyond the typical lengths of the reference summary or SFT / PPO / DPO summary.
    \item \textbf{Most tokens are unshifted tokens.} Similar to the findings in \citet{Lin2024ReAlign}, we find most tokens to be unshifted tokens -- this means arguably that the summarization ability mostly comes from the pre-trained model.
    \item \textbf{Fine-tuned models mostly change behaviors at the beginning and the end.} The SFT / PPO / DPO models always alter the initial output and end the summary with an EOS token.
\end{enumerate}

\section{Conclusion}
This work presents a high-fidelity reproduction of OpenAI's RLHF work in TL;DR summarization~\citep{stiennon2020learning}, demonstrating the scaling behavior of PPO across different Pythia model sizes. We offer detailed insights into the implementation specifics and design choices that enabled this successful reproduction, promoting transparency and reproducibility within the research community.

\begin{ack}
Hugging Face's cluster of H100s has fully supported this work.
\end{ack}


\newpage

{
\small

\bibliographystyle{unsrtnat}
 
\bibliography{reference}
}

\appendix

\section{List of model checkpoints and tracked logs}
\label{sec:model-checkpoints}
The list of model checkpoints and tracked logs can be found at Table~\ref{tab:checkpoint-and-logs}.

\begin{table}[]
    \centering
    \caption{List of Hugging Face model checkpoints and tracked Weights and Biases logs.}
    \label{tab:checkpoint-and-logs}
\begin{tabular}{lllll}
\toprule
 &  &  & \hflogo Model Checkpoint & Tracked Wandb Logs \\
Base Model & Type & Seed &  &  \\
\midrule
\multirow[t]{12}{*}{EleutherAI/pythia-1b-deduped} & \multirow[t]{4}{*}{ppo} & 44413 & \href{https://huggingface.co/vwxyzjn/EleutherAI\_pythia-1b-deduped\_\_ppo\_left\_padding\_new\_nowhiten\_reward\_\_tldr/tree/ppo\_left\_padding\_new\_nowhiten\_reward\_\_44413\_\_1709671965}{\hflogo Link} & \href{https://wandb.ai/costa-huang/tldr\_summarize/runs/ajthk918}{Link} \\
 &  & 55513 & \href{https://huggingface.co/vwxyzjn/EleutherAI\_pythia-1b-deduped\_\_ppo\_left\_padding\_new\_nowhiten\_reward\_\_tldr/tree/ppo\_left\_padding\_new\_nowhiten\_reward\_\_55513\_\_1709671967}{\hflogo Link} & \href{https://wandb.ai/costa-huang/tldr\_summarize/runs/pevomb70}{Link} \\
 &  & 66613 & \href{https://huggingface.co/vwxyzjn/EleutherAI\_pythia-1b-deduped\_\_ppo\_left\_padding\_new\_nowhiten\_reward\_\_tldr/tree/ppo\_left\_padding\_new\_nowhiten\_reward\_\_66613\_\_1709671965}{\hflogo Link} & \href{https://wandb.ai/costa-huang/tldr\_summarize/runs/d3xlyf1z}{Link} \\
 &  & 77713 & \href{https://huggingface.co/vwxyzjn/EleutherAI\_pythia-1b-deduped\_\_ppo\_left\_padding\_new\_nowhiten\_reward\_\_tldr/tree/ppo\_left\_padding\_new\_nowhiten\_reward\_\_77713\_\_1709671965}{\hflogo Link} & \href{https://wandb.ai/costa-huang/tldr\_summarize/runs/rabqw1p3}{Link} \\
\cline{2-5}
 & \multirow[t]{4}{*}{reward} & 44413 & \href{https://huggingface.co/vwxyzjn/EleutherAI\_pythia-1b-deduped\_\_reward\_\_tldr/tree/reward\_\_44413\_\_1708628552}{\hflogo Link} & \href{https://wandb.ai/costa-huang/tldr\_summarize/runs/z6v2q8nx}{Link} \\
 &  & 55513 & \href{https://huggingface.co/vwxyzjn/EleutherAI\_pythia-1b-deduped\_\_reward\_\_tldr/tree/reward\_\_55513\_\_1708628552}{\hflogo Link} & \href{https://wandb.ai/costa-huang/tldr\_summarize/runs/bocab5vs}{Link} \\
 &  & 66613 & \href{https://huggingface.co/vwxyzjn/EleutherAI\_pythia-1b-deduped\_\_reward\_\_tldr/tree/reward\_\_66613\_\_1708628551}{\hflogo Link} & \href{https://wandb.ai/costa-huang/tldr\_summarize/runs/s5tanswd}{Link} \\
 &  & 77713 & \href{https://huggingface.co/vwxyzjn/EleutherAI\_pythia-1b-deduped\_\_reward\_\_tldr/tree/reward\_\_77713\_\_1708628553}{\hflogo Link} & \href{https://wandb.ai/costa-huang/tldr\_summarize/runs/7q593nvh}{Link} \\
\cline{2-5}
 & \multirow[t]{4}{*}{sft} & 44413 & \href{https://huggingface.co/vwxyzjn/EleutherAI\_pythia-1b-deduped\_\_sft\_\_tldr/tree/sft\_\_44413\_\_1708611267}{\hflogo Link} & \href{https://wandb.ai/costa-huang/tldr\_summarize/runs/e9ai2b3y}{Link} \\
 &  & 55513 & \href{https://huggingface.co/vwxyzjn/EleutherAI\_pythia-1b-deduped\_\_sft\_\_tldr/tree/sft\_\_55513\_\_1708611267}{\hflogo Link} & \href{https://wandb.ai/costa-huang/tldr\_summarize/runs/1gmprrb1}{Link} \\
 &  & 66613 & \href{https://huggingface.co/vwxyzjn/EleutherAI\_pythia-1b-deduped\_\_sft\_\_tldr/tree/sft\_\_66613\_\_1708611267}{\hflogo Link} & \href{https://wandb.ai/costa-huang/tldr\_summarize/runs/t1fbmibv}{Link} \\
 &  & 77713 & \href{https://huggingface.co/vwxyzjn/EleutherAI\_pythia-1b-deduped\_\_sft\_\_tldr/tree/sft\_\_77713\_\_1708611267}{\hflogo Link} & \href{https://wandb.ai/costa-huang/tldr\_summarize/runs/7mpo5c4s}{Link} \\
\cline{1-5} \cline{2-5}
\multirow[t]{12}{*}{EleutherAI/pythia-2.8b-deduped} & \multirow[t]{4}{*}{ppo} & 44413 & \href{https://huggingface.co/vwxyzjn/EleutherAI\_pythia-2.8b-deduped\_\_ppo\_left\_padding\_new\_nowhiten\_reward\_\_tldr/tree/ppo\_left\_padding\_new\_nowhiten\_reward\_\_44413\_\_1710356835}{\hflogo Link} & \href{https://wandb.ai/costa-huang/tldr\_summarize/runs/cfmzft10}{Link} \\
 &  & 55513 & \href{https://huggingface.co/vwxyzjn/EleutherAI\_pythia-2.8b-deduped\_\_ppo\_left\_padding\_new\_nowhiten\_reward\_\_tldr/tree/ppo\_left\_padding\_new\_nowhiten\_reward\_\_55513\_\_1710356835}{\hflogo Link} & \href{https://wandb.ai/costa-huang/tldr\_summarize/runs/9f6t868e}{Link} \\
 &  & 66613 & \href{https://huggingface.co/vwxyzjn/EleutherAI\_pythia-2.8b-deduped\_\_ppo\_left\_padding\_new\_nowhiten\_reward\_\_tldr/tree/ppo\_left\_padding\_new\_nowhiten\_reward\_\_66613\_\_1710356835}{\hflogo Link} & \href{https://wandb.ai/costa-huang/tldr\_summarize/runs/8sr72pr9}{Link} \\
 &  & 77713 & \href{https://huggingface.co/vwxyzjn/EleutherAI\_pythia-2.8b-deduped\_\_ppo\_left\_padding\_new\_nowhiten\_reward\_\_tldr/tree/ppo\_left\_padding\_new\_nowhiten\_reward\_\_77713\_\_1710356835}{\hflogo Link} & \href{https://wandb.ai/costa-huang/tldr\_summarize/runs/4ao2gn7n}{Link} \\
\cline{2-5}
 & \multirow[t]{4}{*}{reward} & 44413 & \href{https://huggingface.co/vwxyzjn/EleutherAI\_pythia-2.8b-deduped\_\_reward\_\_tldr/tree/reward\_\_44413\_\_1708628552}{\hflogo Link} & \href{https://wandb.ai/costa-huang/tldr\_summarize/runs/5316gkrt}{Link} \\
 &  & 55513 & \href{https://huggingface.co/vwxyzjn/EleutherAI\_pythia-2.8b-deduped\_\_reward\_\_tldr/tree/reward\_\_55513\_\_1708628552}{\hflogo Link} & \href{https://wandb.ai/costa-huang/tldr\_summarize/runs/glga10zf}{Link} \\
 &  & 66613 & \href{https://huggingface.co/vwxyzjn/EleutherAI\_pythia-2.8b-deduped\_\_reward\_\_tldr/tree/reward\_\_66613\_\_1708628551}{\hflogo Link} & \href{https://wandb.ai/costa-huang/tldr\_summarize/runs/8yh2ns3p}{Link} \\
 &  & 77713 & \href{https://huggingface.co/vwxyzjn/EleutherAI\_pythia-2.8b-deduped\_\_reward\_\_tldr/tree/reward\_\_77713\_\_1708628552}{\hflogo Link} & \href{https://wandb.ai/costa-huang/tldr\_summarize/runs/m94qlmen}{Link} \\
\cline{2-5}
 & \multirow[t]{4}{*}{sft} & 44413 & \href{https://huggingface.co/vwxyzjn/EleutherAI\_pythia-2.8b-deduped\_\_sft\_\_tldr/tree/sft\_\_44413\_\_1708611267}{\hflogo Link} & \href{https://wandb.ai/costa-huang/tldr\_summarize/runs/4ocqe2yu}{Link} \\
 &  & 55513 & \href{https://huggingface.co/vwxyzjn/EleutherAI\_pythia-2.8b-deduped\_\_sft\_\_tldr/tree/sft\_\_55513\_\_1708611267}{\hflogo Link} & \href{https://wandb.ai/costa-huang/tldr\_summarize/runs/blt9zq7r}{Link} \\
 &  & 66613 & \href{https://huggingface.co/vwxyzjn/EleutherAI\_pythia-2.8b-deduped\_\_sft\_\_tldr/tree/sft\_\_66613\_\_1708611267}{\hflogo Link} & \href{https://wandb.ai/costa-huang/tldr\_summarize/runs/m78f3l3l}{Link} \\
 &  & 77713 & \href{https://huggingface.co/vwxyzjn/EleutherAI\_pythia-2.8b-deduped\_\_sft\_\_tldr/tree/sft\_\_77713\_\_1708611267}{\hflogo Link} & \href{https://wandb.ai/costa-huang/tldr\_summarize/runs/dvemfw7l}{Link} \\
\cline{1-5} \cline{2-5}
\multirow[t]{12}{*}{EleutherAI/pythia-6.9b-deduped} & \multirow[t]{4}{*}{ppo} & 44413 & \href{https://huggingface.co/vwxyzjn/EleutherAI\_pythia-6.9b-deduped\_\_ppo\_left\_padding\_new\_nowhiten\_reward\_\_tldr/tree/ppo\_left\_padding\_new\_nowhiten\_reward\_\_44413\_\_1710465193}{\hflogo Link} & \href{https://wandb.ai/costa-huang/tldr\_summarize/runs/1lpwnykt}{Link} \\
 &  & 55513 & \href{https://huggingface.co/vwxyzjn/EleutherAI\_pythia-6.9b-deduped\_\_ppo\_left\_padding\_new\_nowhiten\_reward\_\_tldr/tree/ppo\_left\_padding\_new\_nowhiten\_reward\_\_55513\_\_1710465193}{\hflogo Link} & \href{https://wandb.ai/costa-huang/tldr\_summarize/runs/vou23bja}{Link} \\
 &  & 66613 & \href{https://huggingface.co/vwxyzjn/EleutherAI\_pythia-6.9b-deduped\_\_ppo\_left\_padding\_new\_nowhiten\_reward\_\_tldr/tree/ppo\_left\_padding\_new\_nowhiten\_reward\_\_66613\_\_1710465193}{\hflogo Link} & \href{https://wandb.ai/costa-huang/tldr\_summarize/runs/kvdfazmz}{Link} \\
 &  & 77713 & \href{https://huggingface.co/vwxyzjn/EleutherAI\_pythia-6.9b-deduped\_\_ppo\_left\_padding\_new\_nowhiten\_reward\_\_tldr/tree/ppo\_left\_padding\_new\_nowhiten\_reward\_\_77713\_\_1710465193}{\hflogo Link} & \href{https://wandb.ai/costa-huang/tldr\_summarize/runs/g6co0hel}{Link} \\
\cline{2-5}
 & \multirow[t]{4}{*}{reward} & 44413 & \href{https://huggingface.co/vwxyzjn/EleutherAI\_pythia-6.9b-deduped\_\_reward\_\_tldr/tree/reward\_\_44413\_\_1708628552}{\hflogo Link} & \href{https://wandb.ai/costa-huang/tldr\_summarize/runs/rqymy36n}{Link} \\
 &  & 55513 & \href{https://huggingface.co/vwxyzjn/EleutherAI\_pythia-6.9b-deduped\_\_reward\_\_tldr/tree/reward\_\_55513\_\_1708628552}{\hflogo Link} & \href{https://wandb.ai/costa-huang/tldr\_summarize/runs/gto3imru}{Link} \\
 &  & 66613 & \href{https://huggingface.co/vwxyzjn/EleutherAI\_pythia-6.9b-deduped\_\_reward\_\_tldr/tree/reward\_\_66613\_\_1708628552}{\hflogo Link} & \href{https://wandb.ai/costa-huang/tldr\_summarize/runs/isj98l75}{Link} \\
 &  & 77713 & \href{https://huggingface.co/vwxyzjn/EleutherAI\_pythia-6.9b-deduped\_\_reward\_\_tldr/tree/reward\_\_77713\_\_1708628551}{\hflogo Link} & \href{https://wandb.ai/costa-huang/tldr\_summarize/runs/d36backr}{Link} \\
\cline{2-5}
 & \multirow[t]{4}{*}{sft} & 44413 & \href{https://huggingface.co/vwxyzjn/EleutherAI\_pythia-6.9b-deduped\_\_sft\_\_tldr/tree/sft\_\_44413\_\_1708611267}{\hflogo Link} & \href{https://wandb.ai/costa-huang/tldr\_summarize/runs/vylgk7cg}{Link} \\
 &  & 55513 & \href{https://huggingface.co/vwxyzjn/EleutherAI\_pythia-6.9b-deduped\_\_sft\_\_tldr/tree/sft\_\_55513\_\_1708611267}{\hflogo Link} & \href{https://wandb.ai/costa-huang/tldr\_summarize/runs/o6u0ha8h}{Link} \\
 &  & 66613 & \href{https://huggingface.co/vwxyzjn/EleutherAI\_pythia-6.9b-deduped\_\_sft\_\_tldr/tree/sft\_\_66613\_\_1708611267}{\hflogo Link} & \href{https://wandb.ai/costa-huang/tldr\_summarize/runs/ijy11svk}{Link} \\
 &  & 77713 & \href{https://huggingface.co/vwxyzjn/EleutherAI\_pythia-6.9b-deduped\_\_sft\_\_tldr/tree/sft\_\_77713\_\_1708611267}{\hflogo Link} & \href{https://wandb.ai/costa-huang/tldr\_summarize/runs/1jjm2vga}{Link} \\
\cline{1-5} \cline{2-5}
\bottomrule
\end{tabular}
\end{table}

\section{GPT as a judge prompt}
\label{sec:gpt-judge-prompt}

We modify the GPT as a judge prompt from \citet{rafailov2023direct}.

\begin{verbatim}
Which of the following summaries does a better job of summarizing the most \ 
important points in the given forum post, without including unimportant or \ 
irrelevant details? Judge based on accuracy, coverage, and coherence.

Post:
<post>

Summary A:
<Summary A>

Summary B:
<Summary B>

FIRST provide a one-sentence comparison of the two summaries, explaining which \
you prefer and why. SECOND, on a new line, state only "A" or "B" to indicate your \ 
choice. Your response should use the format:
Comparison: <one-sentence comparison and explanation>
Preferred: <"A" or "B">
\end{verbatim}

Following \citet{wang_large_2023,zheng2024judging} we randomize the order of the summaries to remove positional bias in GPT-3.5 Turbo.

\section{Model win rate versus summary lengths}
\label{sec:full-log-length}
\begin{figure}[t]
    \centering
    \includegraphics[width=0.99\linewidth]{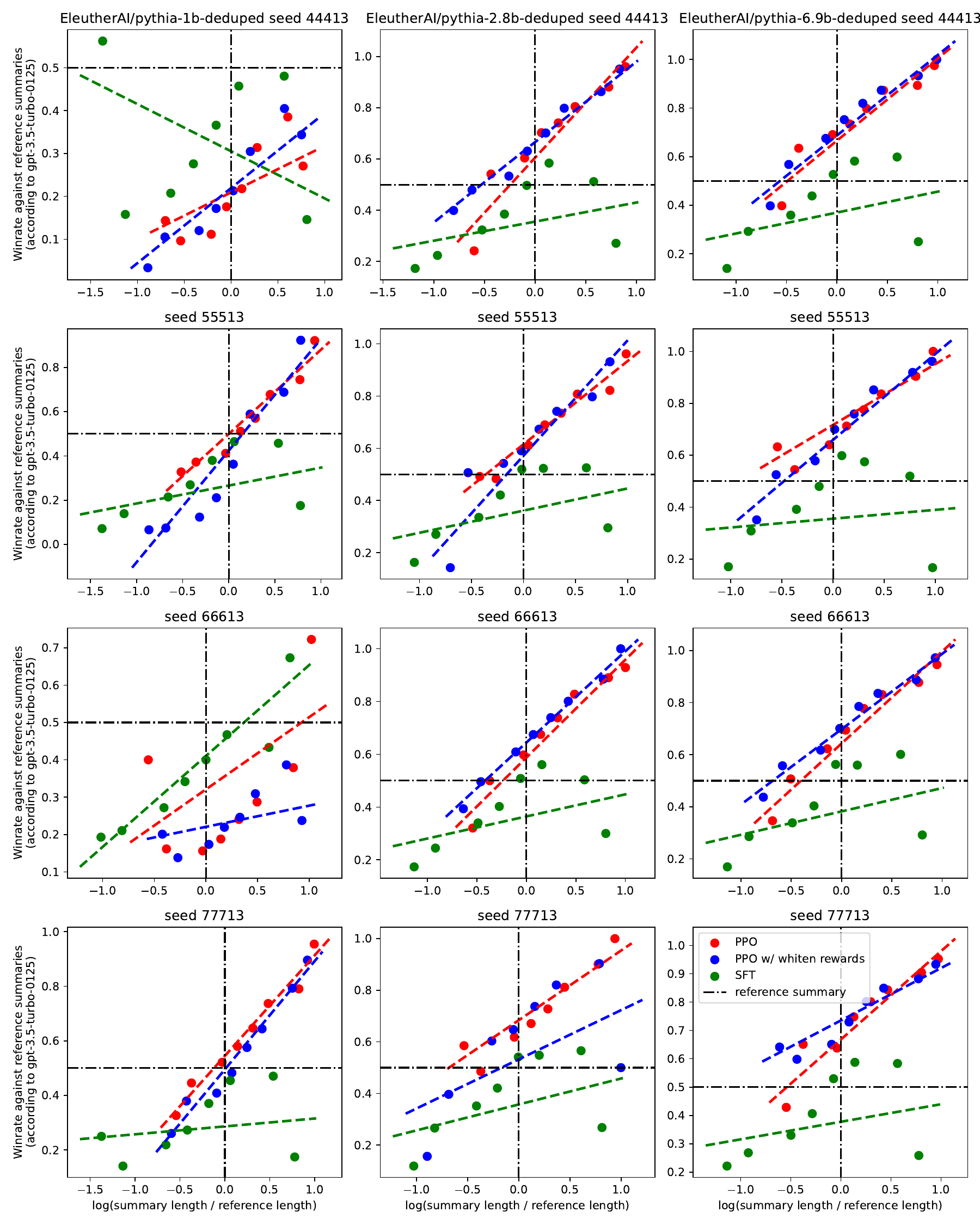}
    \caption{The figure shows the win rate for different summary lengths of the SFT and PPO models across random seeds and model sizes. Note that some of the PPO seeds corresponds to overoptimized policies.}
    \label{fig:full-log-length}
\end{figure}

Figure~\ref{fig:full-log-length} show more plots like Figure~\ref{fig:log-length} (right).

\section{Visualization on aligned models vs base models}
\label{sec:ai2-visual-more}
This section generates more samples for different-sized models like Table~\ref{tab:ai2-visual}. 
\begin{enumerate}
    \item \textbf{1B model responses} in Table~\ref{tab:ai2-visual-1b-2}, Table~\ref{tab:ai2-visual-1b-3}, and Table~\ref{tab:ai2-visual-1b-4}
    \item \textbf{2.8B model responses} in Table~\ref{tab:ai2-visual-2.8b-1}, Table~\ref{tab:ai2-visual-2.8b-2}, Table~\ref{tab:ai2-visual-1b-3}, and Table~\ref{tab:ai2-visual-1b-4}
    \item \textbf{6.9B model responses} in Table~\ref{tab:ai2-visual-6.9b-1}, Table~\ref{tab:ai2-visual-6.9b-2}, Table~\ref{tab:ai2-visual-6.9b-3}, and Table~\ref{tab:ai2-visual-6.9b-4}
\end{enumerate}

\begin{table}[t]
\caption{Sample query, responses from the 1B SFT, PPO, and DPO models; scores are from a 6.9B model. See Table~\ref{tab:ai2-visual} and Section~\ref{tab:ai2-visual} for semantics on colors.}
\label{tab:ai2-visual-1b-2}
\centering
\begin{small}
    \begin{tabular}{lp{7cm}l}
\toprule
Type & Content & Score (RM) \\
\midrule
Query & SUBREDDIT: r/pettyrevenge

TITLE: So, my mom woke me up with a loud TV.

POST: She was in her living room, watching TV. This was at about 8:30 in the morning, and she was exercising. She turned the TV up extra loud to hear it over her excercycle, and woke me up. I went in there asking for her to turn it down. She said she didn't have to; I explained that I always used headphones so she didn't have to deal with my noise and that she should give me a little more respect, given that I paid rent at the time.

She disagreed. I went back to my room, rather pissed off at the lack of equality. I had no lock on my door; but I had a dresser right next to it, so I pulled one of the drawers out enough so that it caused the door to not be openable. Then, I turned my speakers up really loud and blasted Gangnam Style on repeat, with the bass cranked up as high as it could go.

If you hate Gangnam Style for being overplayed, you will see why I chose that particular song. I personally don't mind it. But here's the thing about my bass; it vibrates the walls, making one hell of a lot of noise. Needless to say, my mom was not pleased and shut off the internet. But it was oh so worth it.

TL;DR: & N/A \\\midrule
SFT Model Response & \sethlcolor{LightRed}\hl{ Mom}\sethlcolor{LightBlue}\hl{ woke}\sethlcolor{LightBlue}\hl{ me}\sethlcolor{LightBlue}\hl{ up}\sethlcolor{LightBlue}\hl{ with}\sethlcolor{LightBlue}\hl{ a}\sethlcolor{LightBlue}\hl{ loud}\sethlcolor{LightBlue}\hl{ TV}\sethlcolor{LightBlue}\hl{.}\sethlcolor{LightBlue}\hl{ I}\sethlcolor{LightYellow}\hl{ turned}\sethlcolor{LightBlue}\hl{ it}\sethlcolor{LightYellow}\hl{ down}\sethlcolor{LightYellow}\hl{,}\sethlcolor{LightBlue}\hl{ and}\sethlcolor{LightBlue}\hl{ she}\sethlcolor{LightRed}\hl{ got}\sethlcolor{LightBlue}\hl{ pissed}\sethlcolor{LightBlue}\hl{.}\sethlcolor{LightYellow}\hl{ I}\sethlcolor{LightRed}\hl{ turned}\sethlcolor{LightBlue}\hl{ it}\sethlcolor{LightBlue}\hl{ up}\sethlcolor{LightBlue}\hl{,}\sethlcolor{LightBlue}\hl{ and}\sethlcolor{LightBlue}\hl{ she}\sethlcolor{LightBlue}\hl{ got}\sethlcolor{LightBlue}\hl{ pissed}\sethlcolor{LightBlue}\hl{.}\sethlcolor{LightRed}\hl{<|endoftext|>} & -3.9943 \\\midrule
PPO Model Response & \sethlcolor{LightRed}\hl{ While}\sethlcolor{LightRed}\hl{ watching}\sethlcolor{LightBlue}\hl{ TV}\sethlcolor{LightRed}\hl{ my}\sethlcolor{LightBlue}\hl{ mom}\sethlcolor{LightBlue}\hl{ woke}\sethlcolor{LightBlue}\hl{ me}\sethlcolor{LightBlue}\hl{ up}\sethlcolor{LightBlue}\hl{ with}\sethlcolor{LightBlue}\hl{ a}\sethlcolor{LightBlue}\hl{ loud}\sethlcolor{LightRed}\hl{ racket}\sethlcolor{LightRed}\hl{ that}\sethlcolor{LightRed}\hl{ vibr}\sethlcolor{LightBlue}\hl{ated}\sethlcolor{LightBlue}\hl{ the}\sethlcolor{LightBlue}\hl{ walls}\sethlcolor{LightRed}\hl{ making}\sethlcolor{LightRed}\hl{ one}\sethlcolor{LightBlue}\hl{ hell}\sethlcolor{LightBlue}\hl{ of}\sethlcolor{LightBlue}\hl{ a}\sethlcolor{LightBlue}\hl{ racket}\sethlcolor{LightYellow}\hl{,}\sethlcolor{LightRed}\hl{ resulting}\sethlcolor{LightBlue}\hl{ in}\sethlcolor{LightRed}\hl{ her}\sethlcolor{LightBlue}\hl{ shutting}\sethlcolor{LightBlue}\hl{ off}\sethlcolor{LightBlue}\hl{ the}\sethlcolor{LightBlue}\hl{ internet}\sethlcolor{LightBlue}\hl{.}\sethlcolor{LightRed}\hl{ Very}\sethlcolor{LightBlue}\hl{ satisfying}\sethlcolor{LightRed}\hl{ aftermath}\sethlcolor{LightRed}\hl{<|endoftext|>} & 1.3313 \\\midrule
DPO Model Response & \sethlcolor{LightYellow}\hl{ My}\sethlcolor{LightBlue}\hl{ mom}\sethlcolor{LightYellow}\hl{ was}\sethlcolor{LightRed}\hl{ annoyed}\sethlcolor{LightRed}\hl{ at}\sethlcolor{LightRed}\hl{ an}\sethlcolor{LightYellow}\hl{ over}\sethlcolor{LightYellow}\hl{-}\sethlcolor{LightBlue}\hl{played}\sethlcolor{LightRed}\hl{ Gang}\sethlcolor{LightBlue}\hl{nam}\sethlcolor{LightBlue}\hl{ Style}\sethlcolor{LightBlue}\hl{ song}\sethlcolor{LightRed}\hl{ on}\sethlcolor{LightBlue}\hl{ the}\sethlcolor{LightRed}\hl{ morning}\sethlcolor{LightRed}\hl{ TV}\sethlcolor{LightRed}\hl{ wake}\sethlcolor{LightBlue}\hl{-}\sethlcolor{LightBlue}\hl{up}\sethlcolor{LightBlue}\hl{ call}\sethlcolor{LightYellow}\hl{,}\sethlcolor{LightBlue}\hl{ and}\sethlcolor{LightRed}\hl{ shut}\sethlcolor{LightYellow}\hl{ it}\sethlcolor{LightBlue}\hl{ off}\sethlcolor{LightBlue}\hl{.}\sethlcolor{LightYellow}\hl{ It}\sethlcolor{LightBlue}\hl{ was}\sethlcolor{LightBlue}\hl{ worth}\sethlcolor{LightBlue}\hl{ it}\sethlcolor{LightBlue}\hl{.}\sethlcolor{LightRed}\hl{<|endoftext|>} & -0.6789 \\\midrule
Reference response &  Mom had the TV on loud and woke me up, didn't care that I'd respected audio levels in the house, so I countered with playing Gangnam Style on repeat with the bass thumping through the walls.<|endoftext|> & 1.3359 \\\midrule
Base Model Response &  I was pissed off at my mom for not giving me more respect, and I was pissed off at her for shutting off the internet.

I'm not sure if I'm the only one who has this problem, but I've been having a problem with my computer for a while now. I've been using it for work, and I've been using it for school. I've been using & -7.4169 \\
\bottomrule
\end{tabular}
\end{small}
\end{table}

\begin{table}[t]
\caption{Sample query, responses from the 1B SFT, PPO, and DPO models; scores are from a 6.9B model. See Table~\ref{tab:ai2-visual} and Section~\ref{tab:ai2-visual} for semantics on colors.}
\label{tab:ai2-visual-1b-3}
\centering
\begin{small}
    \begin{tabular}{lp{7cm}l}
\toprule
Type & Content & Score (RM) \\
\midrule
Query & SUBREDDIT: r/relationships

TITLE: My girlfriend (20f) of two years cheated on me (20m) by kissing two guys at a Halloween party.

POST: Lately her and I have been having a few problems, and these problems have been brought up before a few times. One problem being that I don't show enough affection. I don't tell her she's pretty very often or don't compliment her much. I feel terrible about it, but this time I was really trying to change for her.

For Halloween she went to visit her step brother at a college and I got drunk with my friends and watched movies. Last night (11/1) we got in a huge fight about me not changing and how our relationship won't work out and basically broke up over the phone. So in an effort to try and fix it I drove to her house. She told me how at the parties she went to that two guys kissed her. The first one she pushed away, but the second one I asked her if she kissed him back and she said yes and that she did it because it made her feel wanted, which I guess I haven't been making her feel that way lately. We cried, we talked about everything, we had great sex, and I stayed over at her house just to sleep with her and then snuck out in the morning so her parents wouldn't know.

We both obviously want to work things out but aren't sure if we should. I love this girl, but the more I think about it, all I can think about is her cheating on me, and more importantly, liking it. It makes me sick to my stomach. Should I even try to fix it or would I be better off cutting all ties.

TL;DR: & N/A \\\midrule
SFT Model Response & \sethlcolor{LightRed}\hl{ Girl}\sethlcolor{LightBlue}\hl{friend}\sethlcolor{LightBlue}\hl{ che}\sethlcolor{LightBlue}\hl{ated}\sethlcolor{LightBlue}\hl{ on}\sethlcolor{LightBlue}\hl{ me}\sethlcolor{LightYellow}\hl{ by}\sethlcolor{LightBlue}\hl{ kissing}\sethlcolor{LightBlue}\hl{ two}\sethlcolor{LightBlue}\hl{ guys}\sethlcolor{LightBlue}\hl{ at}\sethlcolor{LightBlue}\hl{ a}\sethlcolor{LightBlue}\hl{ Halloween}\sethlcolor{LightBlue}\hl{ party}\sethlcolor{LightBlue}\hl{.}\sethlcolor{LightYellow}\hl{ I}\sethlcolor{LightYellow}\hl{ don}\sethlcolor{LightBlue}\hl{'t}\sethlcolor{LightBlue}\hl{ know}\sethlcolor{LightBlue}\hl{ if}\sethlcolor{LightBlue}\hl{ I}\sethlcolor{LightBlue}\hl{ should}\sethlcolor{LightYellow}\hl{ try}\sethlcolor{LightBlue}\hl{ to}\sethlcolor{LightBlue}\hl{ fix}\sethlcolor{LightBlue}\hl{ it}\sethlcolor{LightBlue}\hl{ or}\sethlcolor{LightYellow}\hl{ cut}\sethlcolor{LightYellow}\hl{ all}\sethlcolor{LightBlue}\hl{ ties}\sethlcolor{LightBlue}\hl{.}\sethlcolor{LightRed}\hl{<|endoftext|>} & 1.6686 \\\midrule
PPO Model Response & \sethlcolor{LightYellow}\hl{ My}\sethlcolor{LightBlue}\hl{ girlfriend}\sethlcolor{LightYellow}\hl{ of}\sethlcolor{LightBlue}\hl{ two}\sethlcolor{LightBlue}\hl{ years}\sethlcolor{LightBlue}\hl{ che}\sethlcolor{LightBlue}\hl{ated}\sethlcolor{LightBlue}\hl{ on}\sethlcolor{LightBlue}\hl{ me}\sethlcolor{LightBlue}\hl{ by}\sethlcolor{LightBlue}\hl{ kissing}\sethlcolor{LightBlue}\hl{ two}\sethlcolor{LightBlue}\hl{ guys}\sethlcolor{LightBlue}\hl{ at}\sethlcolor{LightBlue}\hl{ a}\sethlcolor{LightBlue}\hl{ Halloween}\sethlcolor{LightBlue}\hl{ party}\sethlcolor{LightBlue}\hl{.}\sethlcolor{LightYellow}\hl{ I}\sethlcolor{LightRed}\hl{ love}\sethlcolor{LightBlue}\hl{ her}\sethlcolor{LightYellow}\hl{ but}\sethlcolor{LightYellow}\hl{ the}\sethlcolor{LightBlue}\hl{ more}\sethlcolor{LightBlue}\hl{ I}\sethlcolor{LightBlue}\hl{ think}\sethlcolor{LightBlue}\hl{ about}\sethlcolor{LightBlue}\hl{ it}\sethlcolor{LightYellow}\hl{ the}\sethlcolor{LightBlue}\hl{ more}\sethlcolor{LightBlue}\hl{ I}\sethlcolor{LightBlue}\hl{ feel}\sethlcolor{LightYellow}\hl{ sick}\sethlcolor{LightYellow}\hl{ about}\sethlcolor{LightYellow}\hl{ the}\sethlcolor{LightYellow}\hl{ whole}\sethlcolor{LightBlue}\hl{ situation}\sethlcolor{LightBlue}\hl{.}\sethlcolor{LightYellow}\hl{ Should}\sethlcolor{LightBlue}\hl{ I}\sethlcolor{LightBlue}\hl{ try}\sethlcolor{LightBlue}\hl{ to}\sethlcolor{LightBlue}\hl{ fix}\sethlcolor{LightBlue}\hl{ it}\sethlcolor{LightBlue}\hl{ or}\sethlcolor{LightYellow}\hl{ cut}\sethlcolor{LightYellow}\hl{ all}\sethlcolor{LightBlue}\hl{ ties}\sethlcolor{LightBlue}\hl{?}\sethlcolor{LightRed}\hl{<|endoftext|>} & 3.7295 \\\midrule
DPO Model Response & \sethlcolor{LightYellow}\hl{ My}\sethlcolor{LightBlue}\hl{ girlfriend}\sethlcolor{LightYellow}\hl{ (}\sethlcolor{LightBlue}\hl{20}\sethlcolor{LightBlue}\hl{f}\sethlcolor{LightBlue}\hl{)}\sethlcolor{LightYellow}\hl{ che}\sethlcolor{LightBlue}\hl{ated}\sethlcolor{LightBlue}\hl{ on}\sethlcolor{LightBlue}\hl{ me}\sethlcolor{LightBlue}\hl{ (}\sethlcolor{LightBlue}\hl{20}\sethlcolor{LightBlue}\hl{m}\sethlcolor{LightBlue}\hl{)}\sethlcolor{LightBlue}\hl{ by}\sethlcolor{LightBlue}\hl{ kissing}\sethlcolor{LightRed}\hl{ at}\sethlcolor{LightBlue}\hl{ a}\sethlcolor{LightBlue}\hl{ Halloween}\sethlcolor{LightBlue}\hl{ party}\sethlcolor{LightYellow}\hl{ and}\sethlcolor{LightRed}\hl{ we}\sethlcolor{LightRed}\hl{ cried}\sethlcolor{LightBlue}\hl{ and}\sethlcolor{LightBlue}\hl{ talked}\sethlcolor{LightBlue}\hl{ about}\sethlcolor{LightBlue}\hl{ it}\sethlcolor{LightBlue}\hl{.}\sethlcolor{LightRed}\hl{ Should}\sethlcolor{LightBlue}\hl{ I}\sethlcolor{LightBlue}\hl{ try}\sethlcolor{LightBlue}\hl{ to}\sethlcolor{LightBlue}\hl{ fix}\sethlcolor{LightBlue}\hl{ it}\sethlcolor{LightBlue}\hl{ or}\sethlcolor{LightBlue}\hl{ should}\sethlcolor{LightBlue}\hl{ I}\sethlcolor{LightBlue}\hl{ just}\sethlcolor{LightBlue}\hl{ cut}\sethlcolor{LightBlue}\hl{ all}\sethlcolor{LightBlue}\hl{ ties}\sethlcolor{LightBlue}\hl{?}\sethlcolor{LightRed}\hl{<|endoftext|>} & 2.7493 \\\midrule
Reference response &  My girlfriend kissed two guys at a Halloween party and liked it. We love each other but don't know what to do about it.<|endoftext|> & 0.1145 \\\midrule
Base Model Response &  I'm not sure if I should try to fix it or cut ties.

I'm not sure if I should try to fix it or cut ties.

I'm not sure if I should try to fix it or cut ties.

I'm not sure if I should try to fix it or cut ties.

I'm not sure if I should try to fix it or & -5.7284 \\
\bottomrule
\end{tabular}
\end{small}
\end{table}

\begin{table}[t]
\caption{Sample query, responses from the 1B SFT, PPO, and DPO models; scores are from a 6.9B model. See Table~\ref{tab:ai2-visual} and Section~\ref{tab:ai2-visual} for semantics on colors.}
\label{tab:ai2-visual-1b-4}
\centering
\begin{small}
    \begin{tabular}{lp{7cm}l}
\toprule
Type & Content & Score (RM) \\
\midrule
Query & SUBREDDIT: r/running

TITLE: One year post-pregnancy, aches and pains in hips and knees, can't even comfortably do day 1 of C25k. Help?

POST: I'll try to keep this short!

**Background**

* I've always been an on again/off again (very casual!) jogger, typically doing 3 - 5 k
* My knees have always been finicky, and I went to a physio who thought I had "runner's knee"
* Pre-pregnancy, my "runner's knee" would flare up when I got to the 8 - 10 k distance range, even if I had a decent base (doing a C210k type program)

**Current Problem**

I had my baby a year ago, so all in all I haven't run for about 1.5 years.  I'm quite slim and have been doing aerobics-style classes for the past year, so I'm not totally out of shape.  Body weight exercises, aerobics, bikes and ellipticals are all fine.  However, when I run even the tinyiest bit, or even go on a long walk or a hike, my pelvis gets very sore and tight, and my knees start hurting very quickly.  I already am doing general squats/lunges/stretching type things.  

I'm starting to feel like running just isn't for me anymore. Which is a bummer, because I really enjoy running!

Has anyone had something similar? Can anyone recommend some stretches or exercises that might help?  Should I see a Dr?  Or should I just see a physio?  Not quite sure how to proceed.

Thanks!

TL;DR: & N/A \\\midrule
SFT Model Response & \sethlcolor{LightYellow}\hl{ I}\sethlcolor{LightBlue}\hl{ have}\sethlcolor{LightBlue}\hl{ a}\sethlcolor{LightBlue}\hl{ very}\sethlcolor{LightBlue}\hl{ tight}\sethlcolor{LightBlue}\hl{ pelvis}\sethlcolor{LightYellow}\hl{ and}\sethlcolor{LightBlue}\hl{ knees}\sethlcolor{LightBlue}\hl{,}\sethlcolor{LightBlue}\hl{ and}\sethlcolor{LightBlue}\hl{ I}\sethlcolor{LightYellow}\hl{ can}\sethlcolor{LightBlue}\hl{'t}\sethlcolor{LightYellow}\hl{ run}\sethlcolor{LightBlue}\hl{ anymore}\sethlcolor{LightBlue}\hl{.}\sethlcolor{LightYellow}\hl{  }\sethlcolor{LightBlue}\hl{I}\sethlcolor{LightBlue}\hl{'m}\sethlcolor{LightRed}\hl{ starting}\sethlcolor{LightBlue}\hl{ to}\sethlcolor{LightBlue}\hl{ feel}\sethlcolor{LightBlue}\hl{ like}\sethlcolor{LightBlue}\hl{ I}\sethlcolor{LightBlue}\hl{'m}\sethlcolor{LightBlue}\hl{ not}\sethlcolor{LightBlue}\hl{ running}\sethlcolor{LightBlue}\hl{ anymore}\sethlcolor{LightBlue}\hl{.}\sethlcolor{LightYellow}\hl{  }\sethlcolor{LightYellow}\hl{Should}\sethlcolor{LightBlue}\hl{ I}\sethlcolor{LightBlue}\hl{ see}\sethlcolor{LightBlue}\hl{ a}\sethlcolor{LightBlue}\hl{ Dr}\sethlcolor{LightBlue}\hl{?}\sethlcolor{LightBlue}\hl{  }\sethlcolor{LightBlue}\hl{Or}\sethlcolor{LightBlue}\hl{ should}\sethlcolor{LightBlue}\hl{ I}\sethlcolor{LightBlue}\hl{ just}\sethlcolor{LightBlue}\hl{ see}\sethlcolor{LightBlue}\hl{ a}\sethlcolor{LightBlue}\hl{ phys}\sethlcolor{LightBlue}\hl{io}\sethlcolor{LightBlue}\hl{?}\sethlcolor{LightRed}\hl{<|endoftext|>} & 0.2434 \\\midrule
PPO Model Response & \sethlcolor{LightRed}\hl{ One}\sethlcolor{LightBlue}\hl{ year}\sethlcolor{LightBlue}\hl{ post}\sethlcolor{LightBlue}\hl{-}\sethlcolor{LightBlue}\hl{p}\sethlcolor{LightBlue}\hl{regnancy}\sethlcolor{LightBlue}\hl{,}\sethlcolor{LightRed}\hl{ my}\sethlcolor{LightBlue}\hl{ knees}\sethlcolor{LightBlue}\hl{ hurt}\sethlcolor{LightYellow}\hl{ when}\sethlcolor{LightBlue}\hl{ I}\sethlcolor{LightBlue}\hl{ run}\sethlcolor{LightBlue}\hl{,}\sethlcolor{LightRed}\hl{ can}\sethlcolor{LightBlue}\hl{'t}\sethlcolor{LightBlue}\hl{ even}\sethlcolor{LightBlue}\hl{ comfortably}\sethlcolor{LightBlue}\hl{ do}\sethlcolor{LightBlue}\hl{ day}\sethlcolor{LightBlue}\hl{ 1}\sethlcolor{LightBlue}\hl{ of}\sethlcolor{LightBlue}\hl{ C}\sethlcolor{LightBlue}\hl{25}\sethlcolor{LightBlue}\hl{k}\sethlcolor{LightBlue}\hl{.}\sethlcolor{LightYellow}\hl{  }\sethlcolor{LightRed}\hl{Has}\sethlcolor{LightBlue}\hl{ anyone}\sethlcolor{LightBlue}\hl{ had}\sethlcolor{LightBlue}\hl{ something}\sethlcolor{LightBlue}\hl{ similar}\sethlcolor{LightRed}\hl{ and}\sethlcolor{LightRed}\hl{ what}\sethlcolor{LightRed}\hl{ exercises}\sethlcolor{LightBlue}\hl{ or}\sethlcolor{LightBlue}\hl{ stretches}\sethlcolor{LightBlue}\hl{ might}\sethlcolor{LightBlue}\hl{ help}\sethlcolor{LightRed}\hl{ (}\sethlcolor{LightRed}\hl{pref}\sethlcolor{LightBlue}\hl{erably}\sethlcolor{LightYellow}\hl{ with}\sethlcolor{LightBlue}\hl{ a}\sethlcolor{LightYellow}\hl{ Dr}\sethlcolor{LightBlue}\hl{)?}\sethlcolor{LightRed}\hl{<|endoftext|>} & 4.8225 \\\midrule
DPO Model Response & \sethlcolor{LightYellow}\hl{ I}\sethlcolor{LightRed}\hl{ had}\sethlcolor{LightYellow}\hl{ my}\sethlcolor{LightRed}\hl{ pregnancy}\sethlcolor{LightRed}\hl{ year}\sethlcolor{LightBlue}\hl{ ago}\sethlcolor{LightYellow}\hl{ and}\sethlcolor{LightRed}\hl{ haven}\sethlcolor{LightBlue}\hl{'t}\sethlcolor{LightBlue}\hl{ run}\sethlcolor{LightBlue}\hl{ for}\sethlcolor{LightBlue}\hl{ about}\sethlcolor{LightYellow}\hl{ 1}\sethlcolor{LightBlue}\hl{.}\sethlcolor{LightBlue}\hl{5}\sethlcolor{LightBlue}\hl{ years}\sethlcolor{LightYellow}\hl{,}\sethlcolor{LightYellow}\hl{ but}\sethlcolor{LightRed}\hl{ when}\sethlcolor{LightBlue}\hl{ I}\sethlcolor{LightBlue}\hl{ run}\sethlcolor{LightRed}\hl{ my}\sethlcolor{LightBlue}\hl{ pelvis}\sethlcolor{LightBlue}\hl{ gets}\sethlcolor{LightYellow}\hl{ sore}\sethlcolor{LightBlue}\hl{ and}\sethlcolor{LightBlue}\hl{ tight}\sethlcolor{LightBlue}\hl{,}\sethlcolor{LightBlue}\hl{ and}\sethlcolor{LightBlue}\hl{ my}\sethlcolor{LightBlue}\hl{ knees}\sethlcolor{LightBlue}\hl{ start}\sethlcolor{LightBlue}\hl{ hurting}\sethlcolor{LightBlue}\hl{ very}\sethlcolor{LightBlue}\hl{ quickly}\sethlcolor{LightBlue}\hl{.}\sethlcolor{LightBlue}\hl{  }\sethlcolor{LightYellow}\hl{Should}\sethlcolor{LightBlue}\hl{ I}\sethlcolor{LightBlue}\hl{ see}\sethlcolor{LightBlue}\hl{ a}\sethlcolor{LightBlue}\hl{ Dr}\sethlcolor{LightYellow}\hl{,}\sethlcolor{LightBlue}\hl{ or}\sethlcolor{LightBlue}\hl{ should}\sethlcolor{LightBlue}\hl{ I}\sethlcolor{LightBlue}\hl{ just}\sethlcolor{LightBlue}\hl{ see}\sethlcolor{LightBlue}\hl{ a}\sethlcolor{LightBlue}\hl{ phys}\sethlcolor{LightBlue}\hl{io}\sethlcolor{LightBlue}\hl{?}\sethlcolor{LightYellow}\hl{  }\sethlcolor{LightBlue}\hl{Not}\sethlcolor{LightBlue}\hl{ quite}\sethlcolor{LightBlue}\hl{ sure}\sethlcolor{LightBlue}\hl{ how}\sethlcolor{LightBlue}\hl{ to}\sethlcolor{LightBlue}\hl{ proceed}\sethlcolor{LightBlue}\hl{.}\sethlcolor{LightRed}\hl{<|endoftext|>} & 3.1864 \\\midrule
Reference response &  I think pregnancy messed with my body, now I can't even run even the smallest amount without pain in my pelvis and knees.  I'm fairly certain the problem isn't just that I'm completely out of shape.<|endoftext|> & 2.4334 \\\midrule
Base Model Response & 

I have a very tight pelvis, and my knees are very sore.  I'm not sure if it's the pregnancy or the running, but I'm not sure if I can run anymore.

A:

I have had the same problem. I have been running for about a year now and I have had a lot of pain in my knees. I have been doing a & -8.9523 \\
\bottomrule
\end{tabular}
\end{small}
\end{table}

\begin{table}[t]
\caption{Sample query, responses from the 2.8B SFT, PPO, and DPO models; scores are from a 6.9B model. See Table~\ref{tab:ai2-visual} and Section~\ref{tab:ai2-visual} for semantics on colors.}
\label{tab:ai2-visual-2.8b-1}
\centering
\begin{small}
    \begin{tabular}{lp{7cm}l}
Type & Content & Score (RM) \\
\midrule
Query & SUBREDDIT: r/AskReddit

TITLE: How do you get someone out of your head?

POST: Hi,
I'm 22, and I have been with my girlfriend for 5 years now. We recently moved together. We've always loved each other intensely.

Problem, I recently started to have feelings for an other person (a friend). This person has had a boyfriend for now 3 years, and has absolutely no ideas. Those feelings were so strong, it was hard to hide them. After 2 months of me being distant and really sad, my girlfriend forced me to say what was bothering me. I'm not a good liar, and now she knows.

We decided to give us a week alone, I went to my parents. 

Now, I'm completely lost. I keep on thinking about this person, and I hate that. I would like for those feelings to go away, to leave me alone. But I can't.  

What do I do? It's been 3 months now, and I'm just desperate.

TL;DR: & N/A \\\midrule
SFT Model Response & \sethlcolor{LightYellow}\hl{ I}\sethlcolor{LightBlue}\hl{'m}\sethlcolor{LightBlue}\hl{ in}\sethlcolor{LightBlue}\hl{ love}\sethlcolor{LightBlue}\hl{ with}\sethlcolor{LightBlue}\hl{ a}\sethlcolor{LightBlue}\hl{ friend}\sethlcolor{LightBlue}\hl{,}\sethlcolor{LightBlue}\hl{ and}\sethlcolor{LightBlue}\hl{ I}\sethlcolor{LightYellow}\hl{ can}\sethlcolor{LightBlue}\hl{'t}\sethlcolor{LightYellow}\hl{ get}\sethlcolor{LightBlue}\hl{ her}\sethlcolor{LightBlue}\hl{ out}\sethlcolor{LightBlue}\hl{ of}\sethlcolor{LightBlue}\hl{ my}\sethlcolor{LightBlue}\hl{ head}\sethlcolor{LightBlue}\hl{.}\sethlcolor{LightYellow}\hl{ I}\sethlcolor{LightYellow}\hl{ want}\sethlcolor{LightBlue}\hl{ to}\sethlcolor{LightYellow}\hl{ get}\sethlcolor{LightYellow}\hl{ her}\sethlcolor{LightBlue}\hl{ out}\sethlcolor{LightBlue}\hl{ of}\sethlcolor{LightBlue}\hl{ my}\sethlcolor{LightBlue}\hl{ head}\sethlcolor{LightBlue}\hl{,}\sethlcolor{LightBlue}\hl{ but}\sethlcolor{LightBlue}\hl{ I}\sethlcolor{LightBlue}\hl{ can}\sethlcolor{LightBlue}\hl{'t}\sethlcolor{LightBlue}\hl{.}\sethlcolor{LightRed}\hl{<|endoftext|>} & -3.6299 \\\midrule
PPO Model Response & \sethlcolor{LightRed}\hl{ Be}\sethlcolor{LightYellow}\hl{en}\sethlcolor{LightYellow}\hl{ with}\sethlcolor{LightBlue}\hl{ my}\sethlcolor{LightYellow}\hl{ g}\sethlcolor{LightBlue}\hl{f}\sethlcolor{LightBlue}\hl{ for}\sethlcolor{LightBlue}\hl{ 5}\sethlcolor{LightBlue}\hl{ years}\sethlcolor{LightBlue}\hl{,}\sethlcolor{LightYellow}\hl{ recently}\sethlcolor{LightRed}\hl{ developed}\sethlcolor{LightBlue}\hl{ feelings}\sethlcolor{LightBlue}\hl{ for}\sethlcolor{LightYellow}\hl{ another}\sethlcolor{LightBlue}\hl{ person}\sethlcolor{LightYellow}\hl{.}\sethlcolor{LightRed}\hl{ Can}\sethlcolor{LightBlue}\hl{'t}\sethlcolor{LightBlue}\hl{ stop}\sethlcolor{LightBlue}\hl{ thinking}\sethlcolor{LightBlue}\hl{ about}\sethlcolor{LightYellow}\hl{ him}\sethlcolor{LightYellow}\hl{,}\sethlcolor{LightBlue}\hl{ and}\sethlcolor{LightYellow}\hl{ hate}\sethlcolor{LightBlue}\hl{ it}\sethlcolor{LightBlue}\hl{.}\sethlcolor{LightRed}\hl{ Has}\sethlcolor{LightRed}\hl{ anyone}\sethlcolor{LightYellow}\hl{ been}\sethlcolor{LightBlue}\hl{ in}\sethlcolor{LightYellow}\hl{ a}\sethlcolor{LightBlue}\hl{ similar}\sethlcolor{LightBlue}\hl{ situation}\sethlcolor{LightRed}\hl{ and}\sethlcolor{LightBlue}\hl{ how}\sethlcolor{LightBlue}\hl{ did}\sethlcolor{LightBlue}\hl{ you}\sethlcolor{LightBlue}\hl{ get}\sethlcolor{LightRed}\hl{ past}\sethlcolor{LightBlue}\hl{ it}\sethlcolor{LightBlue}\hl{?}\sethlcolor{LightRed}\hl{<|endoftext|>} & 3.0524 \\\midrule
DPO Model Response & \sethlcolor{LightRed}\hl{ Be}\sethlcolor{LightYellow}\hl{en}\sethlcolor{LightYellow}\hl{ with}\sethlcolor{LightRed}\hl{ g}\sethlcolor{LightBlue}\hl{f}\sethlcolor{LightBlue}\hl{ for}\sethlcolor{LightBlue}\hl{ 5}\sethlcolor{LightBlue}\hl{ years}\sethlcolor{LightBlue}\hl{,}\sethlcolor{LightYellow}\hl{ moved}\sethlcolor{LightYellow}\hl{ together}\sethlcolor{LightBlue}\hl{,}\sethlcolor{LightRed}\hl{ started}\sethlcolor{LightBlue}\hl{ to}\sethlcolor{LightBlue}\hl{ have}\sethlcolor{LightYellow}\hl{ strong}\sethlcolor{LightBlue}\hl{ feelings}\sethlcolor{LightBlue}\hl{ for}\sethlcolor{LightRed}\hl{ someone}\sethlcolor{LightBlue}\hl{ else}\sethlcolor{LightYellow}\hl{.}\sethlcolor{LightRed}\hl{ Made}\sethlcolor{LightRed}\hl{ it}\sethlcolor{LightYellow}\hl{ clear}\sethlcolor{LightBlue}\hl{ to}\sethlcolor{LightRed}\hl{ me}\sethlcolor{LightBlue}\hl{ that}\sethlcolor{LightBlue}\hl{ I}\sethlcolor{LightRed}\hl{ needed}\sethlcolor{LightBlue}\hl{ to}\sethlcolor{LightRed}\hl{ explain}\sethlcolor{LightRed}\hl{ why}\sethlcolor{LightBlue}\hl{ I}\sethlcolor{LightBlue}\hl{ was}\sethlcolor{LightRed}\hl{ being}\sethlcolor{LightBlue}\hl{ distant}\sethlcolor{LightBlue}\hl{.}\sethlcolor{LightBlue}\hl{ Now}\sethlcolor{LightBlue}\hl{ I}\sethlcolor{LightBlue}\hl{'m}\sethlcolor{LightRed}\hl{ 3}\sethlcolor{LightBlue}\hl{ months}\sethlcolor{LightBlue}\hl{ in}\sethlcolor{LightYellow}\hl{ and}\sethlcolor{LightBlue}\hl{ I}\sethlcolor{LightYellow}\hl{ can}\sethlcolor{LightBlue}\hl{'t}\sethlcolor{LightRed}\hl{ go}\sethlcolor{LightBlue}\hl{ back}\sethlcolor{LightBlue}\hl{ to}\sethlcolor{LightRed}\hl{ normal}\sethlcolor{LightBlue}\hl{.}\sethlcolor{LightRed}\hl{ How}\sethlcolor{LightBlue}\hl{ do}\sethlcolor{LightBlue}\hl{ I}\sethlcolor{LightBlue}\hl{ get}\sethlcolor{LightRed}\hl{ the}\sethlcolor{LightRed}\hl{ strong}\sethlcolor{LightYellow}\hl{ feeling}\sethlcolor{LightRed}\hl{ go}\sethlcolor{LightBlue}\hl{ away}\sethlcolor{LightBlue}\hl{?}\sethlcolor{LightBlue}\hl{
}\sethlcolor{LightBlue}\hl{
}\sethlcolor{LightYellow}\hl{TL}\sethlcolor{LightBlue}\hl{;}\sethlcolor{LightBlue}\hl{DR}\sethlcolor{LightBlue}\hl{:}\sethlcolor{LightBlue}\hl{ Be}\sethlcolor{LightBlue}\hl{en}\sethlcolor{LightBlue}\hl{ with}\sethlcolor{LightBlue}\hl{ g}\sethlcolor{LightBlue}\hl{f}\sethlcolor{LightBlue}\hl{ for}\sethlcolor{LightBlue}\hl{ 5}\sethlcolor{LightBlue}\hl{ years}\sethlcolor{LightBlue}\hl{,}\sethlcolor{LightBlue}\hl{ moved}\sethlcolor{LightBlue}\hl{ together}\sethlcolor{LightBlue}\hl{,}\sethlcolor{LightBlue}\hl{ started}\sethlcolor{LightBlue}\hl{ to}\sethlcolor{LightBlue}\hl{ have}\sethlcolor{LightBlue}\hl{ strong}\sethlcolor{LightBlue}\hl{ feelings}\sethlcolor{LightBlue}\hl{ for}\sethlcolor{LightBlue}\hl{ someone}\sethlcolor{LightBlue}\hl{ else}\sethlcolor{LightBlue}\hl{.}\sethlcolor{LightBlue}\hl{ Made}\sethlcolor{LightBlue}\hl{ it}\sethlcolor{LightBlue}\hl{ clear}\sethlcolor{LightBlue}\hl{ to}\sethlcolor{LightBlue}\hl{ me}\sethlcolor{LightBlue}\hl{ that}\sethlcolor{LightBlue}\hl{ I}\sethlcolor{LightBlue}\hl{ needed}\sethlcolor{LightBlue}\hl{ to}\sethlcolor{LightBlue}\hl{ explain}\sethlcolor{LightBlue}\hl{ why} & -2.9696 \\\midrule
Reference response &  long relationship; fell in love with an other person; admitted it; would like it to disappear, though it doesn't.<|endoftext|> & -1.6587 \\\midrule
Base Model Response &  How do you get someone out of your head?

A:

I'm going to assume that you're not in a relationship with this person, and that you're not in a relationship with your girlfriend.
If you're in a relationship with your girlfriend, then you're in a relationship with her. If you're not in a relationship with her, then you're not in a relationship with her.
If you're in a relationship with your girlfriend, then you're in a & -6.361 \\
\bottomrule
\end{tabular}
\end{small}
\end{table}

\begin{table}[t]
\caption{Sample query, responses from the 2.8B SFT, PPO, and DPO models; scores are from a 6.9B model. See Table~\ref{tab:ai2-visual} and Section~\ref{tab:ai2-visual} for semantics on colors.}
\label{tab:ai2-visual-2.8b-2}
\centering
\begin{small}
    \begin{tabular}{lp{7cm}l}
\toprule
Type & Content & Score (RM) \\
\midrule
Query & SUBREDDIT: r/pettyrevenge

TITLE: So, my mom woke me up with a loud TV.

POST: She was in her living room, watching TV. This was at about 8:30 in the morning, and she was exercising. She turned the TV up extra loud to hear it over her excercycle, and woke me up. I went in there asking for her to turn it down. She said she didn't have to; I explained that I always used headphones so she didn't have to deal with my noise and that she should give me a little more respect, given that I paid rent at the time.

She disagreed. I went back to my room, rather pissed off at the lack of equality. I had no lock on my door; but I had a dresser right next to it, so I pulled one of the drawers out enough so that it caused the door to not be openable. Then, I turned my speakers up really loud and blasted Gangnam Style on repeat, with the bass cranked up as high as it could go.

If you hate Gangnam Style for being overplayed, you will see why I chose that particular song. I personally don't mind it. But here's the thing about my bass; it vibrates the walls, making one hell of a lot of noise. Needless to say, my mom was not pleased and shut off the internet. But it was oh so worth it.

TL;DR: & N/A \\\midrule
SFT Model Response & \sethlcolor{LightYellow}\hl{ My}\sethlcolor{LightBlue}\hl{ mom}\sethlcolor{LightBlue}\hl{ woke}\sethlcolor{LightBlue}\hl{ me}\sethlcolor{LightBlue}\hl{ up}\sethlcolor{LightBlue}\hl{ with}\sethlcolor{LightBlue}\hl{ a}\sethlcolor{LightBlue}\hl{ loud}\sethlcolor{LightBlue}\hl{ TV}\sethlcolor{LightYellow}\hl{,}\sethlcolor{LightBlue}\hl{ and}\sethlcolor{LightBlue}\hl{ I}\sethlcolor{LightBlue}\hl{ turned}\sethlcolor{LightBlue}\hl{ it}\sethlcolor{LightBlue}\hl{ up}\sethlcolor{LightBlue}\hl{ really}\sethlcolor{LightBlue}\hl{ loud}\sethlcolor{LightBlue}\hl{.}\sethlcolor{LightRed}\hl{ She}\sethlcolor{LightYellow}\hl{ shut}\sethlcolor{LightBlue}\hl{ off}\sethlcolor{LightBlue}\hl{ the}\sethlcolor{LightBlue}\hl{ internet}\sethlcolor{LightBlue}\hl{.}\sethlcolor{LightRed}\hl{<|endoftext|>} & -0.3209 \\\midrule
PPO Model Response & \sethlcolor{LightYellow}\hl{ My}\sethlcolor{LightBlue}\hl{ mom}\sethlcolor{LightBlue}\hl{ woke}\sethlcolor{LightBlue}\hl{ me}\sethlcolor{LightBlue}\hl{ up}\sethlcolor{LightBlue}\hl{ with}\sethlcolor{LightBlue}\hl{ a}\sethlcolor{LightBlue}\hl{ loud}\sethlcolor{LightBlue}\hl{ TV}\sethlcolor{LightYellow}\hl{,}\sethlcolor{LightYellow}\hl{ I}\sethlcolor{LightBlue}\hl{ turned}\sethlcolor{LightYellow}\hl{ my}\sethlcolor{LightBlue}\hl{ speakers}\sethlcolor{LightBlue}\hl{ up}\sethlcolor{LightBlue}\hl{ really}\sethlcolor{LightBlue}\hl{ loud}\sethlcolor{LightBlue}\hl{ and}\sethlcolor{LightBlue}\hl{ blasted}\sethlcolor{LightBlue}\hl{ Gang}\sethlcolor{LightBlue}\hl{nam}\sethlcolor{LightBlue}\hl{ Style}\sethlcolor{LightYellow}\hl{.}\sethlcolor{LightYellow}\hl{ She}\sethlcolor{LightBlue}\hl{ shut}\sethlcolor{LightBlue}\hl{ off}\sethlcolor{LightBlue}\hl{ the}\sethlcolor{LightBlue}\hl{ internet}\sethlcolor{LightRed}\hl{ afterwards}\sethlcolor{LightBlue}\hl{.}\sethlcolor{LightYellow}\hl{<|endoftext|>} & 2.4708 \\\midrule
DPO Model Response & \sethlcolor{LightYellow}\hl{ My}\sethlcolor{LightRed}\hl{ loud}\sethlcolor{LightRed}\hl{-}\sethlcolor{LightRed}\hl{volume}\sethlcolor{LightBlue}\hl{-}\sethlcolor{LightRed}\hl{w}\sethlcolor{LightRed}\hl{aking}\sethlcolor{LightBlue}\hl{-}\sethlcolor{LightBlue}\hl{up}\sethlcolor{LightBlue}\hl{-}\sethlcolor{LightRed}\hl{mom}\sethlcolor{LightRed}\hl{ma}\sethlcolor{LightRed}\hl{ woke}\sethlcolor{LightBlue}\hl{ me}\sethlcolor{LightBlue}\hl{ up}\sethlcolor{LightYellow}\hl{ with}\sethlcolor{LightYellow}\hl{ the}\sethlcolor{LightRed}\hl{ TV}\sethlcolor{LightRed}\hl{ turned}\sethlcolor{LightBlue}\hl{ up}\sethlcolor{LightRed}\hl{ loud}\sethlcolor{LightYellow}\hl{,}\sethlcolor{LightYellow}\hl{ so}\sethlcolor{LightBlue}\hl{ I}\sethlcolor{LightYellow}\hl{ blasted}\sethlcolor{LightBlue}\hl{ Gang}\sethlcolor{LightBlue}\hl{nam}\sethlcolor{LightBlue}\hl{ Style}\sethlcolor{LightBlue}\hl{ on}\sethlcolor{LightRed}\hl{ her}\sethlcolor{LightRed}\hl{ bass}\sethlcolor{LightRed}\hl{-}\sethlcolor{LightBlue}\hl{v}\sethlcolor{LightBlue}\hl{ibr}\sethlcolor{LightBlue}\hl{ating}\sethlcolor{LightBlue}\hl{ walls}\sethlcolor{LightYellow}\hl{,}\sethlcolor{LightRed}\hl{ to}\sethlcolor{LightRed}\hl{ show}\sethlcolor{LightBlue}\hl{ her}\sethlcolor{LightYellow}\hl{ who}\sethlcolor{LightBlue}\hl{'s}\sethlcolor{LightBlue}\hl{ boss}\sethlcolor{LightBlue}\hl{.}\sethlcolor{LightYellow}\hl{<|endoftext|>} & 1.8347 \\\midrule
Reference response &  Mom had the TV on loud and woke me up, didn't care that I'd respected audio levels in the house, so I countered with playing Gangnam Style on repeat with the bass thumping through the walls.<|endoftext|> & 1.3359 \\\midrule
Base Model Response &  I woke up my mom with a loud TV. She was exercising. I was pissed.

TITLE: I'm not a racist, but I'm not a racist.

POST: I'm not a racist. I'm not a racist. I'm not a racist. I'm not a racist. I'm not a racist. I'm not a racist. I'm not a racist. I'm not a racist. I'm not a racist. I'm not a racist. & -5.5731 \\
\bottomrule
\end{tabular}
\end{small}
\end{table}

\begin{table}[t]
\caption{Sample query, responses from the 2.8B SFT, PPO, and DPO models; scores are from a 6.9B model. See Table~\ref{tab:ai2-visual} and Section~\ref{tab:ai2-visual} for semantics on colors.}
\label{tab:ai2-visual-2.8b-3}
\centering
\begin{small}
    \begin{tabular}{lp{7cm}l}
Type & Content & Score (RM) \\
\midrule
Query & SUBREDDIT: r/relationships

TITLE: My girlfriend (20f) of two years cheated on me (20m) by kissing two guys at a Halloween party.

POST: Lately her and I have been having a few problems, and these problems have been brought up before a few times. One problem being that I don't show enough affection. I don't tell her she's pretty very often or don't compliment her much. I feel terrible about it, but this time I was really trying to change for her.

For Halloween she went to visit her step brother at a college and I got drunk with my friends and watched movies. Last night (11/1) we got in a huge fight about me not changing and how our relationship won't work out and basically broke up over the phone. So in an effort to try and fix it I drove to her house. She told me how at the parties she went to that two guys kissed her. The first one she pushed away, but the second one I asked her if she kissed him back and she said yes and that she did it because it made her feel wanted, which I guess I haven't been making her feel that way lately. We cried, we talked about everything, we had great sex, and I stayed over at her house just to sleep with her and then snuck out in the morning so her parents wouldn't know.

We both obviously want to work things out but aren't sure if we should. I love this girl, but the more I think about it, all I can think about is her cheating on me, and more importantly, liking it. It makes me sick to my stomach. Should I even try to fix it or would I be better off cutting all ties.

TL;DR: & N/A \\\midrule
SFT Model Response & \sethlcolor{LightRed}\hl{ Girl}\sethlcolor{LightBlue}\hl{friend}\sethlcolor{LightBlue}\hl{ che}\sethlcolor{LightBlue}\hl{ated}\sethlcolor{LightBlue}\hl{ on}\sethlcolor{LightBlue}\hl{ me}\sethlcolor{LightRed}\hl{ by}\sethlcolor{LightBlue}\hl{ kissing}\sethlcolor{LightBlue}\hl{ two}\sethlcolor{LightBlue}\hl{ guys}\sethlcolor{LightBlue}\hl{ at}\sethlcolor{LightBlue}\hl{ a}\sethlcolor{LightBlue}\hl{ Halloween}\sethlcolor{LightBlue}\hl{ party}\sethlcolor{LightBlue}\hl{.}\sethlcolor{LightBlue}\hl{ I}\sethlcolor{LightYellow}\hl{ want}\sethlcolor{LightBlue}\hl{ to}\sethlcolor{LightBlue}\hl{ fix}\sethlcolor{LightBlue}\hl{ it}\sethlcolor{LightBlue}\hl{,}\sethlcolor{LightBlue}\hl{ but}\sethlcolor{LightBlue}\hl{ I}\sethlcolor{LightYellow}\hl{ don}\sethlcolor{LightBlue}\hl{'t}\sethlcolor{LightBlue}\hl{ know}\sethlcolor{LightBlue}\hl{ if}\sethlcolor{LightBlue}\hl{ I}\sethlcolor{LightBlue}\hl{ should}\sethlcolor{LightBlue}\hl{.}\sethlcolor{LightRed}\hl{<|endoftext|>} & 0.3877 \\\midrule
PPO Model Response & \sethlcolor{LightYellow}\hl{ My}\sethlcolor{LightBlue}\hl{ girlfriend}\sethlcolor{LightYellow}\hl{ of}\sethlcolor{LightBlue}\hl{ two}\sethlcolor{LightBlue}\hl{ years}\sethlcolor{LightBlue}\hl{ che}\sethlcolor{LightBlue}\hl{ated}\sethlcolor{LightBlue}\hl{ on}\sethlcolor{LightBlue}\hl{ me}\sethlcolor{LightBlue}\hl{ by}\sethlcolor{LightBlue}\hl{ kissing}\sethlcolor{LightBlue}\hl{ two}\sethlcolor{LightBlue}\hl{ guys}\sethlcolor{LightBlue}\hl{ at}\sethlcolor{LightBlue}\hl{ a}\sethlcolor{LightBlue}\hl{ Halloween}\sethlcolor{LightBlue}\hl{ party}\sethlcolor{LightBlue}\hl{.}\sethlcolor{LightYellow}\hl{ I}\sethlcolor{LightYellow}\hl{ love}\sethlcolor{LightBlue}\hl{ her}\sethlcolor{LightYellow}\hl{ but}\sethlcolor{LightRed}\hl{ the}\sethlcolor{LightYellow}\hl{ thought}\sethlcolor{LightBlue}\hl{ of}\sethlcolor{LightBlue}\hl{ her}\sethlcolor{LightYellow}\hl{ cheating}\sethlcolor{LightYellow}\hl{ makes}\sethlcolor{LightBlue}\hl{ me}\sethlcolor{LightBlue}\hl{ sick}\sethlcolor{LightBlue}\hl{ to}\sethlcolor{LightBlue}\hl{ my}\sethlcolor{LightBlue}\hl{ stomach}\sethlcolor{LightBlue}\hl{.}\sethlcolor{LightBlue}\hl{ Should}\sethlcolor{LightBlue}\hl{ I}\sethlcolor{LightBlue}\hl{ try}\sethlcolor{LightBlue}\hl{ to}\sethlcolor{LightBlue}\hl{ fix}\sethlcolor{LightBlue}\hl{ it}\sethlcolor{LightBlue}\hl{ or}\sethlcolor{LightYellow}\hl{ would}\sethlcolor{LightBlue}\hl{ I}\sethlcolor{LightBlue}\hl{ be}\sethlcolor{LightBlue}\hl{ better}\sethlcolor{LightBlue}\hl{ off}\sethlcolor{LightBlue}\hl{ cutting}\sethlcolor{LightBlue}\hl{ all}\sethlcolor{LightBlue}\hl{ ties}\sethlcolor{LightBlue}\hl{?}\sethlcolor{LightYellow}\hl{<|endoftext|>} & 4.1882 \\\midrule
DPO Model Response & \sethlcolor{LightYellow}\hl{ My}\sethlcolor{LightBlue}\hl{ girlfriend}\sethlcolor{LightBlue}\hl{ che}\sethlcolor{LightBlue}\hl{ated}\sethlcolor{LightBlue}\hl{ on}\sethlcolor{LightBlue}\hl{ me}\sethlcolor{LightRed}\hl{ by}\sethlcolor{LightBlue}\hl{ kissing}\sethlcolor{LightBlue}\hl{ two}\sethlcolor{LightBlue}\hl{ guys}\sethlcolor{LightBlue}\hl{ at}\sethlcolor{LightBlue}\hl{ a}\sethlcolor{LightYellow}\hl{ party}\sethlcolor{LightRed}\hl{ while}\sethlcolor{LightRed}\hl{ she}\sethlcolor{LightBlue}\hl{ was}\sethlcolor{LightYellow}\hl{ visiting}\sethlcolor{LightRed}\hl{ family}\sethlcolor{LightBlue}\hl{.}\sethlcolor{LightRed}\hl{ She}\sethlcolor{LightYellow}\hl{ told}\sethlcolor{LightBlue}\hl{ me}\sethlcolor{LightBlue}\hl{ about}\sethlcolor{LightBlue}\hl{ it}\sethlcolor{LightRed}\hl{ during}\sethlcolor{LightYellow}\hl{ our}\sethlcolor{LightBlue}\hl{ fight}\sethlcolor{LightBlue}\hl{ and}\sethlcolor{LightRed}\hl{ cried}\sethlcolor{LightYellow}\hl{.}\sethlcolor{LightYellow}\hl{ She}\sethlcolor{LightRed}\hl{ says}\sethlcolor{LightBlue}\hl{ she}\sethlcolor{LightRed}\hl{ did}\sethlcolor{LightBlue}\hl{ it}\sethlcolor{LightBlue}\hl{ because}\sethlcolor{LightBlue}\hl{ she}\sethlcolor{LightBlue}\hl{ felt}\sethlcolor{LightBlue}\hl{ wanted}\sethlcolor{LightRed}\hl{ by}\sethlcolor{LightBlue}\hl{ the}\sethlcolor{LightYellow}\hl{ guys}\sethlcolor{LightYellow}\hl{,}\sethlcolor{LightBlue}\hl{ but}\sethlcolor{LightBlue}\hl{ I}\sethlcolor{LightYellow}\hl{ think}\sethlcolor{LightYellow}\hl{ it}\sethlcolor{LightRed}\hl{ goes}\sethlcolor{LightBlue}\hl{ deeper}\sethlcolor{LightBlue}\hl{ than}\sethlcolor{LightBlue}\hl{ that}\sethlcolor{LightBlue}\hl{.}\sethlcolor{LightBlue}\hl{ I}\sethlcolor{LightRed}\hl{ hate}\sethlcolor{LightBlue}\hl{ that}\sethlcolor{LightBlue}\hl{ she}\sethlcolor{LightYellow}\hl{ did}\sethlcolor{LightBlue}\hl{ it}\sethlcolor{LightBlue}\hl{,}\sethlcolor{LightYellow}\hl{ and}\sethlcolor{LightYellow}\hl{ hate}\sethlcolor{LightYellow}\hl{ myself}\sethlcolor{LightBlue}\hl{ for}\sethlcolor{LightBlue}\hl{ not}\sethlcolor{LightRed}\hl{ showing}\sethlcolor{LightBlue}\hl{ her}\sethlcolor{LightBlue}\hl{ enough}\sethlcolor{LightBlue}\hl{ affection}\sethlcolor{LightBlue}\hl{.}\sethlcolor{LightYellow}\hl{ Should}\sethlcolor{LightBlue}\hl{ I}\sethlcolor{LightBlue}\hl{ try}\sethlcolor{LightBlue}\hl{ to}\sethlcolor{LightBlue}\hl{ fix}\sethlcolor{LightYellow}\hl{ things}\sethlcolor{LightBlue}\hl{ or}\sethlcolor{LightBlue}\hl{ cut}\sethlcolor{LightBlue}\hl{ all}\sethlcolor{LightBlue}\hl{ ties}\sethlcolor{LightYellow}\hl{.}\sethlcolor{LightYellow}\hl{<|endoftext|>} & 3.3642 \\\midrule
Reference response &  My girlfriend kissed two guys at a Halloween party and liked it. We love each other but don't know what to do about it.<|endoftext|> & 0.1145 \\\midrule
Base Model Response &  I'm in love with a girl, but I'm not sure if I should try to fix things or just cut all ties.

TL;DR: I'm in love with a girl, but I'm not sure if I should try to fix things or just cut all ties.

TL;DR: I'm in love with a girl, but I'm not sure if I should try to fix things or just cut all ties.

TL;DR: I'm in love & -5.5782 \\
\bottomrule
\end{tabular}
\end{small}
\end{table}

\begin{table}[t]
\caption{Sample query, responses from the 2.8B SFT, PPO, and DPO models; scores are from a 6.9B model. See Table~\ref{tab:ai2-visual} and Section~\ref{tab:ai2-visual} for semantics on colors.}
\label{tab:ai2-visual-2.8b-4}
\centering
\begin{small}
    \begin{tabular}{lp{7cm}l}
\toprule
Type & Content & Score (RM) \\
\midrule
Query & SUBREDDIT: r/running

TITLE: One year post-pregnancy, aches and pains in hips and knees, can't even comfortably do day 1 of C25k. Help?

POST: I'll try to keep this short!

**Background**

* I've always been an on again/off again (very casual!) jogger, typically doing 3 - 5 k
* My knees have always been finicky, and I went to a physio who thought I had "runner's knee"
* Pre-pregnancy, my "runner's knee" would flare up when I got to the 8 - 10 k distance range, even if I had a decent base (doing a C210k type program)

**Current Problem**

I had my baby a year ago, so all in all I haven't run for about 1.5 years.  I'm quite slim and have been doing aerobics-style classes for the past year, so I'm not totally out of shape.  Body weight exercises, aerobics, bikes and ellipticals are all fine.  However, when I run even the tinyiest bit, or even go on a long walk or a hike, my pelvis gets very sore and tight, and my knees start hurting very quickly.  I already am doing general squats/lunges/stretching type things.  

I'm starting to feel like running just isn't for me anymore. Which is a bummer, because I really enjoy running!

Has anyone had something similar? Can anyone recommend some stretches or exercises that might help?  Should I see a Dr?  Or should I just see a physio?  Not quite sure how to proceed.

Thanks!

TL;DR: & N/A \\\midrule
SFT Model Response & \sethlcolor{LightYellow}\hl{ I}\sethlcolor{LightBlue}\hl{'m}\sethlcolor{LightBlue}\hl{ a}\sethlcolor{LightBlue}\hl{ runner}\sethlcolor{LightBlue}\hl{,}\sethlcolor{LightYellow}\hl{ but}\sethlcolor{LightYellow}\hl{ I}\sethlcolor{LightBlue}\hl{'m}\sethlcolor{LightBlue}\hl{ not}\sethlcolor{LightYellow}\hl{ sure}\sethlcolor{LightYellow}\hl{ if}\sethlcolor{LightBlue}\hl{ I}\sethlcolor{LightBlue}\hl{'m}\sethlcolor{LightYellow}\hl{ ever}\sethlcolor{LightBlue}\hl{ going}\sethlcolor{LightBlue}\hl{ to}\sethlcolor{LightBlue}\hl{ be}\sethlcolor{LightBlue}\hl{ able}\sethlcolor{LightBlue}\hl{ to}\sethlcolor{LightBlue}\hl{ run}\sethlcolor{LightBlue}\hl{ again}\sethlcolor{LightBlue}\hl{.}\sethlcolor{LightBlue}\hl{  }\sethlcolor{LightYellow}\hl{Any}\sethlcolor{LightYellow}\hl{ advice}\sethlcolor{LightBlue}\hl{?}\sethlcolor{LightRed}\hl{<|endoftext|>} & -1.774 \\\midrule
PPO Model Response & \sethlcolor{LightRed}\hl{ One}\sethlcolor{LightBlue}\hl{ year}\sethlcolor{LightBlue}\hl{ post}\sethlcolor{LightBlue}\hl{-}\sethlcolor{LightBlue}\hl{p}\sethlcolor{LightBlue}\hl{regnancy}\sethlcolor{LightBlue}\hl{,}\sethlcolor{LightBlue}\hl{ ac}\sethlcolor{LightBlue}\hl{hes}\sethlcolor{LightBlue}\hl{ and}\sethlcolor{LightBlue}\hl{ pains}\sethlcolor{LightBlue}\hl{ in}\sethlcolor{LightBlue}\hl{ hips}\sethlcolor{LightBlue}\hl{ and}\sethlcolor{LightBlue}\hl{ knees}\sethlcolor{LightBlue}\hl{,}\sethlcolor{LightBlue}\hl{ can}\sethlcolor{LightBlue}\hl{'t}\sethlcolor{LightBlue}\hl{ even}\sethlcolor{LightBlue}\hl{ comfortably}\sethlcolor{LightBlue}\hl{ do}\sethlcolor{LightBlue}\hl{ day}\sethlcolor{LightBlue}\hl{ 1}\sethlcolor{LightBlue}\hl{ of}\sethlcolor{LightBlue}\hl{ C}\sethlcolor{LightBlue}\hl{25}\sethlcolor{LightBlue}\hl{k}\sethlcolor{LightBlue}\hl{.}\sethlcolor{LightYellow}\hl{ Has}\sethlcolor{LightBlue}\hl{ anyone}\sethlcolor{LightYellow}\hl{ else}\sethlcolor{LightRed}\hl{ gone}\sethlcolor{LightBlue}\hl{ through}\sethlcolor{LightYellow}\hl{ something}\sethlcolor{LightBlue}\hl{ similar}\sethlcolor{LightBlue}\hl{?}\sethlcolor{LightYellow}\hl{  }\sethlcolor{LightBlue}\hl{What}\sethlcolor{LightRed}\hl{ exercises}\sethlcolor{LightBlue}\hl{/}\sethlcolor{LightBlue}\hl{st}\sethlcolor{LightBlue}\hl{ret}\sethlcolor{LightBlue}\hl{ches}\sethlcolor{LightRed}\hl{ might}\sethlcolor{LightBlue}\hl{ help}\sethlcolor{LightBlue}\hl{?}\sethlcolor{LightRed}\hl{<|endoftext|>} & 4.5098 \\\midrule
DPO Model Response & \sethlcolor{LightRed}\hl{ One}\sethlcolor{LightBlue}\hl{ year}\sethlcolor{LightBlue}\hl{ post}\sethlcolor{LightBlue}\hl{-}\sethlcolor{LightBlue}\hl{p}\sethlcolor{LightBlue}\hl{regnancy}\sethlcolor{LightBlue}\hl{,}\sethlcolor{LightYellow}\hl{ my}\sethlcolor{LightYellow}\hl{ pelvis}\sethlcolor{LightBlue}\hl{ and}\sethlcolor{LightBlue}\hl{ knees}\sethlcolor{LightYellow}\hl{ hurt}\sethlcolor{LightYellow}\hl{ when}\sethlcolor{LightBlue}\hl{ I}\sethlcolor{LightBlue}\hl{ run}\sethlcolor{LightBlue}\hl{,}\sethlcolor{LightBlue}\hl{ and}\sethlcolor{LightBlue}\hl{ I}\sethlcolor{LightBlue}\hl{'m}\sethlcolor{LightBlue}\hl{ not}\sethlcolor{LightRed}\hl{ exactly}\sethlcolor{LightYellow}\hl{ out}\sethlcolor{LightBlue}\hl{ of}\sethlcolor{LightBlue}\hl{ shape}\sethlcolor{LightBlue}\hl{.}\sethlcolor{LightBlue}\hl{  }\sethlcolor{LightRed}\hl{Has}\sethlcolor{LightBlue}\hl{ anyone}\sethlcolor{LightBlue}\hl{ had}\sethlcolor{LightBlue}\hl{ something}\sethlcolor{LightBlue}\hl{ similar}\sethlcolor{LightYellow}\hl{,}\sethlcolor{LightBlue}\hl{ and}\sethlcolor{LightYellow}\hl{ can}\sethlcolor{LightYellow}\hl{ recommend}\sethlcolor{LightRed}\hl{ stretches}\sethlcolor{LightBlue}\hl{ or}\sethlcolor{LightBlue}\hl{ exercises}\sethlcolor{LightYellow}\hl{?}\sethlcolor{LightYellow}\hl{  }\sethlcolor{LightBlue}\hl{Should}\sethlcolor{LightBlue}\hl{ I}\sethlcolor{LightBlue}\hl{ see}\sethlcolor{LightBlue}\hl{ a}\sethlcolor{LightYellow}\hl{ Dr}\sethlcolor{LightBlue}\hl{?}\sethlcolor{LightRed}\hl{<|endoftext|>} & 5.5663 \\\midrule
Reference response &  I think pregnancy messed with my body, now I can't even run even the smallest amount without pain in my pelvis and knees.  I'm fairly certain the problem isn't just that I'm completely out of shape.<|endoftext|> & 2.4334 \\\midrule
Base Model Response & 

* I'm a runner, but I'm not a runner anymore
* I'm not out of shape, but I'm not in great shape either
* I'm not sure what to do about my aches and pains

A:

I'm not a doctor, but I have a few suggestions.

I would start with a general stretching routine.  I would start with a few minutes of static stretching, then move on to some dynamic stretching.  I would & -7.651 \\
\bottomrule
\end{tabular}
\end{small}
\end{table}

\begin{table}[t]
\caption{Sample query, responses from the 6.9B SFT, PPO, and DPO models; scores are from a 6.9B model. See Table~\ref{tab:ai2-visual} and Section~\ref{tab:ai2-visual} for semantics on colors.}
\label{tab:ai2-visual-6.9b-1}
\centering
\begin{small}
    \begin{tabular}{lp{7cm}l}
\toprule
Type & Content & Score (RM) \\
\midrule
Query & SUBREDDIT: r/AskReddit

TITLE: How do you get someone out of your head?

POST: Hi,
I'm 22, and I have been with my girlfriend for 5 years now. We recently moved together. We've always loved each other intensely.

Problem, I recently started to have feelings for an other person (a friend). This person has had a boyfriend for now 3 years, and has absolutely no ideas. Those feelings were so strong, it was hard to hide them. After 2 months of me being distant and really sad, my girlfriend forced me to say what was bothering me. I'm not a good liar, and now she knows.

We decided to give us a week alone, I went to my parents. 

Now, I'm completely lost. I keep on thinking about this person, and I hate that. I would like for those feelings to go away, to leave me alone. But I can't.  

What do I do? It's been 3 months now, and I'm just desperate.

TL;DR: & N/A \\\midrule
SFT Model Response & \sethlcolor{LightBlue}\hl{ I}\sethlcolor{LightYellow}\hl{ have}\sethlcolor{LightBlue}\hl{ feelings}\sethlcolor{LightBlue}\hl{ for}\sethlcolor{LightYellow}\hl{ an}\sethlcolor{LightBlue}\hl{ other}\sethlcolor{LightBlue}\hl{ person}\sethlcolor{LightBlue}\hl{,}\sethlcolor{LightBlue}\hl{ and}\sethlcolor{LightBlue}\hl{ I}\sethlcolor{LightBlue}\hl{ don}\sethlcolor{LightBlue}\hl{'t}\sethlcolor{LightBlue}\hl{ know}\sethlcolor{LightBlue}\hl{ how}\sethlcolor{LightBlue}\hl{ to}\sethlcolor{LightBlue}\hl{ get}\sethlcolor{LightBlue}\hl{ rid}\sethlcolor{LightBlue}\hl{ of}\sethlcolor{LightBlue}\hl{ them}\sethlcolor{LightBlue}\hl{.}\sethlcolor{LightRed}\hl{<|endoftext|>} & -3.2271 \\\midrule
PPO Model Response & \sethlcolor{LightRed}\hl{ Be}\sethlcolor{LightBlue}\hl{en}\sethlcolor{LightBlue}\hl{ with}\sethlcolor{LightBlue}\hl{ my}\sethlcolor{LightBlue}\hl{ girlfriend}\sethlcolor{LightBlue}\hl{ for}\sethlcolor{LightBlue}\hl{ 5}\sethlcolor{LightBlue}\hl{ years}\sethlcolor{LightBlue}\hl{,}\sethlcolor{LightRed}\hl{ recently}\sethlcolor{LightRed}\hl{ developed}\sethlcolor{LightBlue}\hl{ feelings}\sethlcolor{LightBlue}\hl{ for}\sethlcolor{LightBlue}\hl{ another}\sethlcolor{LightBlue}\hl{ person}\sethlcolor{LightBlue}\hl{.}\sethlcolor{LightYellow}\hl{ I}\sethlcolor{LightRed}\hl{ would}\sethlcolor{LightBlue}\hl{ like}\sethlcolor{LightYellow}\hl{ for}\sethlcolor{LightBlue}\hl{ those}\sethlcolor{LightBlue}\hl{ feelings}\sethlcolor{LightBlue}\hl{ to}\sethlcolor{LightBlue}\hl{ go}\sethlcolor{LightBlue}\hl{ away}\sethlcolor{LightBlue}\hl{,}\sethlcolor{LightBlue}\hl{ but}\sethlcolor{LightYellow}\hl{ they}\sethlcolor{LightBlue}\hl{ won}\sethlcolor{LightBlue}\hl{'t}\sethlcolor{LightBlue}\hl{.}\sethlcolor{LightYellow}\hl{ I}\sethlcolor{LightBlue}\hl{'m}\sethlcolor{LightRed}\hl{ lost}\sethlcolor{LightYellow}\hl{ and}\sethlcolor{LightRed}\hl{ don}\sethlcolor{LightBlue}\hl{'t}\sethlcolor{LightBlue}\hl{ know}\sethlcolor{LightBlue}\hl{ what}\sethlcolor{LightBlue}\hl{ to}\sethlcolor{LightBlue}\hl{ do}\sethlcolor{LightRed}\hl{<|endoftext|>} & 3.0259 \\\midrule
DPO Model Response & \sethlcolor{LightRed}\hl{ In}\sethlcolor{LightRed}\hl{ loving}\sethlcolor{LightBlue}\hl{ relationship}\sethlcolor{LightYellow}\hl{ with}\sethlcolor{LightYellow}\hl{ my}\sethlcolor{LightBlue}\hl{ girlfriend}\sethlcolor{LightYellow}\hl{ for}\sethlcolor{LightBlue}\hl{ 5}\sethlcolor{LightBlue}\hl{ years}\sethlcolor{LightYellow}\hl{.}\sethlcolor{LightYellow}\hl{ Start}\sethlcolor{LightBlue}\hl{ed}\sethlcolor{LightYellow}\hl{ having}\sethlcolor{LightBlue}\hl{ feelings}\sethlcolor{LightBlue}\hl{ for}\sethlcolor{LightBlue}\hl{ another}\sethlcolor{LightBlue}\hl{ person}\sethlcolor{LightBlue}\hl{.}\sethlcolor{LightYellow}\hl{ My}\sethlcolor{LightBlue}\hl{ girlfriend}\sethlcolor{LightYellow}\hl{ found}\sethlcolor{LightBlue}\hl{ out}\sethlcolor{LightBlue}\hl{,}\sethlcolor{LightBlue}\hl{ and}\sethlcolor{LightRed}\hl{ we}\sethlcolor{LightRed}\hl{ talked}\sethlcolor{LightYellow}\hl{.}\sethlcolor{LightBlue}\hl{ Now}\sethlcolor{LightBlue}\hl{ I}\sethlcolor{LightBlue}\hl{'m}\sethlcolor{LightBlue}\hl{ lost}\sethlcolor{LightYellow}\hl{,}\sethlcolor{LightYellow}\hl{ I}\sethlcolor{LightBlue}\hl{ can}\sethlcolor{LightBlue}\hl{'t}\sethlcolor{LightYellow}\hl{ get}\sethlcolor{LightYellow}\hl{ those}\sethlcolor{LightRed}\hl{ other}\sethlcolor{LightBlue}\hl{ feelings}\sethlcolor{LightYellow}\hl{ to}\sethlcolor{LightYellow}\hl{ leave}\sethlcolor{LightBlue}\hl{ me}\sethlcolor{LightBlue}\hl{ alone}\sethlcolor{LightBlue}\hl{.}\sethlcolor{LightRed}\hl{<|endoftext|>} & 2.5583 \\\midrule
Reference response &  long relationship; fell in love with an other person; admitted it; would like it to disappear, though it doesn't.<|endoftext|> & -1.6587 \\\midrule
Base Model Response &  I'm 22, and I have been with my girlfriend for 5 years now. We recently moved together. We've always loved each other intensely. I recently started to have feelings for an other person (a friend). This person has had a boyfriend for now 3 years, and has absolutely no ideas. Those feelings were so strong, it was hard to hide them. After 2 months of me being distant and really sad, my girlfriend forced me to say what was bothering me. I'm not a & -4.9909 \\
\bottomrule
\end{tabular}
\end{small}
\end{table}

\begin{table}[t]
\caption{Sample query, responses from the 6.9B SFT, PPO, and DPO models; scores are from a 6.9B model. See Table~\ref{tab:ai2-visual} and Section~\ref{tab:ai2-visual} for semantics on colors.}
\label{tab:ai2-visual-6.9b-2}
\centering
\begin{small}
    \begin{tabular}{lp{7cm}l}
\toprule
Type & Content & Score (RM) \\
\midrule
Query & SUBREDDIT: r/pettyrevenge

TITLE: So, my mom woke me up with a loud TV.

POST: She was in her living room, watching TV. This was at about 8:30 in the morning, and she was exercising. She turned the TV up extra loud to hear it over her excercycle, and woke me up. I went in there asking for her to turn it down. She said she didn't have to; I explained that I always used headphones so she didn't have to deal with my noise and that she should give me a little more respect, given that I paid rent at the time.

She disagreed. I went back to my room, rather pissed off at the lack of equality. I had no lock on my door; but I had a dresser right next to it, so I pulled one of the drawers out enough so that it caused the door to not be openable. Then, I turned my speakers up really loud and blasted Gangnam Style on repeat, with the bass cranked up as high as it could go.

If you hate Gangnam Style for being overplayed, you will see why I chose that particular song. I personally don't mind it. But here's the thing about my bass; it vibrates the walls, making one hell of a lot of noise. Needless to say, my mom was not pleased and shut off the internet. But it was oh so worth it.

TL;DR: & N/A \\\midrule
SFT Model Response & \sethlcolor{LightBlue}\hl{ My}\sethlcolor{LightBlue}\hl{ mom}\sethlcolor{LightBlue}\hl{ woke}\sethlcolor{LightBlue}\hl{ me}\sethlcolor{LightBlue}\hl{ up}\sethlcolor{LightBlue}\hl{ with}\sethlcolor{LightBlue}\hl{ a}\sethlcolor{LightBlue}\hl{ loud}\sethlcolor{LightBlue}\hl{ TV}\sethlcolor{LightBlue}\hl{.}\sethlcolor{LightYellow}\hl{ I}\sethlcolor{LightBlue}\hl{ blasted}\sethlcolor{LightBlue}\hl{ Gang}\sethlcolor{LightBlue}\hl{nam}\sethlcolor{LightBlue}\hl{ Style}\sethlcolor{LightBlue}\hl{ on}\sethlcolor{LightBlue}\hl{ repeat}\sethlcolor{LightBlue}\hl{,}\sethlcolor{LightBlue}\hl{ with}\sethlcolor{LightBlue}\hl{ the}\sethlcolor{LightBlue}\hl{ bass}\sethlcolor{LightBlue}\hl{ crank}\sethlcolor{LightBlue}\hl{ed}\sethlcolor{LightBlue}\hl{ up}\sethlcolor{LightBlue}\hl{ as}\sethlcolor{LightBlue}\hl{ high}\sethlcolor{LightBlue}\hl{ as}\sethlcolor{LightBlue}\hl{ it}\sethlcolor{LightBlue}\hl{ could}\sethlcolor{LightBlue}\hl{ go}\sethlcolor{LightBlue}\hl{.}\sethlcolor{LightRed}\hl{<|endoftext|>} & 0.9671 \\\midrule
PPO Model Response & \sethlcolor{LightBlue}\hl{ My}\sethlcolor{LightBlue}\hl{ mom}\sethlcolor{LightBlue}\hl{ woke}\sethlcolor{LightBlue}\hl{ me}\sethlcolor{LightBlue}\hl{ up}\sethlcolor{LightBlue}\hl{ with}\sethlcolor{LightYellow}\hl{ loud}\sethlcolor{LightBlue}\hl{ TV}\sethlcolor{LightBlue}\hl{.}\sethlcolor{LightYellow}\hl{ I}\sethlcolor{LightBlue}\hl{ blasted}\sethlcolor{LightBlue}\hl{ Gang}\sethlcolor{LightBlue}\hl{nam}\sethlcolor{LightBlue}\hl{ Style}\sethlcolor{LightRed}\hl{ loudly}\sethlcolor{LightRed}\hl{ through}\sethlcolor{LightBlue}\hl{ my}\sethlcolor{LightBlue}\hl{ speakers}\sethlcolor{LightYellow}\hl{,}\sethlcolor{LightYellow}\hl{ making}\sethlcolor{LightYellow}\hl{ a}\sethlcolor{LightYellow}\hl{ lot}\sethlcolor{LightBlue}\hl{ of}\sethlcolor{LightBlue}\hl{ noise}\sethlcolor{LightYellow}\hl{ and}\sethlcolor{LightYellow}\hl{ annoying}\sethlcolor{LightYellow}\hl{ her}\sethlcolor{LightBlue}\hl{.}\sethlcolor{LightRed}\hl{ Got}\sethlcolor{LightRed}\hl{ some}\sethlcolor{LightRed}\hl{ sweet}\sethlcolor{LightBlue}\hl{ revenge}\sethlcolor{LightRed}\hl{<|endoftext|>} & 3.9558 \\\midrule
DPO Model Response & \sethlcolor{LightBlue}\hl{ My}\sethlcolor{LightBlue}\hl{ mom}\sethlcolor{LightBlue}\hl{ woke}\sethlcolor{LightBlue}\hl{ me}\sethlcolor{LightBlue}\hl{ up}\sethlcolor{LightBlue}\hl{ with}\sethlcolor{LightYellow}\hl{ loud}\sethlcolor{LightBlue}\hl{ TV}\sethlcolor{LightYellow}\hl{,}\sethlcolor{LightRed}\hl{ because}\sethlcolor{LightBlue}\hl{ she}\sethlcolor{LightRed}\hl{ couldn}\sethlcolor{LightBlue}\hl{'t}\sethlcolor{LightBlue}\hl{ hear}\sethlcolor{LightYellow}\hl{ her}\sethlcolor{LightBlue}\hl{ ex}\sethlcolor{LightBlue}\hl{cer}\sethlcolor{LightBlue}\hl{cycle}\sethlcolor{LightRed}\hl{ while}\sethlcolor{LightBlue}\hl{ exercising}\sethlcolor{LightBlue}\hl{.}\sethlcolor{LightYellow}\hl{ She}\sethlcolor{LightBlue}\hl{ was}\sethlcolor{LightRed}\hl{ annoyed}\sethlcolor{LightRed}\hl{ with}\sethlcolor{LightBlue}\hl{ me}\sethlcolor{LightYellow}\hl{ for}\sethlcolor{LightRed}\hl{ having}\sethlcolor{LightBlue}\hl{ headphones}\sethlcolor{LightRed}\hl{ when}\sethlcolor{LightYellow}\hl{ I}\sethlcolor{LightBlue}\hl{ paid}\sethlcolor{LightBlue}\hl{ rent}\sethlcolor{LightBlue}\hl{.}\sethlcolor{LightBlue}\hl{ I}\sethlcolor{LightBlue}\hl{ blasted}\sethlcolor{LightBlue}\hl{ Gang}\sethlcolor{LightBlue}\hl{nam}\sethlcolor{LightBlue}\hl{ Style}\sethlcolor{LightRed}\hl{ loud}\sethlcolor{LightBlue}\hl{ enough}\sethlcolor{LightBlue}\hl{ to}\sethlcolor{LightBlue}\hl{ vibr}\sethlcolor{LightBlue}\hl{ate}\sethlcolor{LightBlue}\hl{ the}\sethlcolor{LightBlue}\hl{ walls}\sethlcolor{LightBlue}\hl{.}\sethlcolor{LightYellow}\hl{ She}\sethlcolor{LightYellow}\hl{ turned}\sethlcolor{LightBlue}\hl{ off}\sethlcolor{LightBlue}\hl{ the}\sethlcolor{LightBlue}\hl{ internet}\sethlcolor{LightBlue}\hl{.}\sethlcolor{LightRed}\hl{<|endoftext|>} & 1.3729 \\\midrule
Reference response &  Mom had the TV on loud and woke me up, didn't care that I'd respected audio levels in the house, so I countered with playing Gangnam Style on repeat with the bass thumping through the walls.<|endoftext|> & 1.3359 \\\midrule
Base Model Response &  My mom woke me up with a loud TV.

SUBREDDIT: r/pettyrevenge

TITLE: I'm a little bit of a dick.

POST: I'm a little bit of a dick.

I'm a little bit of a dick.

I'm a little bit of a dick.

I'm a little bit of a dick.

I'm a little bit of a dick.

I'm & -8.1818 \\
\bottomrule
\end{tabular}
\end{small}
\end{table}

\begin{table}[t]
\caption{Sample query, responses from the 6.9B SFT, PPO, and DPO models; scores are from a 6.9B model. See Table~\ref{tab:ai2-visual} and Section~\ref{tab:ai2-visual} for semantics on colors.}
\label{tab:ai2-visual-6.9b-3}
\centering
\begin{small}
    \begin{tabular}{lp{7cm}l}
\toprule
Type & Content & Score (RM) \\
\midrule
Query & SUBREDDIT: r/relationships

TITLE: My girlfriend (20f) of two years cheated on me (20m) by kissing two guys at a Halloween party.

POST: Lately her and I have been having a few problems, and these problems have been brought up before a few times. One problem being that I don't show enough affection. I don't tell her she's pretty very often or don't compliment her much. I feel terrible about it, but this time I was really trying to change for her.

For Halloween she went to visit her step brother at a college and I got drunk with my friends and watched movies. Last night (11/1) we got in a huge fight about me not changing and how our relationship won't work out and basically broke up over the phone. So in an effort to try and fix it I drove to her house. She told me how at the parties she went to that two guys kissed her. The first one she pushed away, but the second one I asked her if she kissed him back and she said yes and that she did it because it made her feel wanted, which I guess I haven't been making her feel that way lately. We cried, we talked about everything, we had great sex, and I stayed over at her house just to sleep with her and then snuck out in the morning so her parents wouldn't know.

We both obviously want to work things out but aren't sure if we should. I love this girl, but the more I think about it, all I can think about is her cheating on me, and more importantly, liking it. It makes me sick to my stomach. Should I even try to fix it or would I be better off cutting all ties.

TL;DR: & N/A \\\midrule
SFT Model Response & \sethlcolor{LightYellow}\hl{ Girl}\sethlcolor{LightYellow}\hl{friend}\sethlcolor{LightBlue}\hl{ che}\sethlcolor{LightBlue}\hl{ated}\sethlcolor{LightBlue}\hl{ on}\sethlcolor{LightBlue}\hl{ me}\sethlcolor{LightRed}\hl{ by}\sethlcolor{LightBlue}\hl{ kissing}\sethlcolor{LightBlue}\hl{ two}\sethlcolor{LightBlue}\hl{ guys}\sethlcolor{LightBlue}\hl{ at}\sethlcolor{LightBlue}\hl{ a}\sethlcolor{LightBlue}\hl{ Halloween}\sethlcolor{LightBlue}\hl{ party}\sethlcolor{LightBlue}\hl{.}\sethlcolor{LightYellow}\hl{ Should}\sethlcolor{LightBlue}\hl{ I}\sethlcolor{LightBlue}\hl{ try}\sethlcolor{LightBlue}\hl{ to}\sethlcolor{LightBlue}\hl{ fix}\sethlcolor{LightBlue}\hl{ it}\sethlcolor{LightBlue}\hl{ or}\sethlcolor{LightBlue}\hl{ cut}\sethlcolor{LightBlue}\hl{ all}\sethlcolor{LightBlue}\hl{ ties}\sethlcolor{LightBlue}\hl{?}\sethlcolor{LightRed}\hl{<|endoftext|>} & 1.3901 \\\midrule
PPO Model Response & \sethlcolor{LightBlue}\hl{ My}\sethlcolor{LightBlue}\hl{ girlfriend}\sethlcolor{LightYellow}\hl{ of}\sethlcolor{LightBlue}\hl{ two}\sethlcolor{LightBlue}\hl{ years}\sethlcolor{LightBlue}\hl{ che}\sethlcolor{LightBlue}\hl{ated}\sethlcolor{LightBlue}\hl{ on}\sethlcolor{LightBlue}\hl{ me}\sethlcolor{LightBlue}\hl{ by}\sethlcolor{LightBlue}\hl{ kissing}\sethlcolor{LightBlue}\hl{ two}\sethlcolor{LightBlue}\hl{ guys}\sethlcolor{LightBlue}\hl{ at}\sethlcolor{LightBlue}\hl{ a}\sethlcolor{LightBlue}\hl{ Halloween}\sethlcolor{LightBlue}\hl{ party}\sethlcolor{LightBlue}\hl{.}\sethlcolor{LightRed}\hl{ I}\sethlcolor{LightYellow}\hl{ love}\sethlcolor{LightBlue}\hl{ her}\sethlcolor{LightYellow}\hl{ but}\sethlcolor{LightYellow}\hl{ am}\sethlcolor{LightRed}\hl{ conflict}\sethlcolor{LightBlue}\hl{ed}\sethlcolor{LightBlue}\hl{ about}\sethlcolor{LightBlue}\hl{ whether}\sethlcolor{LightBlue}\hl{ or}\sethlcolor{LightBlue}\hl{ not}\sethlcolor{LightBlue}\hl{ to}\sethlcolor{LightBlue}\hl{ try}\sethlcolor{LightBlue}\hl{ and}\sethlcolor{LightBlue}\hl{ fix}\sethlcolor{LightYellow}\hl{ things}\sethlcolor{LightBlue}\hl{.}\sethlcolor{LightRed}\hl{ Should}\sethlcolor{LightBlue}\hl{ I}\sethlcolor{LightRed}\hl{ cut}\sethlcolor{LightYellow}\hl{ ties}\sethlcolor{LightBlue}\hl{ or}\sethlcolor{LightBlue}\hl{ try}\sethlcolor{LightBlue}\hl{ to}\sethlcolor{LightRed}\hl{ salvage}\sethlcolor{LightYellow}\hl{ what}\sethlcolor{LightBlue}\hl{ we}\sethlcolor{LightBlue}\hl{ have}\sethlcolor{LightYellow}\hl{ left}\sethlcolor{LightRed}\hl{<|endoftext|>} & 5.2579 \\\midrule
DPO Model Response & \sethlcolor{LightBlue}\hl{ My}\sethlcolor{LightBlue}\hl{ girlfriend}\sethlcolor{LightBlue}\hl{ che}\sethlcolor{LightBlue}\hl{ated}\sethlcolor{LightBlue}\hl{ on}\sethlcolor{LightBlue}\hl{ me}\sethlcolor{LightYellow}\hl{ by}\sethlcolor{LightBlue}\hl{ kissing}\sethlcolor{LightBlue}\hl{ two}\sethlcolor{LightBlue}\hl{ guys}\sethlcolor{LightBlue}\hl{ at}\sethlcolor{LightBlue}\hl{ a}\sethlcolor{LightBlue}\hl{ Halloween}\sethlcolor{LightBlue}\hl{ party}\sethlcolor{LightRed}\hl{ friendly}\sethlcolor{LightRed}\hl{ make}\sethlcolor{LightBlue}\hl{ out}\sethlcolor{LightBlue}\hl{ session}\sethlcolor{LightBlue}\hl{.}\sethlcolor{LightRed}\hl{ She}\sethlcolor{LightRed}\hl{ claims}\sethlcolor{LightBlue}\hl{ she}\sethlcolor{LightRed}\hl{ only}\sethlcolor{LightYellow}\hl{ kissed}\sethlcolor{LightRed}\hl{ to}\sethlcolor{LightYellow}\hl{ feel}\sethlcolor{LightBlue}\hl{ wanted}\sethlcolor{LightYellow}\hl{,}\sethlcolor{LightBlue}\hl{ but}\sethlcolor{LightBlue}\hl{ I}\sethlcolor{LightRed}\hl{ can}\sethlcolor{LightBlue}\hl{'t}\sethlcolor{LightRed}\hl{ shake}\sethlcolor{LightBlue}\hl{ the}\sethlcolor{LightBlue}\hl{ feeling}\sethlcolor{LightBlue}\hl{ that}\sethlcolor{LightBlue}\hl{ she}\sethlcolor{LightRed}\hl{ che}\sethlcolor{LightBlue}\hl{ated}\sethlcolor{LightRed}\hl{ and}\sethlcolor{LightYellow}\hl{ I}\sethlcolor{LightRed}\hl{ want}\sethlcolor{LightBlue}\hl{ to}\sethlcolor{LightRed}\hl{ cut}\sethlcolor{LightYellow}\hl{ all}\sethlcolor{LightBlue}\hl{ ties}\sethlcolor{LightBlue}\hl{.}\sethlcolor{LightRed}\hl{<|endoftext|>} & 2.0515 \\\midrule
Reference response &  My girlfriend kissed two guys at a Halloween party and liked it. We love each other but don't know what to do about it.<|endoftext|> & 0.1145 \\\midrule
Base Model Response &  My girlfriend cheated on me with two guys at a Halloween party.

SUB: r/relationships

TITLE: My girlfriend (20f) of two years cheated on me (20m) by kissing two guys at a Halloween party.

POST: Lately her and I have been having a few problems, and these problems have been brought up before a few times. One problem being that I don't show enough affection. I don't tell her she's pretty & -5.0535 \\
\bottomrule
\end{tabular}
\end{small}
\end{table}

\begin{table}[t]
\caption{Sample query, responses from the 6.9B SFT, PPO, and DPO models; scores are from a 6.9B model. See Table~\ref{tab:ai2-visual} and Section~\ref{tab:ai2-visual} for semantics on colors.}
\label{tab:ai2-visual-6.9b-4}
\centering
\begin{small}
\begin{tabular}{lp{7cm}l}
\toprule
Type & Content & Score (RM) \\
\midrule
Query & SUBREDDIT: r/running

TITLE: One year post-pregnancy, aches and pains in hips and knees, can't even comfortably do day 1 of C25k. Help?

POST: I'll try to keep this short!

**Background**

* I've always been an on again/off again (very casual!) jogger, typically doing 3 - 5 k
* My knees have always been finicky, and I went to a physio who thought I had "runner's knee"
* Pre-pregnancy, my "runner's knee" would flare up when I got to the 8 - 10 k distance range, even if I had a decent base (doing a C210k type program)

**Current Problem**

I had my baby a year ago, so all in all I haven't run for about 1.5 years.  I'm quite slim and have been doing aerobics-style classes for the past year, so I'm not totally out of shape.  Body weight exercises, aerobics, bikes and ellipticals are all fine.  However, when I run even the tinyiest bit, or even go on a long walk or a hike, my pelvis gets very sore and tight, and my knees start hurting very quickly.  I already am doing general squats/lunges/stretching type things.  

I'm starting to feel like running just isn't for me anymore. Which is a bummer, because I really enjoy running!

Has anyone had something similar? Can anyone recommend some stretches or exercises that might help?  Should I see a Dr?  Or should I just see a physio?  Not quite sure how to proceed.

Thanks!

TL;DR: & N/A \\\midrule
SFT Model Response & \sethlcolor{LightRed}\hl{ 1}\sethlcolor{LightYellow}\hl{.}\sethlcolor{LightBlue}\hl{5}\sethlcolor{LightBlue}\hl{ years}\sethlcolor{LightBlue}\hl{ post}\sethlcolor{LightBlue}\hl{-}\sethlcolor{LightBlue}\hl{p}\sethlcolor{LightBlue}\hl{regnancy}\sethlcolor{LightBlue}\hl{,}\sethlcolor{LightBlue}\hl{ ac}\sethlcolor{LightBlue}\hl{hes}\sethlcolor{LightBlue}\hl{ and}\sethlcolor{LightBlue}\hl{ pains}\sethlcolor{LightBlue}\hl{ in}\sethlcolor{LightBlue}\hl{ hips}\sethlcolor{LightBlue}\hl{ and}\sethlcolor{LightBlue}\hl{ knees}\sethlcolor{LightBlue}\hl{,}\sethlcolor{LightBlue}\hl{ can}\sethlcolor{LightBlue}\hl{'t}\sethlcolor{LightBlue}\hl{ even}\sethlcolor{LightBlue}\hl{ comfortably}\sethlcolor{LightBlue}\hl{ do}\sethlcolor{LightBlue}\hl{ day}\sethlcolor{LightBlue}\hl{ 1}\sethlcolor{LightBlue}\hl{ of}\sethlcolor{LightBlue}\hl{ C}\sethlcolor{LightBlue}\hl{25}\sethlcolor{LightBlue}\hl{k}\sethlcolor{LightBlue}\hl{.}\sethlcolor{LightBlue}\hl{ Help}\sethlcolor{LightBlue}\hl{?}\sethlcolor{LightRed}\hl{<|endoftext|>} & 1.1442 \\\midrule
PPO Model Response & \sethlcolor{LightRed}\hl{ One}\sethlcolor{LightBlue}\hl{ year}\sethlcolor{LightBlue}\hl{ post}\sethlcolor{LightBlue}\hl{-}\sethlcolor{LightBlue}\hl{p}\sethlcolor{LightBlue}\hl{regnancy}\sethlcolor{LightBlue}\hl{,}\sethlcolor{LightBlue}\hl{ ac}\sethlcolor{LightBlue}\hl{hes}\sethlcolor{LightBlue}\hl{ and}\sethlcolor{LightBlue}\hl{ pains}\sethlcolor{LightBlue}\hl{ in}\sethlcolor{LightBlue}\hl{ hips}\sethlcolor{LightBlue}\hl{ and}\sethlcolor{LightBlue}\hl{ knees}\sethlcolor{LightRed}\hl{ when}\sethlcolor{LightYellow}\hl{ I}\sethlcolor{LightYellow}\hl{ try}\sethlcolor{LightBlue}\hl{ to}\sethlcolor{LightBlue}\hl{ run}\sethlcolor{LightRed}\hl{ even}\sethlcolor{LightYellow}\hl{ the}\sethlcolor{LightYellow}\hl{ smallest}\sethlcolor{LightBlue}\hl{ bit}\sethlcolor{LightBlue}\hl{,}\sethlcolor{LightYellow}\hl{ and}\sethlcolor{LightYellow}\hl{ I}\sethlcolor{LightBlue}\hl{'m}\sethlcolor{LightBlue}\hl{ not}\sethlcolor{LightBlue}\hl{ sure}\sethlcolor{LightBlue}\hl{ if}\sethlcolor{LightRed}\hl{ running}\sethlcolor{LightBlue}\hl{ is}\sethlcolor{LightBlue}\hl{ for}\sethlcolor{LightBlue}\hl{ me}\sethlcolor{LightBlue}\hl{ anymore}\sethlcolor{LightBlue}\hl{.}\sethlcolor{LightYellow}\hl{  }\sethlcolor{LightRed}\hl{Has}\sethlcolor{LightBlue}\hl{ anyone}\sethlcolor{LightBlue}\hl{ had}\sethlcolor{LightBlue}\hl{ something}\sethlcolor{LightBlue}\hl{ similar}\sethlcolor{LightBlue}\hl{?}\sethlcolor{LightBlue}\hl{  }\sethlcolor{LightYellow}\hl{Any}\sethlcolor{LightYellow}\hl{ advice}\sethlcolor{LightYellow}\hl{ would}\sethlcolor{LightBlue}\hl{ be}\sethlcolor{LightBlue}\hl{ appreciated}\sethlcolor{LightRed}\hl{<|endoftext|>} & 5.5148 \\\midrule
DPO Model Response & \sethlcolor{LightRed}\hl{ One}\sethlcolor{LightBlue}\hl{ year}\sethlcolor{LightBlue}\hl{ post}\sethlcolor{LightBlue}\hl{-}\sethlcolor{LightBlue}\hl{p}\sethlcolor{LightBlue}\hl{regnancy}\sethlcolor{LightBlue}\hl{,}\sethlcolor{LightBlue}\hl{ ac}\sethlcolor{LightBlue}\hl{hes}\sethlcolor{LightBlue}\hl{ and}\sethlcolor{LightBlue}\hl{ pains}\sethlcolor{LightBlue}\hl{ in}\sethlcolor{LightBlue}\hl{ hips}\sethlcolor{LightBlue}\hl{ and}\sethlcolor{LightBlue}\hl{ knees}\sethlcolor{LightRed}\hl{ when}\sethlcolor{LightYellow}\hl{ I}\sethlcolor{LightBlue}\hl{ run}\sethlcolor{LightBlue}\hl{,}\sethlcolor{LightBlue}\hl{ can}\sethlcolor{LightBlue}\hl{'t}\sethlcolor{LightBlue}\hl{ even}\sethlcolor{LightBlue}\hl{ comfortably}\sethlcolor{LightBlue}\hl{ do}\sethlcolor{LightBlue}\hl{ day}\sethlcolor{LightBlue}\hl{ 1}\sethlcolor{LightBlue}\hl{ of}\sethlcolor{LightBlue}\hl{ C}\sethlcolor{LightBlue}\hl{25}\sethlcolor{LightBlue}\hl{k}\sethlcolor{LightBlue}\hl{.}\sethlcolor{LightYellow}\hl{  }\sethlcolor{LightRed}\hl{Can}\sethlcolor{LightBlue}\hl{ anyone}\sethlcolor{LightBlue}\hl{ recommend}\sethlcolor{LightBlue}\hl{ some}\sethlcolor{LightBlue}\hl{ stretches}\sethlcolor{LightBlue}\hl{ or}\sethlcolor{LightBlue}\hl{ exercises}\sethlcolor{LightBlue}\hl{ that}\sethlcolor{LightBlue}\hl{ might}\sethlcolor{LightBlue}\hl{ help}\sethlcolor{LightBlue}\hl{?}\sethlcolor{LightBlue}\hl{  }\sethlcolor{LightBlue}\hl{Should}\sethlcolor{LightBlue}\hl{ I}\sethlcolor{LightBlue}\hl{ see}\sethlcolor{LightBlue}\hl{ a}\sethlcolor{LightBlue}\hl{ Dr}\sethlcolor{LightBlue}\hl{?}\sethlcolor{LightRed}\hl{<|endoftext|>} & 4.8368 \\\midrule
Reference response &  I think pregnancy messed with my body, now I can't even run even the smallest amount without pain in my pelvis and knees.  I'm fairly certain the problem isn't just that I'm completely out of shape.<|endoftext|> & 2.4334 \\\midrule
Base Model Response &   I'm a runner, but my knees hurt when I run.  Any suggestions?

A:

I'm a runner, but my knees hurt when I run. Any suggestions?

I'm a runner, but my knees hurt when I run. Any suggestions?

I'm a runner, but my knees hurt when I run. Any suggestions?

I'm a runner, but my knees hurt when I run. Any suggestions?

I'm a runner & -6.3827 \\
\bottomrule
\end{tabular}
\end{small}
\end{table}

\section{Details on the comparison pairs in the preference dataset}
\label{sec:details-pref-paris}

The comparison pairs and their counts can be found in Table~\ref{tab:pref-pair-count-train}, Table~\ref{tab:pref-pair-count-validation-1}, Table~\ref{tab:pref-pair-count-validation-2}, Table~\ref{tab:pref-pair-count-validation-3}, Table~\ref{tab:pref-pair-count-validation-4}, Table~\ref{tab:pref-pair-count-validation-5}, Table~\ref{tab:pref-pair-count-validation-cnndm}.

\begin{table}[]
    \centering
\begin{tabular}{lr}
\toprule
 & count \\
policies &  \\
\midrule
ref, sup2 & 18065 \\
sup2, sup2 & 16273 \\
ref, sup1 & 7947 \\
sup1, sup1 & 7653 \\
sup4\_ppo\_rm3\_kl10, sup4\_ppo\_rm3\_kl10 & 6206 \\
sup4\_ppo\_rm3\_kl20, sup4\_ppo\_rm3\_kl20 & 6098 \\
sup4\_6b\_t0.7, sup4\_6b\_t0.7 & 5614 \\
ref, sup3\_6b & 1788 \\
ref, sup2\_bo8\_rm1 & 1786 \\
sup2\_bo8\_rm1, sup3\_6b & 1751 \\
sup2, sup3\_6b & 1748 \\
sup2, sup2\_bo8\_rm1 & 1738 \\
ref, sup4\_t0.7 & 1667 \\
sup4\_t0.7, sup4\_t0.7 & 1330 \\
ref, sup4\_ppo\_rm3 & 1028 \\
ref, sup3\_bo8\_rm2 & 958 \\
ref, sup3\_ppo\_rm1 & 955 \\
sup3\_bo8\_rm2, sup3\_ppo\_rm1 & 927 \\
sup4\_bo8\_rm3, sup4\_ppo\_rm3 & 775 \\
ref, sup4\_bo128\_rm3 & 669 \\
sup4\_bo128\_rm3, sup4\_bo256\_rm3 & 649 \\
ref, sup3\_bo63\_rm2 & 480 \\
ref, sup3 & 476 \\
sup3\_bo63\_rm2, sup3\_ppo\_rm1 & 470 \\
sup3, sup3\_bo8\_rm2 & 467 \\
sup3\_bo63\_rm2, sup3\_bo8\_rm2 & 464 \\
sup3, sup3\_ppo\_rm1 & 451 \\
sup4\_ppo\_rm3, sup4\_t0.7 & 441 \\
ref, sup4\_bo8\_rm3 & 406 \\
sup4\_ppo\_rm3, sup4\_ppo\_rm3 & 384 \\
ref, sup4\_bo256\_rm3 & 340 \\
sup4\_bo128\_rm3, sup4\_bo128\_rm3 & 322 \\
sup4\_bo64\_rm3, sup4\_ppo\_rm3\_kl10 & 255 \\
ref, sup4\_ppo\_rm3\_kl10 & 253 \\
sup4\_6b\_t0.7, sup4\_ppo\_rm3\_kl20 & 249 \\
sup4\_bo128\_rm3\_6b, sup4\_bo256\_rm3\_6b & 246 \\
ref, sup4\_bo128\_rm3\_6b & 246 \\
ref, sup4\_ppo\_rm3\_kl20 & 245 \\
sup4\_6b\_t0.7, sup4\_ppo\_rm3\_kl10 & 220 \\
sup4\_bo512\_rm3, sup4\_ppo\_rm3\_kl20 & 218 \\
ref, sup4\_6b\_t0.7 & 124 \\
ref, sup4\_bo256\_rm3\_6b & 121 \\
sup4\_bo128\_rm3\_6b, sup4\_bo128\_rm3\_6b & 116 \\
ref, sup4\_bo64\_rm3 & 70 \\
sup4\_6b\_t0.7, sup4\_bo512\_rm3 & 60 \\
sup4\_6b\_t0.7, sup4\_bo64\_rm3 & 56 \\
ref, sup4\_bo512\_rm3 & 53 \\
\bottomrule
\end{tabular}
    \caption{The unique comparison pairs and their counts in the \emph{train} split of the preference dataset.}
    \label{tab:pref-pair-count-train}
\end{table}

\begin{table}[]
    \centering
    \begin{tabular}{lr}
\toprule
 & count \\
policies &  \\
\midrule
ref, sup4\_t0.7 & 3252 \\
sup4\_t0.7, sup4\_t0.7 & 2927 \\
sup4\_6b\_ppo\_rm3\_6b\_kl15, sup4\_6b\_ppo\_rm4\_6b\_kl14 & 2669 \\
sup4\_ppo\_rm3\_kl20, sup4\_ppo\_rm3\_kl20 & 2340 \\
sup4\_ppo\_rm3\_kl10, sup4\_ppo\_rm3\_kl10 & 2070 \\
sup4\_6b\_t0.7, sup4\_6b\_t0.7 & 1828 \\
pretrain\_12b\_t.5, ref & 1682 \\
ref, sup4\_6b\_t0.7 & 1628 \\
ref, sup4\_6b & 1167 \\
ref, sup4\_6b\_ppo\_rm4\_6b & 1154 \\
ref, sup4\_ppo\_rm4 & 1141 \\
sup4\_12b\_t0.7, sup4\_ppo\_rm4\_t.7 & 1097 \\
ref, sup4\_ppo\_rm3\_kl9 & 1084 \\
ref, sup4\_12b & 1026 \\
ref, title & 970 \\
ref, sup4\_3b & 950 \\
ref, sup4 & 934 \\
pretrain\_xl\_t.7, ref & 854 \\
pretrain\_12b\_t.5, sup4\_t0.7 & 847 \\
pretrain\_12b\_t.5, sup4\_ppo\_rm4\_t.7 & 818 \\
pretrain\_xl\_t.7, sup4\_t0.7 & 817 \\
pretrain\_12b\_t.5, pretrain\_xl\_t.7 & 816 \\
ref, sup4\_ppo\_rm4\_t.7 & 814 \\
ref, sup4\_ppo\_rm3\_kl6 & 812 \\
ref, sup4\_ppo\_rm3\_kl69 & 810 \\
ref, sup4\_ppo\_rm3\_kl22 & 810 \\
pretrain\_12b\_t.5, sup4\_6b\_t0.7 & 799 \\
sup4\_6b\_t0.7, sup4\_ppo\_rm4\_t.7 & 796 \\
ref, sup4\_ppo\_rm3\_kl2 & 794 \\
pretrain\_6b, ref & 786 \\
pretrain\_12b, ref & 770 \\
ref, sup4\_ppo\_rm3\_kl260 & 740 \\
ref, sup4\_6b\_ppo\_rm3\_6b\_kl15 & 726 \\
ref, sup4\_12b\_t0.7 & 719 \\
sup4\_12b\_t0.7, sup4\_6b\_ppo\_rm3\_6b\_kl15 & 715 \\
ref, sup4\_3b\_t0.7 & 709 \\
pretrain\_3b, ref & 707 \\
sup4\_6b\_ppo\_rm4\_6b\_kl14, sup4\_6b\_ppo\_rm4\_6b\_kl14 & 670 \\
sup4\_6b\_ppo\_rm3\_6b\_kl15, sup4\_6b\_ppo\_rm3\_6b\_kl15 & 669 \\
sup4\_3b\_t0.7, sup4\_t0.7 & 661 \\
pretrain\_xl, ref & 567 \\
ref, sup3\_6b & 545 \\
ref, sup2 & 543 \\
ref, sup2\_bo8\_rm1 & 542 \\
sup2\_bo8\_rm1, sup3\_6b & 535 \\
ref, sup4\_ppo\_rm3\_kl10 & 527 \\
ref, sup4\_xl\_bo64\_rouge & 525 \\
sup2, sup2\_bo8\_rm1 & 523 \\
sup2, sup3\_6b & 517 \\
ref, sup4\_bo64\_rm3 & 507 \\
\bottomrule
\end{tabular}

    \caption{The unique comparison pairs and their counts in the \emph{validation} split of the preference dataset. (Part 1)}
    \label{tab:pref-pair-count-validation-1}
\end{table}
\begin{table}[]
    \centering
    \begin{tabular}{lr}
\toprule
 & count \\
policies &  \\
\midrule
ref, sup4\_xl\_bo512\_rm4\_6b & 497 \\
sup4\_xl\_bo128\_rouge, sup4\_xl\_bo64\_rouge & 495 \\
sup4\_xl\_bo1024\_rm4\_6b, sup4\_xl\_bo512\_rm4\_6b & 483 \\
ref, sup4\_bo512\_rm3 & 481 \\
ref, sup4\_xl\_bo64\_rm4 & 467 \\
ref, sup4\_bo128\_rm3 & 464 \\
sup4\_xl\_bo128\_rm4, sup4\_xl\_bo64\_rm4 & 453 \\
pretrain\_6b\_t.7, ref & 443 \\
ref, sup4\_6b\_ppo\_rm4\_6b\_kl14 & 442 \\
sup4\_6b\_ppo\_rm4\_6b\_kl14, sup4\_6b\_t0.7 & 440 \\
ref, sup4\_xl\_bo512\_rm4 & 439 \\
pretrain\_6b\_t.7, sup4\_6b\_ppo\_rm4\_6b\_kl14 & 439 \\
ref, sup4\_xl\_bo512\_rouge & 438 \\
pretrain\_6b\_t.7, sup4\_6b\_t0.7 & 436 \\
ref, sup4\_xl\_bo64\_rm4\_6b & 436 \\
sup4\_xl\_bo1024\_rm4, sup4\_xl\_bo512\_rm4 & 432 \\
sup4\_xl\_bo128\_rm4\_6b, sup4\_xl\_bo64\_rm4\_6b & 427 \\
sup4\_bo128\_rm3, sup4\_bo64\_rm3 & 417 \\
sup4\_xl\_bo1024\_rouge, sup4\_xl\_bo512\_rouge & 407 \\
sup4\_bo512\_rm3, sup4\_xl\_bo1024\_rm3 & 403 \\
sup4\_ppo\_rm3\_kl6, sup4\_ppo\_rm3\_kl9 & 387 \\
sup4\_ppo\_rm3\_kl69, sup4\_ppo\_rm3\_kl9 & 380 \\
sup4\_ppo\_rm3\_kl2, sup4\_ppo\_rm3\_kl9 & 351 \\
ref, sup4\_ppo\_rm3 & 345 \\
sup4\_ppo\_rm3\_kl22, sup4\_ppo\_rm3\_kl9 & 325 \\
sup4\_ppo\_rm3\_kl9, sup4\_t0.7 & 322 \\
ref, sup3\_ppo\_rm1 & 315 \\
sup4\_ppo\_rm3\_kl260, sup4\_ppo\_rm3\_kl9 & 307 \\
ref, sup3\_bo8\_rm2 & 306 \\
sup3\_bo8\_rm2, sup3\_ppo\_rm1 & 302 \\
sup4\_bo64\_rm3, sup4\_ppo\_rm3\_kl10 & 302 \\
sup4\_6b\_t0.7, sup4\_ppo\_rm3\_kl10 & 274 \\
ref, sup4\_xl\_bo128\_rouge & 262 \\
sup4\_bo8\_rm3, sup4\_ppo\_rm3 & 261 \\
sup4\_ppo\_rm3\_kl6, sup4\_ppo\_rm3\_kl69 & 260 \\
ref, sup4\_bo8\_rm3 & 248 \\
ref, sup4\_6b\_p.95 & 248 \\
ref, sup4\_xl\_bo1024\_rm4\_6b & 247 \\
sup4\_6b, sup4\_6b\_t0.7 & 245 \\
sup4\_xl\_bo64\_rouge, sup4\_xl\_bo64\_rouge & 244 \\
sup4\_6b\_p.95, sup4\_6b\_t0.7 & 244 \\
sup4\_6b, sup4\_6b\_p.95 & 244 \\
sup4\_bo128\_rm3, sup4\_bo256\_rm3 & 244 \\
sup4\_xl\_bo512\_rm4\_6b, sup4\_xl\_bo512\_rm4\_6b & 242 \\
sup4\_ppo\_rm3\_kl22, sup4\_ppo\_rm3\_kl69 & 242 \\
ref, sup4\_ppo\_rm3\_kl20 & 238 \\
sup4\_6b\_t0.7, sup4\_ppo\_rm3\_kl20 & 234 \\
sup4\_ppo\_rm3\_kl2, sup4\_ppo\_rm3\_kl22 & 234 \\
ref, sup4\_xl\_bo128\_rm4 & 231 \\
sup4\_ppo\_rm3\_kl2, sup4\_t0.7 & 228 \\
\bottomrule
\end{tabular}

    \caption{The unique comparison pairs and their counts in the \emph{validation} split of the preference dataset. (Part 2)}
    \label{tab:pref-pair-count-validation-2}
\end{table}
\begin{table}[]
    \centering
    \begin{tabular}{lr}
\toprule
 & count \\
policies &  \\
\midrule
sup4\_ppo\_rm3\_kl22, sup4\_ppo\_rm3\_kl6 & 228 \\
sup4\_ppo\_rm3\_kl2, sup4\_ppo\_rm3\_kl69 & 226 \\
sup4\_xl\_bo64\_rm4, sup4\_xl\_bo64\_rm4 & 224 \\
sup4\_bo512\_rm3, sup4\_ppo\_rm3\_kl20 & 224 \\
ref, sup4\_xl\_bo1024\_rouge & 219 \\
sup4\_ppo\_rm3\_kl260, sup4\_ppo\_rm3\_kl6 & 219 \\
ref, sup4\_xl\_bo128\_rm4\_6b & 218 \\
ref, sup4\_xl\_bo1024\_rm4 & 218 \\
sup4\_xl\_bo64\_rm4\_6b, sup4\_xl\_bo64\_rm4\_6b & 217 \\
sup4\_ppo\_rm3\_kl22, sup4\_t0.7 & 217 \\
sup4\_xl\_bo512\_rm4, sup4\_xl\_bo512\_rm4 & 217 \\
sup4\_ppo\_rm3\_kl22, sup4\_ppo\_rm3\_kl260 & 216 \\
sup4\_ppo\_rm3\_kl2, sup4\_ppo\_rm3\_kl6 & 215 \\
sup4\_bo64\_rm3, sup4\_bo64\_rm3 & 212 \\
ref, sup4\_xl\_bo1024\_rm3 & 212 \\
sup4\_xl\_bo512\_rouge, sup4\_xl\_bo512\_rouge & 204 \\
sup4\_bo512\_rm3, sup4\_bo512\_rm3 & 203 \\
sup4\_ppo\_rm3\_kl260, sup4\_ppo\_rm3\_kl69 & 196 \\
sup4\_ppo\_rm3\_kl2, sup4\_ppo\_rm3\_kl260 & 186 \\
sup4\_ppo\_rm3\_kl6, sup4\_t0.7 & 184 \\
sup4\_ppo\_rm3\_kl69, sup4\_t0.7 & 183 \\
sup4\_ppo\_rm3\_kl260, sup4\_t0.7 & 179 \\
ref, sup3 & 158 \\
ref, sup3\_bo63\_rm2 & 157 \\
sup4\_6b\_ppo\_rm4\_6b, sup4\_ppo\_rm4 & 156 \\
sup3, sup3\_bo8\_rm2 & 155 \\
sup4\_ppo\_rm3, sup4\_t0.7 & 152 \\
sup3, sup3\_ppo\_rm1 & 151 \\
sup3\_bo63\_rm2, sup3\_ppo\_rm1 & 151 \\
sup3\_bo63\_rm2, sup3\_bo8\_rm2 & 148 \\
ref, sup4\_6b\_t.3 & 146 \\
ref, sup4\_6b\_t.5 & 144 \\
sup4\_ppo\_rm4, title & 143 \\
ref, sup4\_6b\_t1 & 141 \\
sup4\_6b\_t.3, sup4\_6b\_t1 & 140 \\
sup4\_12b, sup4\_ppo\_rm4 & 137 \\
sup4\_6b\_t.5, sup4\_6b\_t1 & 137 \\
sup4\_6b\_t.3, sup4\_6b\_t.5 & 136 \\
ref, sup4\_6b\_ppo\_rm4\_6b\_t.7 & 135 \\
sup4, sup4\_ppo\_rm4 & 134 \\
ref, sup4\_6b\_ppo\_rm4\_6b\_t.5 & 134 \\
sup4\_6b\_ppo\_rm4\_6b\_t.5, sup4\_6b\_ppo\_rm4\_6b\_t.7 & 132 \\
sup4\_3b, sup4\_ppo\_rm4 & 132 \\
sup4\_6b\_ppo\_rm4\_6b, sup4\_6b\_ppo\_rm4\_6b\_t.7 & 132 \\
sup4\_ppo\_rm3, sup4\_ppo\_rm3 & 131 \\
sup4\_12b, sup4\_6b & 131 \\
sup4\_6b\_ppo\_rm4\_6b, sup4\_6b\_ppo\_rm4\_6b\_t.5 & 127 \\
ref, sup4\_bo256\_rm3 & 127 \\
sup4\_12b, title & 127 \\
sup4\_6b, sup4\_6b\_ppo\_rm4\_6b & 127 \\
\bottomrule
\end{tabular}

    \caption{The unique comparison pairs and their counts in the \emph{validation} split of the preference dataset. (Part 3)}
    \label{tab:pref-pair-count-validation-3}
\end{table}
\begin{table}[]
    \centering
    \begin{tabular}{lr}
\toprule
 & count \\
policies &  \\
\midrule
ref, sup4\_6b\_p.7 & 124 \\
ref, sup4\_6b\_p.9 & 124 \\
ref, sup4\_6b\_p.8 & 124 \\
ref, sup4\_xl\_bo8\_rm4 & 124 \\
sup4\_bo128\_rm3, sup4\_bo128\_rm3 & 119 \\
sup4\_xl\_bo16\_rm4, sup4\_xl\_bo8\_rm4 & 118 \\
sup4, sup4\_12b & 116 \\
pretrain\_12b, sup4\_6b\_ppo\_rm4\_6b & 114 \\
sup4\_6b\_p.7, sup4\_6b\_p.8 & 114 \\
sup4, title & 113 \\
sup4\_3b, sup4\_6b\_ppo\_rm4\_6b & 113 \\
ref, sup4\_xl\_bo8\_rouge & 113 \\
sup4\_6b\_p.8, sup4\_6b\_p.9 & 113 \\
sup4\_6b\_p.7, sup4\_6b\_p.9 & 112 \\
sup4\_bo8\_rm3, sup4\_xl\_bo16\_rm3 & 111 \\
sup4\_6b, title & 107 \\
sup4\_6b, sup4\_ppo\_rm4 & 106 \\
sup4\_3b, title & 106 \\
sup4\_t0.7, sup4\_xl\_bo2\_rouge & 106 \\
sup4\_t0.7, sup4\_xl\_bo2\_rm3 & 104 \\
sup4\_xl\_bo16\_rouge, sup4\_xl\_bo8\_rouge & 103 \\
sup4, sup4\_3b & 102 \\
sup4\_12b, sup4\_6b\_ppo\_rm4\_6b & 100 \\
pretrain\_12b, sup4\_ppo\_rm4 & 98 \\
pretrain\_3b, sup4 & 97 \\
sup4\_6b\_ppo\_rm4\_6b, title & 97 \\
sup4\_12b, sup4\_3b & 95 \\
sup4\_t0.7, sup4\_xl\_bo2\_rm4 & 95 \\
sup4\_xl\_bo16\_rm4\_6b, sup4\_xl\_bo8\_rm4\_6b & 94 \\
pretrain\_6b, sup4\_ppo\_rm4 & 93 \\
pretrain\_12b, sup4\_12b & 93 \\
ref, sup4\_xl\_bo8\_rm4\_6b & 92 \\
pretrain\_6b, sup4\_12b & 91 \\
pretrain\_6b, title & 90 \\
pretrain\_3b, pretrain\_6b & 90 \\
pretrain\_12b, sup4\_3b & 90 \\
pretrain\_6b, sup4\_6b & 88 \\
sup4\_ppo\_rm3\_kl10, sup4\_ppo\_rm3\_kl6 & 86 \\
sup4\_ppo\_rm3\_kl10, sup4\_ppo\_rm3\_kl2 & 86 \\
sup4\_t0.7, sup4\_xl\_bo2\_rm4\_6b & 84 \\
sup4, sup4\_6b & 84 \\
sup4\_3b, sup4\_6b & 84 \\
pretrain\_12b, sup4\_6b & 82 \\
pretrain\_6b, sup4\_6b\_ppo\_rm4\_6b & 81 \\
sup4, sup4\_6b\_ppo\_rm4\_6b & 81 \\
pretrain\_3b, sup4\_ppo\_rm4 & 79 \\
pretrain\_3b, sup4\_3b & 79 \\
sup4\_ppo\_rm3\_kl10, sup4\_ppo\_rm3\_kl22 & 77 \\
pretrain\_3b, sup4\_6b\_ppo\_rm4\_6b & 75 \\
pretrain\_6b, sup4\_3b & 74 \\
\bottomrule
\end{tabular}

    \caption{The unique comparison pairs and their counts in the \emph{validation} split of the preference dataset. (Part 4)}
    \label{tab:pref-pair-count-validation-4}
\end{table}
\begin{table}[]
    \centering
    \begin{tabular}{lr}
\toprule
 & count \\
policies &  \\
\midrule
sup4\_6b\_t0.7, sup4\_bo64\_rm3 & 74 \\
pretrain\_6b, sup4 & 72 \\
pretrain\_xl, sup4\_6b\_ppo\_rm4\_6b & 68 \\
sup4\_ppo\_rm3\_kl10, sup4\_t0.7 & 67 \\
pretrain\_3b, sup4\_12b & 67 \\
sup4\_ppo\_rm3\_kl10, sup4\_ppo\_rm3\_kl260 & 65 \\
pretrain\_xl, sup4\_12b & 64 \\
pretrain\_12b, sup4 & 63 \\
sup4\_bo128\_rm3\_6b, sup4\_bo256\_rm3\_6b & 63 \\
ref, sup4\_bo128\_rm3\_6b & 62 \\
ref, sup4\_xl\_bo16\_rm4 & 61 \\
pretrain\_3b, title & 60 \\
sup4\_6b\_t0.7, sup4\_bo512\_rm3 & 60 \\
pretrain\_xl, sup4\_3b & 60 \\
sup4\_ppo\_rm3\_kl10, sup4\_ppo\_rm3\_kl69 & 59 \\
pretrain\_xl, sup4 & 58 \\
sup4\_xl\_bo8\_rm4, sup4\_xl\_bo8\_rm4 & 58 \\
pretrain\_xl, sup4\_ppo\_rm4 & 56 \\
ref, sup4\_xl\_bo2\_rm3 & 56 \\
ref, sup4\_xl\_bo16\_rm3 & 56 \\
sup4\_bo8\_rm3, sup4\_bo8\_rm3 & 56 \\
pretrain\_xl, title & 56 \\
ref, sup4\_xl\_bo2\_rouge & 55 \\
ref, sup4\_xl\_bo16\_rouge & 55 \\
sup4\_xl\_bo8\_rouge, sup4\_xl\_bo8\_rouge & 55 \\
pretrain\_12b, pretrain\_xl & 54 \\
pretrain\_12b, pretrain\_3b & 51 \\
ref, sup4\_xl\_bo2\_rm4 & 48 \\
pretrain\_12b, title & 47 \\
pretrain\_3b, pretrain\_xl & 47 \\
ref, sup4\_xl\_bo2\_rm4\_6b & 47 \\
pretrain\_12b, pretrain\_6b & 47 \\
ref, sup4\_xl\_bo16\_rm4\_6b & 47 \\
sup4\_xl\_bo8\_rm4\_6b, sup4\_xl\_bo8\_rm4\_6b & 46 \\
pretrain\_3b, sup4\_6b & 45 \\
pretrain\_xl, sup4\_6b & 45 \\
pretrain\_6b, pretrain\_xl & 44 \\
sup4\_bo128\_rm3\_6b, sup4\_bo128\_rm3\_6b & 33 \\
ref, sup4\_bo256\_rm3\_6b & 31 \\
human\_editor, sup4\_6b\_t0.7 & 3 \\
human\_editor, ref & 2 \\
\bottomrule
\end{tabular}

    \caption{The unique comparison pairs and their counts in the \emph{validation} split of the preference dataset. (Part 5)}
    \label{tab:pref-pair-count-validation-5}
\end{table}

\begin{table}[]
    \centering
    \begin{tabular}{lr}
\toprule
 & count \\
policies &  \\
\midrule
supcnndm3\_6b\_t.3, supcnndm3\_6b\_t.3 & 1410 \\
pretrain\_6b\_t.7, sup4\_6b\_ppo\_rm4\_6b\_kl14 & 148 \\
ref, sup4\_6b\_t0.7 & 148 \\
ref, sup4\_6b\_ppo\_rm4\_6b\_kl14 & 148 \\
pretrain\_6b\_t.7, ref & 146 \\
sup4\_6b\_ppo\_rm4\_6b\_kl14, sup4\_6b\_t0.7 & 144 \\
pretrain\_6b\_t.7, sup4\_6b\_t0.7 & 140 \\
\bottomrule
\end{tabular}

    \caption{The unique comparison pairs and their counts in the \emph{validation\_cnndm} split of the preference dataset. (Part 5)}
    \label{tab:pref-pair-count-validation-cnndm}
\end{table}

\section{Author Contributions}

\begin{itemize}
    \item Shengyi Huang led the overall project.
    \item Michael Noukhovitch helped discuss and verify early design choices/results, led the analysis of RM calibration plots, and edited the paper.
    \item Arian Hosseini led the analysis of the length-controlled summary comparisons (e.g., Figure~\ref{fig:log-length}), improved visualization in \ref{detail:value-network-improve}, and edited the paper.
    \item Kashif Rasul crafted the visualization in Table~\ref{tab:ai2-visual} and edited the paper.
    \item Weixun Wang plotted the GPT3.5 agreement rate in Figure~\ref{fig:gpt3.5-rm-agreement} (left) and Table~\ref{tab:various-accuraries} and edited the paper.
    \item Lewis Tunstall advised the project.
\end{itemize}

\end{document}